# FairFLRep: Fairness aware fault localization and repair of Deep Neural Networks

MOSES OPENJA, Polytechnique Montréal, Canada
PAOLO ARCAINI, National Institute of Informatics, Japan
FOUTSE KHOMH, Polytechnique Montréal, Canada
FUYUKI ISHIKAWA, National Institute of Informatics, Japan

Deep neural networks (DNNs) are being utilized in various aspects of our daily lives, including high-stakes decision-making applications that impact individuals. However, these systems reflect and amplify bias from the data used during training and testing, potentially resulting in biased behavior and inaccurate decisions. For instance, having different misclassification rates between white and black sub-populations. However, effectively and efficiently identifying and correcting biased behavior in DNNs is a challenge. This paper introduces FairFLRep, an automated fairness-aware fault localization and repair technique that identifies and corrects potentially bias-inducing neurons in DNN classifiers. FairFLRep focuses on adjusting neuron weights associated with sensitive attributes, such as race or gender, that contribute to unfair decisions. By analyzing the input-output relationships within the network, FairFLRep corrects neurons responsible for disparities in predictive quality parity. We evaluate FairFLRep on four image classification datasets using two DNN classifiers, and four tabular datasets with a DNN model. The results show that FairFLRep consistently outperforms existing methods in improving fairness while preserving accuracy. An ablation study confirms the importance of considering fairness during both fault localization and repair stages. Our findings also show that FairFLRep is more efficient than the baseline approaches in repairing the network.

## 1 Introduction

Deep neural networks (DNNs) have witnessed rapid advancement in recent years, revolutionizing various industries with their ability to learn complex patterns from large datasets. From autonomous systems and medical diagnostics [10, 26, 45] to natural language processing (NLP) and image recognition [41, 60], DNNs are now at the core of many high-stakes applications that directly affect individuals' lives. In these domains, the decisions made by DNN-based models carry significant weight. For instance, an image recognition system may determine an individual's identity or diagnosis, while a hiring algorithm might influence employment outcomes. These applications demand not only high accuracy but also *fairness*, i.e., impartial treatment of different sensitive groups under classification.

Despite their powerful capabilities, DNNs are prone to inheriting biases present in their training data. These biases can manifest in unfair predictions, disproportionately affecting certain sensitive

Authors' Contact Information: Moses Openja, openja.moses@polymtl.ca, Polytechnique Montréal, Montreal, Quebéc, Canada; Paolo Arcaini, arcaini@nii.ac.jp, National Institute of Informatics, Tokyo, Japan; Foutse Khomh, foutse.khomh@polymtl.ca, Polytechnique Montréal, Montreal, Quebéc, Canada; Fuyuki Ishikawa, f-ishikawa@nii.ac.jp, National Institute of Informatics, Tokyo, Japan.

groups. For example, a facial recognition model may show higher misclassification rates for darker-skinned individuals compared to lighter-skinned individuals, as shown in [8, 39]. In such cases, DNN models may perpetuate or even amplify societal inequalities. This is particularly concerning for Convolutional Neural Networks (CNNs), which are widely used, as they may unintentionally learn to prioritize features correlated with sensitive attributes, such as race or gender, leading to unfair outcomes [54, 55, 75].

As DNNs are increasingly integrated into decision-making systems that impact people's lives, addressing fairness issues has become critical. Unfair models can cause considerable harm, particularly in sensitive applications like hiring, loan approvals, or facial recognition. These biases not only lead to unjust outcomes but also erode public trust in AI systems [50, 61]. Fairness of AI systems is now required by different international regulations; for example, the EU AI Act, in its Recital 27, requires that AI systems avoid "discriminatory impacts and unfair biases"; the U.S. Equal Employment Opportunity Commission (EEOC) in its guidelines requires that the selection rate for an unprivileged group should be at least 80% of the selection rate for the privileged group [67]; GDPR requires that data is "processed lawfully, fairly and in a transparent manner in relation to the data subject" [22].

Fairness in DNNs is commonly tackled through *group fairness* (ensuring equal treatment for different sensitive groups) or *individual fairness* (ensuring similar treatment for similar individuals). While various fairness metrics [29, 57, 74] (e.g., *Statistical Parity Difference* (*SPD*), *Disparate Impact* (*DI*), *Equalized Odds* (*EOD*), and *False Positive Rate* (*FPR*)) have been developed, addressing unfairness in practice remains challenging, as sensitive attributes are often not explicitly available but inferred by the model. In other words, CNNs may make biased decisions based on features correlated with sensitive attributes like race or gender, even if those attributes are not provided directly as inputs.

Traditionally, methods to address fairness in DNNs have focused on *pre-processing* (modifying the training data), *in-processing* (modifying the model's training process), or *post-processing* (adjusting predictions after training) [29]. While these techniques can mitigate some bias, they come with limitations. Pre-processing, for example, balances the training data but does not always resolve the root cause of bias encoded in the network's learned representations, as neural networks can still pick up subtle correlations between features and sensitive attributes. In-processing methods, which modify the training process, or post-processing techniques, which adjust predictions, can also be computationally expensive, require retraining, and may fail to generalize well across different datasets. Moreover, retraining the model—while capable of modifying weights—might not efficiently address the specific neurons responsible for biased behavior unless fairness-specific objectives are incorporated into the training process. Retraining can be computationally costly, especially for large models, and may have unintended consequences such as degrading overall model accuracy without fully addressing fairness concerns. This highlights the need for more targeted approaches that directly intervene at the level of bias-inducing neurons to achieve fairer outcomes without the overhead of retraining the entire model.

In response to these challenges, we propose applying repair techniques that have proven effective in other contexts—such as model debugging [35, 42, 43, 47, 48, 59, 62, 64, 66, 76] (i.e., process of diagnosing, and fixing issues or errors in a DL model) and improving robustness [73]—to mitigate fairness issues in DNNs. Instead of retraining the model or modifying the data, these approaches focus on repairing the trained model by directly targeting and repairing the problematic neurons responsible for bias. This method has been successful in improving model accuracy and robustness in other domains, as demonstrated by `Arachne` [64]. We believe that repair techniques can also effectively mitigate fairness issues in DNNs by targeting specific neural weights responsible for biased decisions. This approach enables precise, neuron-level interventions that improve fairness without significantly affecting model performance. It is particularly advantageous in real-world scenarios where retraining a model from scratch may be computationally infeasible or impractical due to time constraints.

Given the complexity of CNNs and their widespread use, an efficient repair technique that targets biased neurons is particularly necessary. In addition to improving fairness, this method should minimize disruptions to the model's accuracy and be computationally efficient, making it practical for deployment in real-world systems. To address these challenges, we introduce `FairFLRep`, an automated fairness-aware fault localization and repair technique specifically designed for DNN classifiers. In this context, a *fault* refers to a *biased behavior* (or *bias bug*) within the model, where certain neuron weights contribute to unfair predictions against specific sensitive groups. `FairFLRep` identifies potentially biased neurons in a DNN by analyzing the input-output relationships within the network. The method prioritizes identifying and correcting neuron weights associated with sensitive attributes, such as race or gender, which contribute to unfair decisions. By focusing on the neurons most responsible for biased behavior, `FairFLRep` offers a targeted repair strategy that minimizes the impact on overall model accuracy. `FairFLRep` operates in two key stages:

(1) *Fairness-aware fault localization* identifies neural weights in the network that contribute the most to unfair behavior. In this step, the fault is understood as the bias bug embedded in the neural weights, which causes unfair treatment of certain subgroups. The goal is to pinpoint the specific neuron weights that amplify these biases.
(2) *Fairness-aware repair* corrects these neural weights to mitigate their contribution to bias, ensuring that the model produces fairer predictions.

We conduct an extensive empirical evaluation of `FairFLRep` across multiple image classification benchmarks (i.e., `FairFace` [33], `UTKFace` [79], Labeled Faces in the Wild (`LFW`) [31], and `CelebA` [46]), tabular classification benchmarks (`Student` [15], `COMPAS` [68], `Adult` [4], and `MEPS` [2]), and fairness metrics (*SPD*, *DI*, *FPR* and *EOD*). Our results reveal that `FairFLRep` not only improves fairness, but also preserves model accuracy, consistently outperforming existing methods such as `Arachne`. Moreover, `FairFLRep` is designed to be computationally efficient, enabling faster execution times and making it scalable for large-scale applications.

In summary, we make the following main contributions:

- We propose `FairFLRep`, a novel fairness-aware fault localization and repair technique designed to address bias in DNNs by analyzing input-output relationships and correcting biased neurons. `FairFLRep` provides flexibility in addressing biases directly tied to sensitive attributes (e.g., race, gender) and those indirectly influencing predictions, ensuring comprehensive bias mitigation.
- We conduct an extensive empirical evaluation across four widely used image classification datasets and four tabular classification benchmarks, demonstrating `FairFLRep`'s superiority over existing methods in improving fairness across multiple fairness metrics (*SPD*, *DI*, *EOD*, *FPR*) while preserving accuracy.
- We perform an ablation study confirming the importance of applying fairness constraints during both the fault localization and repair stages, resulting in more robust fairness improvements.
- `FairFLRep` offers computational efficiency, with significantly faster execution times compared to the baseline approaches, ensuring scalability for large datasets and complex models.
- By explicitly targeting biased neural behaviors and reducing subgroup disparities, `FairFLRep` advances the development of AI systems that are not only technically robust but also legally and ethically compliant with fairness mandates such as those in EU AI Act, GDPR, and EEOC regulation.
- We make the code of `FairFLRep` available online at https://github.com/openjamoses/FairFLRep.

4 Moses Openja, Paolo Arcaini, Foutse Khomh, and Fuyuki Ishikawa

**Paper organization.** Section 2 provides background information on DNNs and fairness, offering the foundational concepts necessary for understanding the fairness issues in DNNs and also discussed fairness metrics. Section 3 discusses related works on fairness testing and mitigation techniques in the context of DNNs, reviewing existing approaches and their limitations. Section 4 details the proposed `FairFLRep` technique introduced in this study, outlining the methodology for fairness-aware fault localization and repair in DNNs. Section 5 presents the experiment design used to evaluate `FairFLRep`, including the datasets, models, and the compared approaches. Section 6 reports the evaluation results. Section 7 outlines potential threats to the validity of this work, acknowledging limitations and areas for further research. Section 8 further highlights key insights of our findings, lessons learned, the broader implications and finally concludes the paper.

## 2 Preliminary

### 2.1 Notation and Definition

In this study, we focus on addressing the fairness problem within the context of binary classification. Unless otherwise specified when describing a given algorithm or implementation, the following notations are used consistently throughout this paper:

- $\mathcal{M}$ – *Model*: denotes the DNN model for binary classification.
- $\mathcal{D}$ – *Dataset*: it is the dataset used in the problem (e.g., training, validation, and test sets). The dataset used in this study consists of binary classification labels (e.g., race, gender).
- $(X, Y, S)$: they are the variables associated with the DNN model $\mathcal{M}$ being repaired:
  - $X$ – *Input Feature Vector*: it is the input data to the model.
  - $Y \in \{0, 1\}$ – *True Label*: the ground truth binary label (class label) for the instance.
  - $\hat{Y}$ – *Predicted label*: it is the label predicted by the model (i.e., $\hat{Y} = \mathcal{M}(X)$).
  - $S \in \{s_0, s_1\}$ – *Sensitive Attribute* (or *protected attribute*): it is a binary feature in the dataset that is considered sensitive due to its potential to contribute to unfair treatment, discrimination, or bias (e.g., race, gender). For example, if $S$ is gender, $s_0$ can represent the female community and $s_1$ can represent the male community. Unless otherwise specified, we assume that objects satisfying: $S = s_0$ represent the deprived community (e.g., female), while objects satisfying $S = s_1$ represent the favored community (e.g., male). However, our method dynamically determines whether a sub-group belongs to the deprived or favored community (see Section 4.5.2).
- In our method, these two datasets are used:
  - $\mathcal{D}rep$ – *Repair Dataset*: it represents the dataset used during the repair process. This dataset is not seen during training but is used to guide the fairness improvements in the model. The data points of $\mathcal{D}rep$ are further categorized into positive and negative sets, defined as follows:
    * $\mathcal{D}rep_{pos}$ – *Positive set*: it is the set of instances in the dataset that are correctly classified by the model, i.e., $\mathcal{D}rep_{pos} = \{(X, Y, S) \in \mathcal{D}rep \mid \hat{Y} = Y\}$.
    * $\mathcal{D}rep_{neg}$ – *Negative set*: it is the set of instances in the dataset that are misclassified by the model (predicted label differs from the true label), i.e., $\mathcal{D}rep_{neg} = \{(X, Y, S) \in \mathcal{D}rep \mid \hat{Y} \neq Y\}$.
  - $\mathcal{D}_{test}$ – *Test Dataset*: it is the dataset used for evaluating the effectiveness of the repair approach and comparing it to baseline models. This dataset is also unseen during training and repair.
- $\mathcal{F}$ – *Fairness Metric*: it represents the fairness metric (e.g., *SPD, DI, FPR, EOD*) used to evaluate fairness in binary classification.

## 2.2 Deep Neural Network

Deep Neural Networks (DNNs) $\mathcal{M}$ are a class of artificial neural networks (ANNs) characterized by multiple layers between the input and output layers. The depth of a DNN allows it to capture intricate patterns and relationships in large datasets, making it suitable for various applications, such as autonomous systems, image and speech recognition, and natural language processing. A typical DNN consists of: (i) *Input Layer*: it receives the input data. (ii) *Hidden Layers*: they consist of multiple neurons. These layers progressively extract more abstract features as data flows through the network. For example, in image classification tasks, early hidden layers capture basic visual patterns (edges, textures), while deeper hidden layers detect higher-level features (shapes, objects). (iii) *Output Layer*: it produces the final output, typically through a softmax function for classification tasks or a linear function for regression tasks.

*2.2.1 Neurons in DNN.* Neurons are the fundamental units of a DNN, analogous to biological neurons in the human brain. Each of the neurons in a DNN plays a critical role in extracting and representing features from the input data. As data passes from the input layer to the final prediction layer, the features extracted by neurons become increasingly abstract. In the following, we provide a detailed breakdown of how neurons and their associated weights function within a DNN, and how bias in the model can affect its performance:

*Feature Extraction*: Neurons receive raw data through the input layer. For example, in image recognition, these might be pixel values. As data moves through hidden layers, neurons extract progressively higher-level features. Early layers might capture simple patterns like edges or textures, while deeper layers might capture complex patterns like shapes or objects. The final layer compiles these abstract features to make a prediction. In classification tasks, this layer outputs probabilities of different classes.

*Representation of Features*: Each neuron represents a combination of features it receives from the previous layer. The activation function of the neuron determines the output based on the weighted sum of inputs. Common activation functions include the sigmoid, hyperbolic tangent (tanh), and rectified linear unit (ReLU). Weights are parameters that define the importance of each input feature. During training, the model adjusts these weights to minimize error, thereby learning which features are more significant for the task.

*2.2.2 Feature Importance and Unfairness.* The importance of different features is encoded in the weights. Neurons with higher weights for certain features indicate those features are more critical for the decision-making process. If a model is biased, it might focus on inappropriate features. For instance, in a hiring AI, if the model gives high weight to the gender feature instead of skills or experience, it indicates bias. This can lead to unfair and unethical decisions.

This research focuses on addressing the fairness problem in Convolutional Neural Networks (CNNs). CNNs utilize both forward and backward propagation for learning. Below, we describe how the forward impact and gradient loss are computed, on which our proposed method is based.

*2.2.3 Forward impact of the DNN.* During the forward propagation, the input data is processed sequentially through each layer of the network to produce an output. Each neuron in the hidden layers accepts the data, performs a computation using its activation function (and weight and bias), and passes the result to the next layer. This process continues until the final output is generated.

*2.2.4 Loss Function.* Once the neural network output is generated, the loss function $\mathcal{L}$ is used to quantify the difference between the predicted output $\hat{Y}$ and the actual output $Y$. For example, the cross-entropy loss for classification tasks is defined as: $\mathcal{L}(Y, \hat{Y}) = -\sum Y log(\hat{Y})$.

*2.2.5 Backpropagation.* It is the method by which the network computes the gradient of the loss function with respect to the model's weights and biases. Using this gradient, the network updates its weights through gradient descent, enabling it to learn from its errors and improve its predictions.

Section 2.3 sets the foundation for understanding the fairness concerns related to the forward and backward propagation processes in DNNs, which play a pivotal role in how bias is embedded and addressed during training. Section 2.4 describes how fairness issues manifest in DNNs and methods to measure them.

## 2.3 DNN Fairness

Bias in DNN models can arise when sensitive attributes, such as race or gender, inadvertently influence decision-making, even if these attributes are not provided directly as inputs. For instance, an AI system designed for hiring might inadvertently favor certain genders over others, basing decisions on attributes indirectly correlated with gender instead of qualifications. This type of bias can subtly manifest in the model's internal representations or activations, emphasizing certain features that correlate with protected attributes. Although it is difficult to directly link high-level features to specific model weights, these biased representations can still impact decision-making, potentially skewing outcomes. Bias in DNNs can be broadly categorized into either *prediction outcome discrimination* and *prediction quality disparity* [18].

*2.3.1 Prediction Outcome Discrimination.* It occurs when DNN models deliver unequal treatment to individuals based on membership in sensitive groups. For example, a recruiting AI could exhibit bias by favoring male candidates over female candidates. Training data often includes biases, whether from inherent noise or annotators' subjective judgments [28]. As DNNs are designed to fit these data, they may reinforce these biases or even develop further implicit associations, amplifying societal stereotypes [80]. This can result in algorithmic discrimination, which has critical implications for access to resources in fields like hiring, lending, and facial recognition. Outcome discrimination can be further classified into:

*Discrimination via Input*: Discrimination in outcomes can stem from input data even when sensitive attributes, like race or gender, are not explicitly included. DNN models may inadvertently link these attributes through correlated features, leading to biased predictions [32].

*Discrimination via Representation*: This refers to situations where predictive outcome discrimination is analyzed from a representational perspective, particularly when attributing bias to the input is nearly impossible. For instance, a CNN might identify gender from an image of the retina, even though this information is undetectable by humans. Such cases result in varying intermediate representations for different sensitive groups. Decisions are thus made based on this indirectly encoded group information, which can lead to biased classification outcomes detectable within the model's deep representations.

*2.3.2 Prediction Quality Disparity.* DNN models sometimes deliver lower-quality predictions for certain sensitive groups, often due to underrepresentation in the training data. When specific populations are either inadequately represented or have less reliable data, the model may attempt to minimize the overall error, often leading to a performance gap between majority and minority groups [9, 19]. While optimizing accuracy for the overall population, this approach can result in subpar performance for underrepresented groups, creating a disparity in prediction quality for these populations.

## 2.4 Measurements of Fairness

Fairness metrics $\mathcal{F}$ are quantitative measures used to quantify/evaluate fairness and inequality in classification models, particularly in terms of how they treat different sensitive groups (e.g., based on race, gender, age) [50]. Fairness metrics are important because they help ensure that the outcomes of machine learning models do not disproportionately favor or harm specific groups, addressing concerns about bias and discrimination. This study will focus on *group fairness*, which seeks for equal statistical measures for the individual across the group according to a particular protected attribute, and statistics about model predictions is calculated for each group and compared across groups. Several group fairness metrics are commonly used:

*Statistical Parity Difference* (*SPD*): This metric [20] compares the probability of individuals receiving the desired algorithmic decisions (positive outcomes) between an unprivileged class (i.e., $S = s_0$) and a privileged class ($S = s_1$), such as race or gender. It defines fairness as different subgroups having an equal probability of being classified with the positive label. Formally, this is expressed as: $SPD := P(\hat{Y} = 1 \mid S = s_0) - P(\hat{Y} = 1 \mid S = s_1)$. Ideally, a value of $SPD = 0$ would indicate that the model's prediction decision $\hat{Y}$ is statistically independent of the sensitive attribute $S$, suggesting a fair model. However, achieving an exact $SPD = 0$ is nearly impossible, so the value is typically compared against a threshold $SPD \leq \epsilon$. The threshold $\epsilon$ is determined based on specific real-world requirements. Sensitive parity is independent of the ground truth labels, making it particularly useful in situations where reliable ground truth information is not available, such as in employment and justice contexts.

When a ratio is used instead of a difference to represent the equality of predictions across different groups, the fairness metric is referred to as *Disparate Impact* (*DI*). Formally, it is defined as: $DI := \frac{P(\hat{Y}=1|S=s_0)}{P(\hat{Y}=1|S=s_1)}$. In this case, a value $DI = 1$ would indicate a fair model. In real-world situations, the value of *DI* is compared with a fairness threshold, such as $DI \geq \tau$. The choice for $\tau$ is usually guided by the social implication, for example, a threshold of 0.8 represents the 80% rule [23]. In this context, the 80% rule states that the selection rate for a disadvantaged (unprivileged) group should be at least 80% of the selection rate for the advantaged (privileged) group. If the ratio falls below 80%, it may indicate potential discrimination or an unfair impact on the unprivileged group. Moreover, to remain consistent with other fairness metrics, *DI* can be converted to $abs(1 - DI)$, indicating that a lower value is better (0 means no bias).

*Equality of Opportunity/Equalized Odds* (*EOD*): A limitation of *SPD* and *DI* is that they do not consider potential differences in the compared subgroups. *EOD* overcomes this by taking into consideration the distributions of different groups in terms of the true positive rate of the label. This is expressed as $P(\hat{Y} = 1 \mid S = s, Y = 1) = P(\hat{Y} = 1 \mid Y = 1), \forall s$. Essentially, this compares the true positive rate (*TPR*) across different groups. In other words, the model's predictions should be equally accurate across different sensitive groups when conditioned on the true outcome. A similar measurement can be calculated for the false positive rate (*FPR*) (i.e., the probability of incorrectly predicting a positive outcome): $P(\hat{Y} = 1 \mid S = s, Y = 0) = P(\hat{Y} = 1 \mid Y = 0) \forall s$. By enforcing both true positive rate parity and false positive rate parity simultaneously (i.e., considering both $Y = 1$ and $Y = 0$), we achieve *Equalized Odds* as described by [29]: $P(\hat{Y} = 1 \mid S = s) = P(\hat{Y} = 1)\forall s$.

*Predictive Quality Parity*: This metric measures prediction quality difference between different subgroups. The quality denotes quantitative model performance in terms of model predictions and ground truth. In this paper we focus on accuracy for binary classification.

While these metrics provide important tools for assessing fairness in DNNs, they do not always capture all dimensions of bias. Some metrics, like counterfactual fairness [37] and fairness through unawareness [37], causal fairness [11] explore alternative dimensions, but the metrics discussed above form the foundation for measuring group fairness in DNN models. Understanding these metrics is crucial for designing and evaluating fairness-aware systems.

## 2.5 Fault Localization and Repair in DNN

Fault localization [70] is a debugging technique aimed at pinpointing the defective components of a program based on testing outcomes. In conventional software, a program component (e.g., a statement) is deemed highly faulty or suspicious if it is covered by more test cases that reveal failures and fewer test cases that pass. This concept has been extended to other software systems, including DNNs. In [64], a fault localization method and a repair method for DNNs are proposed, which operate as follows:

(1) it selects a subset of positive and negative inputs, where positive inputs are instances correctly classified by the model $\mathcal{M}$, and negative inputs are those misclassified by $\mathcal{M}$. Negative inputs reveal the model's incorrect behavior and highlight areas needing correction;
(2) the forward impact and the gradient loss of each weight connected to the last layer are calculated. The gradient loss is determined using classic backpropagation. The rationale for focusing solely on the weights of the last layer is to minimize the search space;
(3) weights that have both high forward impact and high gradient loss are extracted (i.e., those on the Pareto Front of these two metrics). The rationale is that these weights are likely to be the most responsible for the model's incorrect behavior (high gradient loss) while also significantly influencing its decisions (forward impact);
(4) in the repair phase, the identified neuron weights are adjusted to reduce bias and improve fairness within the DNN. The process involves searching for alternative weight values that adjust the model's behavior to improve fairness without causing a notable drop in overall accuracy.

## 3 Related Work

Li et al. [40] introduce Faire, a method that aims to repair fairness in neural networks by focusing on neuron conditions and modifying the model's internal structure to mitigate bias. While this approach shares some similarities with our proposed `FairFLRep` technique, there are key differences and limitations of Faire in comparison to our method: (i) The Faire method primarily targets individual fairness and is designed for tabular data, which presents significant challenges when trying to apply it to image classification tasks like the ones we are addressing in this study. In the context of group fairness for DNNs on image datasets (e.g., `UTKFace`, `FairFace`), Faire's focus on individual fairness rather than group-level disparities (e.g., race, gender) makes it less applicable. (ii) Faire uses DeepLIFT, an external framework for analyzing neuron contributions, to identify biased neurons. While DeepLIFT is effective for neuron attribution, its integration into the Faire workflow introduces additional complexity when adapting it for different model architectures or fairness scenarios. `FairFLRep`, in contrast, performs neuron analysis directly based on gradient-based fault localization, which is more integrated with the fairness repair process and does not require external frameworks.

`Arachne` [64] is a more recent method designed to repair DNNs by addressing misclassification errors through fault localization and model repair techniques. Although not originally intended for fairness, `Arachne` can be adapted by prioritizing input data from deprived subgroups to identify neurons linked to biased predictions. However, `Arachne` focuses primarily on general model repair and not fairness-specific issues. When adapted for fairness, it may overlook more complex bias patterns, especially if not properly tuned for fairness-specific tasks.

Maheshwari and Perrot [49] present FairGrad, a method that integrates fairness considerations directly into the training process by modifying the gradient descent optimization. Its main limitations are: (i) FairGrad modifies the optimization algorithm, which may limit its application to models where gradient-based optimization is appropriate. In contrast, `FairFLRep` focuses on neuron analysis and repair, targeting specific areas of the model that contribute to unfair behavior, providing more targeted intervention at the neuron level. (ii) FairGrad applies fairness at the gradient level, affecting all weights in the model during training. `FairFLRep` takes a more localized approach, identifying and modifying specific neurons responsible for bias, allowing for finer-grained control over fairness while preserving model performance. (iii) While FairGrad integrates fairness into the overall model optimization, it does not explicitly distinguish between the fault localization and repair stages as `FairFLRep` does. `FairFLRep` provides a two-stage process: first identifying the unfair behavior (fault localization), and then applying targeted repairs, which may result in a more precise fairness intervention.

FairGRAPE proposed by Lin et al. [44] is a technique designed to mitigate bias in DNNs, particularly in the domain of face attribute classification. The method introduces fairness into the model by pruning certain neurons, based on gradient information, to reduce unfair behavior while maintaining the model's overall performance. Model pruning—the process of removing unnecessary or redundant neurons to make the model more efficient—is a well-known technique in deep learning for reducing model size and complexity. FairGRAPE extends this idea by incorporating fairness into the pruning process, aiming to reduce bias along with the usual benefits of pruning, such as lower computational costs and memory usage. The pruning process in FairGRAPE is driven by fairness considerations rather than just model efficiency, ensuring that fairness is improved as neurons are pruned. Its main limitations are: (i) FairGRAPE removes neurons that are associated with biased behavior. In contrast, `FairFLRep` focuses on repairing the faulty neurons responsible for unfair behavior, which may result in finer control over the fairness intervention. `FairFLRep` does not necessarily reduce the model size but adjusts neuron weights to mitigate bias. (ii) FairGRAPE is constrained by its pruning approach, meaning that it reduces the model's capacity by removing neurons. In contrast, `FairFLRep` takes a broader approach by identifying and repairing unfair neurons, allowing the model to retain its full capacity while still improving fairness. (iii) FairGRAPE needs to be applied during training, making it less flexible for situations where a pretrained model must be repaired for fairness. On the other hand, `FairFLRep` can be applied post-training, offering greater flexibility for real-world use cases where retraining a model may not be feasible.

FairNeuron [25], introduced by Gao et al., improves DNN fairness through adversarial training, focusing on selective neurons that contribute to biased predictions. FairNeuron provides a lightweight approach that is able to scale to large models, and showed its effectiveness on three datasets. Despite its advantages, there are still limitations: (i) FairNeuron uses adversarial training, which requires an additional adversarial network. This adds complexity, while `FairFLRep` focuses on post-training repair, which is more straightforward and applicable to pretrained models. (ii) FairNeuron is primarily designed for MLP architectures with tabular inputs. Its direct applicability to CNNs or hybrid models is limited due to assumptions about fixed neuron layer sizes. Additionally, FairNeuron assumes the availability of two types of labels—binary ground truth labels and regression-style labels (e.g., 1 to 10) used for path-based contribution analysis. Such label configurations may not be available in many real-world settings, particularly in binary classification tasks. In contrast, `FairFLRep` generalizes across diverse architectures, including convolutional models and domain-shifted scenarios (e.g., image datasets), making it suitable for a broader range of real-world applications. (iii) The simultaneous training of both the main network and the adversary in FairNeuron increases computational overhead. `FairFLRep` avoids this by focusing on fault localization and repair without the need for an adversarial network.

MOON (Multi-Objective White-Box Test Input Selection) [27] is a spectrum-based fault localization technique originally developed for test input selection. It identifies suspicious neurons contributing to erroneous decisions by analyzing neuron activation spectra and employs multi-objective optimization to select discriminative inputs. These inputs are then used to retrain the model in an attempt to mitigate bias. While MOON is effective for identifying problematic inputs and exposing model weaknesses, its fairness repair mechanism—retraining based on test input activation patterns—is inherently indirect. It does not directly target or adjust the weights of the neurons responsible for biased outcomes. Moreover, its reliance on statistical correlations from neuron activation frequencies limits its ability to capture causal dependencies driving group-level disparities. In contrast, `FairFLRep` performs gradient-based fault localization and explicitly identifies and modifies biased neuron weights, enabling precise and effective fairness interventions without full retraining. As such, while MOON contributes to fairness-aware testing, its repair mechanism is less fine-grained and insufficient for deeply embedded biases in complex DNNs.

`NeuronFair` [81] identifies biased neurons and employs white-box fairness testing techniques for better interpretability in model decision-making. Unlike `NeuronFair`, which remains primarily diagnostic, `FairFLRep` offers an actionable repair phase, directly mitigating the biases it identifies rather than merely reporting them.

`Fairify` [7] introduces a framework for verifying fairness properties in neural networks using formal methods. `Fairify` aims to provide formal guarantees by encoding fairness constraints (i.e., individual fairness) into SMT problems and systematically verifying whether the model satisfies these constraints across all possible inputs. However, `Fairify` does not allow to improve the model (as done by `FairFLRep`) once unfairness is detected. Moreover, `Fairify` primarily targets smaller, fully connected networks due to the computational complexity of formal verification, limiting its scalability for deep architectures like CNNs. In contrast, `FairFLRep` is designed for efficient post-training fairness repair in large-scale deep models.

Latent Imitator (`LIMI`) [71] is a method for generating natural discriminatory instances for black-box fairness testing. `LIMI` approximates a surrogate decision boundary in the latent space of a generative model and probes samples near this boundary to generate discriminatory test inputs that retain naturalness and semantic realism. To further demonstrate the importance of the naturalness of the generated instances, the authors of `LIMI` conducted retraining experiments, combining the generated discriminatory instances with the original training data. They then evaluated fairness improvements using metrics such as IFR (Individual Fairness Rate), IFO (Individual Fairness Outliers), *SPD*, and *EOD*, reporting noticeable improvements in model fairness. However, it is important to note that while `LIMI` contributes meaningfully to fairness testing and demonstrates that retraining on targeted samples can enhance fairness, its focus remains primarily on generating test inputs and probing latent spaces, rather than directly identifying or repairing specific fairness faults in the model's internal decision-making processes, as `FairFLRep` does. Moreover, `LIMI`'s reliance on surrogate boundary approximation may introduce variability, and its performance may fluctuate depending on data complexity and the quality of the underlying generative model, potentially limiting its robustness across different real-world scenarios.

`MetaRepair` [72] proposes a meta-learning framework that learns to repair DNNs by leveraging prior repair experiences. It trains a meta-repair model that generalizes across repair tasks and can quickly adapt to new models or faults using few-shot learning. Unlike `FairFLRep`, which directly targets fairness-related faults, `MetaRepair` focuses on general model errors and the repair strategy does not explicitly consider fairness. Moreover, `MetaRepair` requires a meta-training phase with access to multiple faulty models, where as `FairFLRep` operates directly on a given pretrained model without needing prior repair knowledge.

Jain et al. [52] propose a performance-aware fairness repair framework that uses Auto-Sklearn AutoML toolkit [24] to automatically search for fairness-preserving model modifications while maintaining high predictive performance. However, this approach is limited to classical machine learning models (and requires retraining or architectural changes) and is not applicable to DNNs.

Das et al. [17] introduce a multi-task CNN framework that jointly learns to classify gender, age, and ethnicity while mitigating inter-group bias. By exploiting correlations among these tasks and balancing feature learning across groups, the model improves fairness in visual attribute classification during model training.

Hendricks et al. [30] address bias in image captioning models, specifically gender bias arising from co-occurrence patterns in training data. It uses training and counterfactual data augmentation strategy to reduce bias in gender-specific captions without degrading overall caption quality. This work targets semantic bias in generative models, while `FairFLRep` repairs fairness in classification tasks.

Overall, `FairFLRep` stands out by offering a more integrated, flexible, and precise approach to fairness repair. It provides neuron-level analysis and repair without the need for retraining or external frameworks, making it more applicable in real-world scenarios where group fairness and model flexibility are key concerns.

## 4 Proposed Approach

### 4.1 Problem Definition

The goal of this study is to address fairness issues in a binary classification where the *sensitive attribute*, such as gender or race, is either the same as or different from the *target classification label*. The challenge is to mitigate biased behaviour in DNNs, particularly when the network disproportionately misclassifies or treats one group less favorably than another.

The bias bug being localized refers to the systematic unfair treatment or misclassification of certain groups, such as minority races, age, or genders. This bias often manifests as differences in misclassification rates or predictive performance between sensitive groups [5, 16]. For example, a model might misclassify individuals from minority groups at a higher rate than those from the majority group. Addressing these misclassification patterns, especially when they correlate with sensitive attributes, helps reduce bias by forcing the model to treat different groups more equitably [16, 38]. While maintaining accuracy is important, the primary goal of `FairFLRep` is to improve fairness by reducing disparities in predictive outcomes between groups. The objective is to develop a method that identifies and repairs the neuron weights responsible for this biased behavior while preserving the overall accuracy of the model $M$.

The proposed approach seeks to localize faulty neuron weights (those contributing to biased outcomes) using both gradient loss and forward impact metrics, prioritizing the weights that contribute the most to the unfair treatment of deprived groups. By targeting these biased weights, the goal is to improve fairness in the model while maintaining acceptable accuracy. Before formally stating the optimization problem, we review the key components of the proposed method.

### 4.2 Overview of `FairFLRep`

The overview of the proposed fairness-aware fault localization and repair approach, `FairFLRep`, depicted in Figure 1, aims to address bias bugs (unfairness) in DNN models. The approach operates by first identifying biased neuron weights responsible for unfair behavior and then repairing them to ensure fairness, while preserving the model's accuracy. `FairFLRep` operates as follows:

(1) ***During the fault localization step***, the process identifies the set of neuron weights that potentially lead to biased behavior in the model's decisions by performing gradient loss analysis and forward impact analysis on each data point for each sub-group in the repair dataset. Both correctly classified (positive) and misclassified (negative) inputs are analyzed, with a stronger emphasis on the negative inputs to correct biased behaviors while preserving correct classifications.

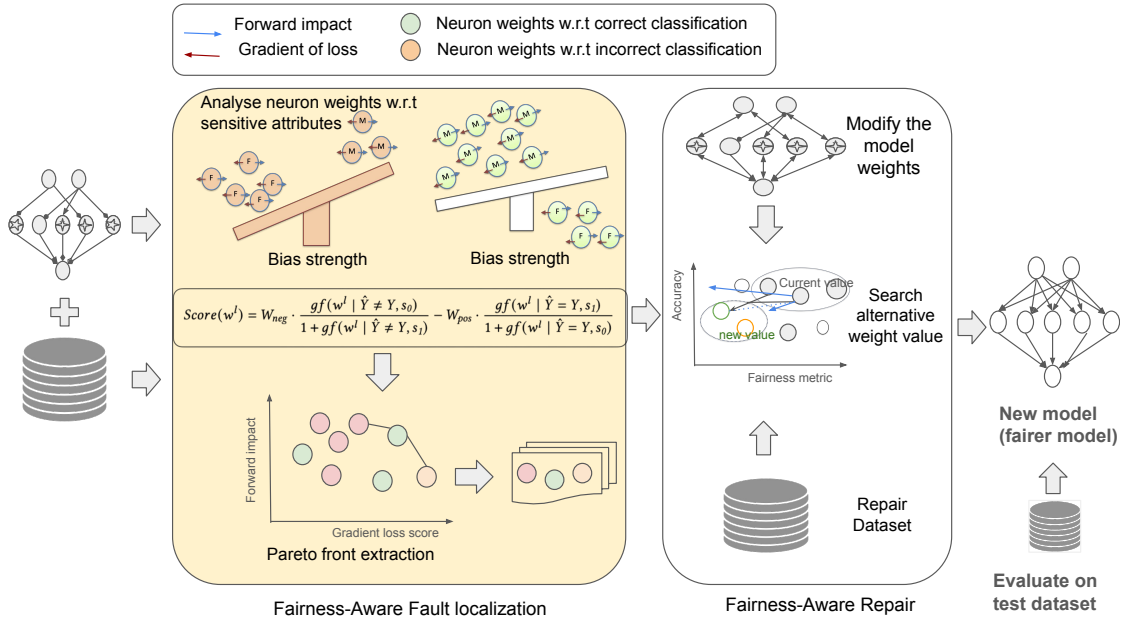

Fig. 1. Overview of `FairFLRep`

(2) **During the repair phase**, the approach searches for a set of neural weights that would correct the behavior of the DNN under repair based on Particle Swarm Optimization (PSO) [58]. PSO was selected for its effectiveness in optimization and its relatively fast convergence, essential in high-dimensional DNN weight adjustment tasks. During this search, a new model is defined by the search individual (i.e., an assignment to the search variables) that updates the original model with the new weights. A search individual is assessed using a fitness function that defines specific fairness metrics, such as *SPD*, *DI*, *FPR* and *EOD*, provided by the user.

*Rationale for Using PSO.* PSO is utilized in `FairFLRep` due to its advantages in continuous optimization and lower computational demands compared to some other evolutionary algorithms [63], making it a practical choice for fairness repairs where resources may be constrained. PSO's versatility across a range of optimization tasks also makes it effective for balancing fairness and accuracy during model repair. However, alternative evolutionary algorithms like Genetic Algorithms (GA) or Differential Evolution (DE) could also be suitable. For example, GA may offer robustness for tuning discrete parameters [78], while DE is well-suited for complex multimodal optimization landscapes due to its high convergence speed and accuracy [51, 56, 65]. This flexibility indicates that `FairFLRep`'s framework could accommodate various optimization algorithms, allowing adaptability based on specific application needs and resource limitations.

We detail the localization and repair phases in the following sections.

### 4.3 Proposed Fault Localization method

In this study, we build on the fault localization method proposed in [64], with a specific focus on fairness. We propose a novel fair-aware fault localization method designed to identify neural weights that are suspected of contributing to unfair behavior in the model. This allows for more effective repairs aimed at addressing unfairness.

DEFINITION 4.1. *(Fair-Aware Fault localization approach), denoted as* `FairFL`, *is an approach that identifies the neural weights at layer l in a DNN model $M$ that contribute to unfair behavior with respect to a sensitive attribute S. Given a repair dataset $\mathcal{D}rep$, the method combines gradient analysis and forward impact to localize these problematic weights and returns top–k weights $\mathbf{w_r}$ with the highest unfairness contributions.*

The fair-aware fault-localization steps proceed as follows:
- *Categorization of data*: The repair dataset $\mathcal{D}rep$ is divided into groups based on the correctness of the model's predictions and sensitive attribute values. The sensitive attribute $S$ takes on values in $\{s_0, s_1\}$, representing the deprived and favored community, respectively. For instance, if the sensitive attribute is gender, $s_0$ might correspond to female and $s_1$ to male. Data points are grouped accordingly based on the sensitive attribute. These subsets are used to examine unfair model predictions and the neural weights impacting such behavior.
- *Estimate the level of bias from input samples*: The method estimates the level of bias based on the prediction outcomes of model $M$ for different sensitive subgroups, such as the deprived and favored communities. This bias quantification is then used as a constraint when selecting neuron weights corresponding to model behaviors.
- *Neuron analysis*: It is performed with respect to the data samples that differ by the sensitive attribute $S$. The goal is to identify neuron weights that are highly influential for the deprived and favored sensitive attribute and assess their impact on classification decisions.
  Notably, for each weight connected to the last layer, we calculate the forward impact and gradient loss, respectively. The focus is on the last layer because it is most directly connected to classification, and this also reduces computational complexity, given that DNNs typically contain thousands of weight parameters. Prior work, such as adversarial debiasing techniques [68, 75], has demonstrated that directly applying mitigation techniques to the output layer promotes model fairness.
- *Scoring function*: A scoring function is employed to prioritize neural weights that maximize the gradient loss and forward impact for the deprived sensitive community relative to the favored community. This function ensures that neural weights contributing to misclassification, especially in the deprived community, are given higher priority. The rationale behind this approach is twofold: to address fairness issues by focusing on correcting biases affecting the deprived community, and to correct misclassifications without negatively impacting the correctly classified instances during the repair process.
- *Pareto front extraction*: The neuron weights $\mathbf{w_r}$ that form a Pareto front in terms of both high forward impact and high gradient loss are selected. A Pareto front consists of weights that cannot be improved in both metrics simultaneously. These weights are more likely to contribute to biased behavior, and improving them can lead to a reduction in bias without compromising model accuracy. Neurons on the Pareto front represent the most influential and biased neurons, making them ideal targets for repair.

Our fair-aware fault localization approach `FairFL` considers two forms of the sensitive attributes:
- `SettSame`: Attribute under prediction ($Y$) = Sensitive Attribute ($S$). This case applies when the sensitive attribute (e.g., gender) is the same as the class label.
- `SettDiff`: Class label ($Y$) ≠ Sensitive Attribute ($S$). This case applies when the sensitive attribute (e.g., race) differs from the class label (e.g., predicting gender but using race as the sensitive attribute).

In the next sections, we explain the details of the categorization, bias estimation, scoring, and repair processes, which together form the `FairFLRep` framework.

### 4.4 Categorization of data

The categorization step provides a framework for analyzing bias in models across two different cases where the attribute under prediction $Y$ and the sensitive attribute $S$ are the same or differ. This dual-categorization setup enables flexibility in the fault localization approach, allowing for more nuanced identification of unfair behavior of the model $M$ based on different relationships between the sensitive attribute and class labels. This flexible setup enables the fault localization and repair methods to be adaptable to different scenarios of bias, thus enhancing the fairness of the model across different sensitive attributes and class labels. Moreover, by isolating and analyzing targeted subgroups (e.g., misclassified minority samples), this categorization significantly reduces the search space during fault localization, thereby lowering the computational cost and improving the efficiency of the repair process.

*4.4.1 `SettSame`.* In this case, the sensitive attribute is the same as the class label. For example, if the model is predicting gender, the sensitive attribute would also be gender. The categorization is based on whether the data points are correctly or incorrectly classified:

- *Correctly Classified Samples by Sensitive Attribute*:
  - $\mathcal{D}rep_{pos,s_0} = \{(X, Y, S) \in \mathcal{D}rep_{pos} \mid S = s_0\}$: This set contains all correctly classified samples where the sensitive attribute $S = s_0$ (e.g., female).
  - $\mathcal{D}rep_{pos,s_1} = \{(X, Y, S) \in \mathcal{D}rep_{pos} \mid S = s_1\}$: This set contains all correctly classified samples where the sensitive attribute $S = s_1$ (e.g., male).
- *Misclassified Samples by Sensitive Attribute*:
  - $\mathcal{D}rep_{neg,s_0} = \{(X, Y, S) \in \mathcal{D}rep_{neg} \mid S = s_0\}$: This set contains all misclassified samples where the sensitive attribute $S = s_0$ (e.g., female).
  - $\mathcal{D}rep_{neg,s_1} = \{(X, Y, S) \in \mathcal{D}rep_{neg} \mid S = s_1\}$: This set contains all misclassified samples where the sensitive attribute $S = s_1$ (e.g., male).

This setup directly targets the bias in the sensitive attribute by grouping the data points into misclassified and correctly classified categories. The goal is to identify neuron weights that contribute to misclassification, particularly for instances where the model disproportionately misclassifies data points from certain sensitive groups (e.g., the minority group).

*4.4.2 `SettDiff`.* In this case, the sensitive attribute and class label are different. For example, the model might be predicting gender, but the sensitive attribute is race. The categorization is based on the class label $Y \in \{0, 1\}$, and sensitive attribute $S \in \{s_0, s_1\}$.

- Samples belonging to class $Y = 1$ (e.g., gender=male).
  - $\mathcal{D}rep_{1,s_0} = \{(X, Y, S) \in \mathcal{D}rep_{pos} \mid S = s_0, Y = 1\}$: instances where $Y = 1$ and the sensitive attribute $S = s_0$ (e.g., gender=male, race = minority).
  - $\mathcal{D}rep_{1,s_1} = \{(X, Y, S) \in \mathcal{D}rep_{pos} \mid S = s_1, Y = 1\}$: instances where $Y = 1$ and the sensitive attribute $S = s_1$ (e.g., gender=male, race = majority).
- Samples belonging to class $Y = 0$ (e.g., gender=female).
  - $\mathcal{D}rep_{0,s_0} = \{(X, Y, S) \in \mathcal{D}rep_{pos} \mid S = s_0, Y = 0\}$: instances where $Y = 0$ and the sensitive attribute $S = s_0$.
  - $\mathcal{D}rep_{0,s_1} = \{(X, Y, S) \in \mathcal{D}rep_{pos} \mid S = s_1, Y = 0\}$: instances where $Y = 0$ and the sensitive attribute $S = s_1$.

## 4.5 Estimating the level of bias

*4.5.1* `SettSame` *– Estimating the level of bias.* We compute the level of bias in both the correctly classified instances and the misclassified instances with respect to the sensitive attribute $S \in \{s_0, s_1\}$. This involves estimating how classifications are biased toward or against specific sensitive groups.

The weighing scores add constraints to ensure neurons are selected fairly, preventing overcompensation toward a specific sensitive group.

- **Estimating the level of bias** $W_{neg}$ **in misclassified data** $\mathcal{D}rep_{neg}$

We aim to estimate the level of bias in the misclassified data points produced by model $\mathcal{M}$ for different sub-groups, such as the deprived community $s_0$ and the favored community $s_1$. This bias quantification can then be used as a constraint when selecting the neuron weights corresponding to misclassification behaviors.

— *Expected probability of bias-free misclassification*: If the model $\mathcal{M}$ were unbiased with respect to the dataset (i.e., $S$ is statistically independent of misclassification), the expected probability $\mathcal{P}_{exp}$ of misclassifying data points with sensitive attribute $S = s_0$ would be:

$$\mathcal{P}_{exp}(\mathcal{D}rep_{neg,s_0}) := \frac{|\{(X,Y,S) \in \mathcal{D}rep \mid S = s_0\}|}{|\mathcal{D}rep|} \times \frac{|\mathcal{D}rep_{neg}|}{|\mathcal{D}rep|} \quad (1)$$

This term computes the expected proportion of deprived community members who would be misclassified by an unbiased model $\mathcal{M}$. Essentially, it calculates the fraction of deprived samples in the entire dataset and scales this by the overall misclassification rate.

— *Observed probability of misclassification*: In reality, the observed probability $\mathcal{P}_{obs}$ of misclassifying data points with the sensitive attribute $S = s_0$ may differ from the expected probability. The observed probability $\mathcal{P}_{obs}$ is defined as:

$$\mathcal{P}_{obs}(\mathcal{D}rep_{neg,s_0}) := \frac{|\mathcal{D}rep_{neg,s_0}|}{|\mathcal{D}rep|} \quad (2)$$

i.e., the proportion of misclassified instances in the dataset that belong to the deprived community.

— *Cost of misclassified instances*: If the observed probability $\mathcal{P}_{obs}(\mathcal{D}rep_{neg,s_0})$ is lower than the expected probability $\mathcal{P}_{exp}(\mathcal{D}rep_{neg,s_0})$, $\mathcal{M}$ exhibits bias against the deprived community. To quantify this bias, we assign the following cost to the neuron weights when the input is the misclassified instances from the deprived community:

$$cost(\mathcal{D}rep_{neg,s_0}) := \frac{\mathcal{P}_{exp}(\mathcal{D}rep_{neg,s_0})}{\mathcal{P}_{obs}(\mathcal{D}rep_{neg,s_0})} \quad (3)$$

This cost represents the level of bias correction needed to ensure the deprived group is not unfairly penalized in the model's misclassifications. The cost scales inversely with the observed probability—if the deprived group is misclassified less frequently than expected, this cost will increase.

Similarly, the cost for misclassified instances from the favored community $S = s_1$ is:

$$cost(\mathcal{D}rep_{neg,s_1}) := \frac{\mathcal{P}_{exp}(\mathcal{D}rep_{neg,s_1})}{\mathcal{P}_{obs}(\mathcal{D}rep_{neg,s_1})} \quad (4)$$

— *Level of bias in misclassified instances*: The strength of bias in the misclassified instances $\mathcal{D}rep_{neg}$ can be quantified by comparing the bias costs between the deprived and favored groups. This is expressed as the ratio of the cost for the deprived community (Eq. 3) to the cost for the favored community (Eq. 4):

$$W_{neg} := \frac{cost(\mathcal{D}rep_{neg,s_1})}{cost(\mathcal{D}rep_{neg,s_0})} \quad (5)$$

The ratio $W_{neg}$ quantifies the level of bias between the deprived and favored groups in the misclassified data points. If $W_{neg} > 1$, the model is more biased against the deprived group (i.e., their misclassification cost is higher than expected), and if $W_{neg} < 1$, the model is more biased against the favored group. $W_{neg}$ can be used as a constraint or a guiding metric when selecting neuron weights corresponding to deprived relative to favored group.

• **Estimating the level of bias $W_{pos}$ in correctly classified data $\mathcal{D}rep_{pos}$**

Given the correctly classified data points, we estimate the level of bias represented by each sub-group, $s_0$ (deprived group) and $s_1$ (favored group), by calculating the cost per group following the same steps as in the previous analysis (Section 4.5.1).

— *Cost for the deprived community $s_0$:* The cost for the deprived community $s_0$ in the correctly classified set $\mathcal{D}rep_{pos}$ is calculated using the following formula:

$$cost(\mathcal{D}rep_{pos,s_0}) := \frac{\mathcal{P}_{exp}(\mathcal{D}rep_{pos,s_0})}{\mathcal{P}_{obs}(\mathcal{D}rep_{pos,s_0})} \quad (6)$$

This formula accounts for the proportion of the dataset that belongs to the deprived community and compares that to the proportion of correctly classified instances from the deprived community. This cost reflects how well the deprived group is represented in the correctly classified set relative to its overall presence in the dataset.

— *Cost for the Favored Community $s_1$:* Similarly, the cost for the favored community $s_1$ in the correctly classified set $\mathcal{D}rep_{pos}$ is calculated as:

$$cost(\mathcal{D}rep_{pos,s_1}) := \frac{\mathcal{P}_{exp}(\mathcal{D}rep_{pos,s_1})}{\mathcal{P}_{obs}(\mathcal{D}rep_{pos,s_1})} \quad (7)$$

This formula mirrors the calculation for the deprived community, measuring the bias relative to the correctly classified instances for the favored group.

— *Level of bias in correctly classified instances*: Once the costs for both the deprived and favored communities are computed, the strength of bias in the correctly classified instances $\mathcal{D}rep_{pos}$ is given by the ratio of the cost for the deprived community to the cost for the favored community:

$$W_{pos} := \frac{cost(\mathcal{D}rep_{pos,s_0})}{cost(\mathcal{D}rep_{pos,s_1})} \quad (8)$$

This ratio $W_{pos}$ quantifies the relative bias between the deprived and favored groups among the correctly classified instances. A higher value of $W_{pos}$ indicates a greater level of bias on the deprived community for the correct resultsIn this case: (i) $W_{pos} > 1$ indicates more biased toward favored group; this means the deprived community is likely to be least classified correctly by the model $\mathcal{M}$ than expected. (ii) $W_{pos} < 1$ indicates fairness on the favored group; this indicates the favored community is more likely to be correctly classified by the model $\mathcal{M}$ as expected, given their presence in the dataset.

This measure of bias in the correctly classified instances complements the bias measure computed in the misclassified instances $\mathcal{D}rep_{neg}$, providing a more holistic understanding of how the model behaves across different sensitive groups. The value of bias strength $W_{pos}$ can

Table 1. Examples for SettSame and SettDiff.

(a) SettSame

| # | Y | S | Correct | Miss |
|---|---|---|---------|------|
| 1 | 1 | male(1) | Yes | No |
| 2 | 1 | male(1) | Yes | No |
| 3 | 1 | male(1) | Yes | No |
| 4 | 0 | female(0) | No | Yes |
| 5 | 0 | female(0) | No | Yes |
| 6 | 1 | male(1) | No | Yes |
| 7 | 0 | female(0) | Yes | No |
| 8 | 0 | female(0) | No | Yes |
| 9 | 1 | male(1) | Yes | No |
| 10 | 0 | female(0) | Yes | No |

(b) SettDiff

| # | Y | S | Correct | Miss |
|---|---|---|---------|------|
| 1 | 1 | white | Yes | No |
| 2 | 0 | white | No | Yes |
| 3 | 0 | black | Yes | No |
| 4 | 1 | white | No | Yes |
| 5 | 0 | black | No | Yes |
| 6 | 1 | white | No | Yes |
| 7 | 0 | white | Yes | No |
| 8 | 1 | black | Yes | No |
| 9 | 1 | white | Yes | No |
| 10 | 0 | black | Yes | No |

then be used to guide the selection of neuron weights from the correctly classified inputs with respect to sensitive attribute $S \in \{s_0, s_1\}$

*Example 4.2.* To illustrate how the bias estimation works, we apply the above formulas to the simple repair dataset $\mathcal{D}rep$ shown in Table 1a. In this example, the sensitive attribute is *gender* and the class label is also *gender*; we consider 10 samples to illustrate the application of the formulas. So, according to the table, the sizes of the subsets of $\mathcal{D}rep$ are as follows:

- Total samples. $|\mathcal{D}rep| = 10$;
- Female samples. $|\{(X, Y, S) \in \mathcal{D}rep \mid S = s_0\}| = 5$;
- Misclassified samples. $|\mathcal{D}rep_{neg}| = 4$;
- Misclassified female samples. $|\mathcal{D}rep_{neg,s_0}| = 3$;
- Misclassified male samples. $|\mathcal{D}rep_{neg,s_1}| = 1$;
- Correctly classified samples. $|\mathcal{D}rep_{pos}| = 6$;
- Correctly classified female samples. $|\mathcal{D}rep_{pos,s_0}| = 2$;
- Correctly classified male samples. $|\mathcal{D}rep_{pos,s_1}| = 4$.

Given these values, the levels of bias are computed as follows:

- Bias in Misclassified Data:

$$\mathcal{P}_{exp}(\mathcal{D}rep_{neg,s_0}) := \frac{5}{10} \times \frac{4}{10} = 0.2 \qquad \mathcal{P}_{obs}(\mathcal{D}rep_{neg,s_0}) := \frac{3}{10} = 0.3$$

$$cost(\mathcal{D}rep_{neg,s_0}) := \frac{0.2}{0.3} \approx 0.67 \qquad cost(\mathcal{D}rep_{neg,s_1}) := \frac{0.2}{0.1} = 2.0$$

$$W_{neg} := \frac{2.0}{0.67} \approx 2.99$$

- Bias in Correctly Classified Data:

$$cost(\mathcal{D}rep_{pos,s_1}) := \frac{0.3}{0.2} = 1.5 \qquad cost(\mathcal{D}rep_{pos,s_1}) := \frac{0.3}{0.4} = 0.75$$

$$W_{pos} := \frac{1.5}{0.75} = 2.0$$

*4.5.2* `SettDiff` *– Estimating the level of bias.* This section outlines the procedure for estimating bias in `SettDiff`, where the sensitive attribute $S$ differs from the class label $Y$. The bias estimation follows a similar approach as previously described but focuses on class-sensitive attribute combinations.

- **Bias in the class label**

  We first estimate bias in predicting class labels $Y = 0$ and $Y = 1$, as follows:
  
  — ***Expected probability of bias-free prediction for*** $Y = 0$:

  $$\mathcal{P}_{exp}(Y = 0) := \frac{|\{(X, Y, S) \in \mathcal{D}rep \mid Y = 0\}|}{|\mathcal{D}rep|} \times \frac{|\mathcal{D}rep_{pos}|}{|\mathcal{D}rep|} \qquad (9)$$

  This term computes the expected proportion of correctly classified instances for $Y = 0$.

  — ***Observed probability within the class*** $Y = 0$:

  $$\mathcal{P}_{obs}(Y = 0) := \frac{|\{(X, Y, S) \in \mathcal{D}rep_{pos} \mid Y = 0\}|}{|\mathcal{D}rep|} \qquad (10)$$

  This reflects the actual classification rate of the model $\mathcal{M}$ for instances with $Y = 0$.

  — ***Estimating the cost of classification for class label*** $Y = 0$:

  $$cost(Y = 0) := \frac{\mathcal{P}_{exp}(Y = 0)}{\mathcal{P}_{obs}(Y = 0)} \qquad (11)$$

  This cost quantifies how much correction is needed for the model to achieve bias-free classification of $Y = 0$.

  Similarly, the classification cost for the class label $Y = 1$ is calculated as:

  $$cost(Y = 1) := \frac{\mathcal{P}_{exp}(Y = 1)}{\mathcal{P}_{obs}(Y = 1)} \qquad (12)$$

  — ***The level of bias in class label***

    The strength of bias reflected in the two class labels can be quantified by comparing the bias costs between the $Y = 0$ and $Y = 1$: if $cost(Y = 0) > cost(Y = 1)$, the model is biased towards $Y = 0$, otherwise, it is biased towards $Y = 1$. This information will guide the scoring function in prioritizing class labels to achieve equitable outcomes across class-sensitive attribute combinations.

- **Bias strength for deprived community**

  We aim to estimate the level of bias for the subset of data where $Y = 0$ and the sensitive attribute is from deprived community, $S = s_0$.

  — ***Expected probability of deprived community*** $s_0$ ***in*** $Y = 0$:

  $$\mathcal{P}_{exp}(\mathcal{D}rep_{0,s_0}) := \frac{|\{(X, Y, S) \in \mathcal{D}rep \mid S = s_0\}|}{|\mathcal{D}rep|} \times \frac{|\{(X, Y, S) \in \mathcal{D}rep \mid Y = 0\}|}{|\mathcal{D}rep|} \qquad (13)$$

  — ***Observed probability of deprived community*** $s_0$ ***within the class*** $Y = 0$:

  $$\mathcal{P}_{obs}(\mathcal{D}rep_{0,s_0}) := \frac{|\mathcal{D}rep_{0,s_0}|}{|\mathcal{D}rep|} \qquad (14)$$

  — ***Estimating the cost for deprived community in*** $Y = 0$:

  $$cost(\mathcal{D}rep_{0,s_0}) := \frac{\mathcal{P}_{exp}(\mathcal{D}rep_{0,s_0})}{\mathcal{P}_{obs}(\mathcal{D}rep_{0,s_0})} \qquad (15)$$

Similarly, the cost for the favored community $s_1$ in $Y = 0$ is calculated as:

$$cost(\mathcal{D}rep_{0,s_1}) := \frac{\mathcal{P}_{exp}(\mathcal{D}rep_{0,s_1})}{\mathcal{P}_{obs}(\mathcal{D}rep_{0,s_1})} \tag{16}$$

— **Computing the level of bias in class $Y = 0$:**

$$W_0 := \frac{cost(\mathcal{D}rep_{0,s_0})}{cost(\mathcal{D}rep_{0,s_1})} \tag{17}$$

This ratio $W_0$ quantifies the level of bias between the deprived and favored groups in the instance of dataset for class labeled category $Y = 0$.

• **Estimating the level of bias within class $Y = 1$**

The bias for $Y = 1$ is computed similarly to $Y = 0$, using the same methodology to assess bias between the deprived and favored communities within $Y = 1$.

— Cost for the deprived community $s_0$ in class $Y = 1$:

$$cost(\mathcal{D}rep_{1,s_0}) := \frac{\mathcal{P}_{exp}(\mathcal{D}rep_{1,s_0})}{\mathcal{P}_{obs}(\mathcal{D}rep_{1,s_0})}$$

and similarly for favored community as:

$$cost(\mathcal{D}rep_{1,s_1}) := \frac{\mathcal{P}_{exp}(\mathcal{D}rep_{1,s_1})}{\mathcal{P}_{obs}(\mathcal{D}rep_{1,s_1})}$$

— **Computing the level of bias in $Y = 1$:**

$$W_1 := \frac{cost(\mathcal{D}rep_{1,s_0})}{cost(\mathcal{D}rep_{1,s_1})} \tag{18}$$

This framework estimates the bias strength for both $Y = 0$ and $Y = 1$, offering a comprehensive view of how the model behaves with respect to different sensitive attributes in each class. These bias measures can then guide the selection of neuron weights to be adjusted in the repair process to ensure equitable treatment of both deprived and favored communities.

*Example 4.3.* Consider the scenario in Table 1b, where we are predicting gender ($Y \in \{0, 1\}$, where 1 = male, 0 = female), and the sensitive attribute $S \in \{s_0, s_1\}$ represents race ($s_0$ = black, $s_1$ = white). In this case, $Y$ and $S$ are not the same, and gender prediction should be independent of race. This fits the `SettDiff` condition.

We compute the *Bias in Class Label* as follows:

• For $Y = 0$

$$cost(Y = 0) := \frac{\frac{5}{10} \times \frac{5}{10}}{\frac{2}{10}} = \frac{0.25}{0.2} = 1.25$$

• For $Y = 1$

$$cost(Y = 1) := \frac{\frac{5}{10} \times \frac{5}{10}}{\frac{3}{10}} = \frac{0.25}{0.3} \approx 0.83$$

• Bias in class $Y = 0$

$$\mathcal{P}_{exp}(\mathcal{D}rep_{0,black}) := \frac{4}{10} \times \frac{5}{10} = 0.20 \qquad \mathcal{P}_{exp}(\mathcal{D}rep_{0,white}) := \frac{6}{10} \times \frac{5}{10} = 0.30$$

$$\mathcal{P}_{obs}(\mathcal{D}rep_{0,black}) := \frac{2}{10} = 0.2 \qquad \mathcal{P}_{obs}(\mathcal{D}rep_{0,white}) := \frac{1}{10} = 0.1$$

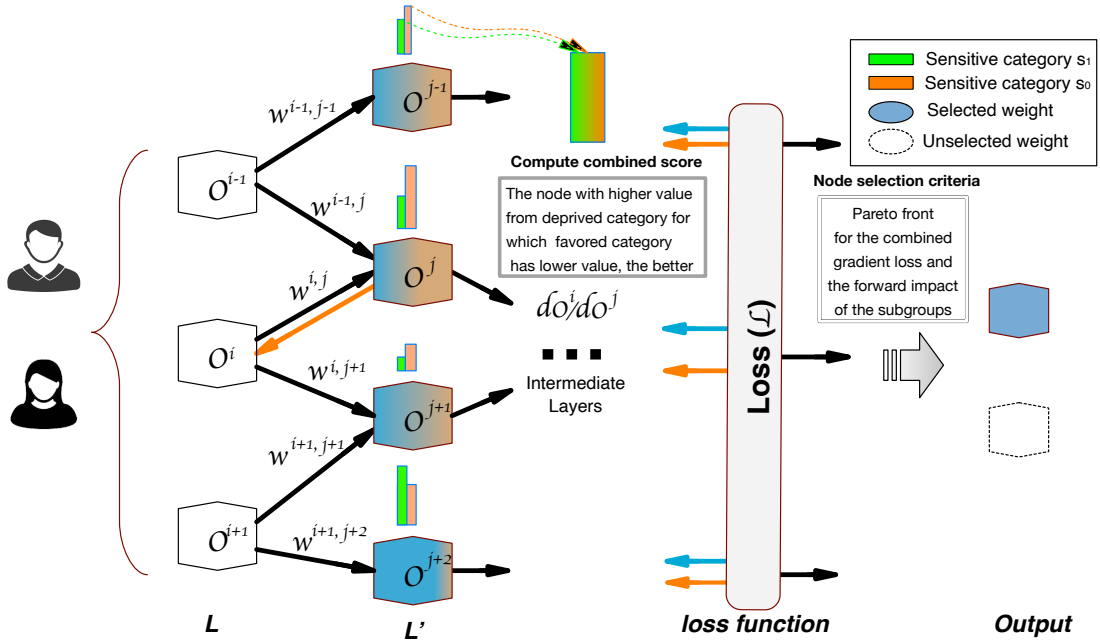

Fig. 2. Propose `FairFL` fault localization technique. We first compute the individual gradient loss and forward impact of each individual weight for sub-groups within a selected layer. Based on the Pareto front, we then select the weights from the group with combined higher gradient and forward impact. In the figure, black arrows represent forward pass, whereas blue and orange arrows are backpropagation. Black and blue arrows combined compute the forward impact, whereas the orange arrow denotes the gradient loss. The gradient loss and the forward impact combined evaluate the likelihood of $w^{i,j}$ being localized by our technique.

$$W_0 := \frac{\frac{0.20}{0.20}}{\frac{0.30}{0.1}} = \frac{1.0}{3.0} \approx 0.333$$

- Bias in class $Y = 1$

$$\mathcal{P}_{exp}(\mathcal{D}rep_{1,black}) := \frac{1}{10} \times \frac{5}{10} = 0.05 \qquad \mathcal{P}_{exp}(\mathcal{D}rep_{1,white}) := \frac{4}{10} \times \frac{5}{10} = 0.2$$

$$\mathcal{P}_{obs}(\mathcal{D}rep_{1,black}) := \frac{1}{10} = 0.1 \qquad \mathcal{P}_{obs}(\mathcal{D}rep_{1,white}) := \frac{2}{10} = 0.2$$

$$W_1 := \frac{\frac{0.05}{0.1}}{\frac{0.2}{0.2}} = 0.5$$

### 4.6 Neuron Analysis with respect to data samples

As depicted in Figure 2, we conduct neuron analysis with respect to the data samples that differ by the sensitive attribute $S \in \{s_0, s_1\}$. The goal is to understand how the weights of individual neurons contribute to the model's incorrect behavior (i.e., unfair predictions). To achieve this, we calculate both the *forward impact* and *gradient loss* of each neuron weight in a given layer of the neural network.

*4.6.1 Gradient Loss.* The gradient loss measures the responsibility that a particular neural weight has for the misclassification within a specific sensitive attribute group (e.g., the deprived group). A higher gradient loss indicates that the weight is more strongly associated with causing misclassification errors. The intuition is that, by fixing the misclassification with respect to such neurons, we have the potential of improving the unfair treatment for the different sub-groups of sensitive attribute.

Given an input sample from a deprived subset (e.g., $\mathcal{D}rep_{neg,s_0}$), we compute the gradient loss with respect to each weight $w^l$ in layer $l$ (typically, the last layer is used in this study) of the neural network by taking the gradient of the loss function $\mathcal{L}$ with respect to these weights. We introduce the following notation:

- $\mathcal{L}_{neg,s_0} = \mathcal{L}(\hat{Y}_{neg,s_0}, Y_{neg,s_0})$: The loss (e.g., cross-entropy) between the predicted output $\hat{Y}_{neg,s_0}$ and the true label $Y_{neg,s_0}$, for the input sample from the deprived community $\mathcal{D}rep_{neg,s_0}$.
- $w^l \in W^l$: The individual weights of the layer $l$ of the neural network $\mathcal{M}$.
- $z^l$: The activations (pre-activation outputs) of the layer $l$ before applying the output activation function (e.g., softmax).

The gradient loss with respect to the neuron weight $w^l$ in layer $l$ for a given input sample $\mathcal{D}rep_{neg,s_0}$ is calculated as:

$$grad\_loss_{w^l(\mathcal{D}rep_{neg,s_0})} = \frac{\delta \mathcal{L}_{neg,s_0}}{\delta w^l} = \frac{\delta \mathcal{L}_{neg,s_0}}{\delta z^l} \cdot \frac{\delta z^l}{\delta w^l}$$

where:

- $\frac{\delta \mathcal{L}_{neg,s_0}}{\delta w^l}$ is the gradient of the loss function with respect to the pre-activation outputs $z^l$. This quantifies how changes in the activations of layer $l$ affect the loss for the missclasified instance of deprived sample.
- $\frac{\delta z^l}{\delta w^l}$ is the gradient of the pre-activation outputs with respect to the weights in layer $l$. This quantifies how changes in the weights affect the activations in layer $l$.

A high value of the gradient loss indicates that the weight $w^l$ is contributing significantly to the incorrect behavior (i.e., misclassifications) of the input from the deprived group. Adjusting such weights can help reduce unfair misclassifications, thereby improving fairness for the sensitive attribute group.

Similarly, the gradient loss of each neuron weights $w^l$ in layer $l$ is computed using the same methodology for other subsets of data. For an input sample from the favored community $\mathcal{D}rep_{neg,s_1}$, the gradient loss is calculated as:

$$grad\_loss_{w^l(\mathcal{D}rep_{neg,s_1})} = \frac{\delta \mathcal{L}_{neg,s_1}}{\delta z^l} \cdot \frac{\delta z^l}{\delta w^l} = \frac{\delta \mathcal{L}_{neg,s_1}}{\delta w^l}$$

*4.6.2 Forward impact analysis.* The forward impact measures the influence of a neural weight on the final classification decision. This metric evaluates how strongly each weight affects the prediction outcome, regardless of whether the prediction is correct or incorrect. Similar to gradient loss, forward impact is calculated for each neural weight with respect to different sensitive groups (i.e., $\mathcal{D}rep_{s_0}$, $\mathcal{D}rep_{s_1}$).

Estimating the forward impact of a neural weight $w^l$ on the final output is more complex than computing the gradient loss. Neural weights that strongly influence the activation value of the connected output neuron, where the output neuron subsequently influences the final classification result, are considered to have a significant impact on the model's predictions. We compute forward impact as follows:

(1) We begin by estimating the influence of a neural weight on the activation of the output neuron, denoted as $o^l w^l$. This is done by multiplying the neural weight $w^l$ by the average activation value $z_\mu^{l-1}$ of the neuron it is connected to in the previous layer, resulting in:

$$o^l w^l = w^l * z_\mu^{l-1} + b^l$$

where $b^l$ is the bias vector term on the output neuron.

The computed influence is then normalized by dividing it by the total influence of all neural weights connected to the same output neuron:

$$\text{Normalized influence} = \frac{o^l w^l}{\sum_i^{|L|} o^l w^l}$$

where $|L|$ is the total number of weights connected to the output neuron.

This normalization step ensures that the forward impact is measured as a proportion of the total influence on the output neuron.

(2) Next, we calculate the influence of the output neuron's activation on the final output $O$. This is done by computing the gradient of the final output $O$ with respect to the activation of the output neuron $o'_j$.

$$\text{Influence on final output} = \frac{\delta O}{\delta o'_j}$$

This gradient captures how much the final output changes in response to changes in the activation of the output neuron.

Finally, we compute the forward impact ($fwd\_impact_{w^l}$) of a neural weight $w^l$ as:

$$fwd\_impact_{w^l} = \frac{o^l w^l}{\sum_i^{|L|} o^l w^l} \times \frac{\delta O}{\delta o'_j} \qquad (19)$$

This provides a comprehensive measure of the impact the weight has on the final classification decision.

As with the gradient loss, we estimate the forward impact of the neural weights with respect to different sub-groups of sensitive attributes. For example:

- $fwd\_impact_{w^l(\mathcal{D}rep_{neg,s_0})}$: Forward impact of neuron weight $w^l$ when the input is the misclassified samples from the deprived community $\mathcal{D}rep_{neg,s_0}$.
- $fwd\_impact_{w^l(\mathcal{D}rep_{pos,s_1})}$: Forward impact of neuron weight $w^l$ when the input is the correctly classified samples from the favored community $\mathcal{D}rep_{pos,s_1}$.

### 4.7 Scoring of neuron weights

We aggregate the values of the forward impact independently of the gradient loss based on a scoring function that identifies faulty neurons contributing to the biased behavior. The information from both the gradient loss and forward impact of each neuron weight allows us to treat these as two competing objectives to optimize. Neurons with higher bias impact on the deprived community are prioritized for repair. The bias strength scores (See Sections 4.5.1 and 4.5.2) help control how much each neuron is selected for adjustment, ensuring that the selected weights is proportional to the bias reflected in $\mathcal{M}$.

As pointed out in Section 4.4, the dual categorization of the data are based on the version of fault-location technique: `SettSame` when $Y = S$, and `SettDiff` when $Y \neq S$. Similarly, the scoring function differs slightly depending on the version of the fault-location technique used.

4.7.1 **SettSame** *(fair-aware fault localization when Y = S)*. The scoring function focuses on selecting neuron weights contributing more to misclassifications from the deprived community relative to those contributing to correct classifications from the favored community. The goal is to prioritize the neuron weights that disproportionately impact deprived groups, relative to the favored groups. Thus, for each weight $w^l$, we compute:

$$score_{w^l} = W_{neg} \cdot \frac{gf(w^l \mid \mathcal{D}rep_{neg,s_0})}{1 + gf(w^l \mid \mathcal{D}rep_{neg,s_1})} - W_{pos} \cdot \frac{gf(w^l \mid \mathcal{D}rep_{pos,s_0})}{1 + gf(w^l \mid \mathcal{D}rep_{pos,s_1})}, \qquad (20)$$

where:
- $w^l$ is the specific weight being scored
- $W_{neg}$ and $W_{pos}$ are the level of bias in misclassified and correctly classified instances, computed based on Eqs. 5 and 8,
- $gf$ is either the measure of gradient of loss or forward impact of weight $w^l$ with respect to the input samples.

The numerator in these equations captures the gradient loss and forward impact on the deprived community, while the denominator captures the impact on the favored community. Neuron weights with higher gradient loss and forward impact on the deprived community are maximized hence are more likely to be selected for repair, while neurons with high impact on the favored community are minimize hence scored low.

The pseudo-code of the FairFL method, given $Y = S$, is presented in Algorithm 1. The algorithm works as follows.
(1) It starts by randomly sampling the same number of positive inputs as negative ones since, typically, a fully trained model has far more positive inputs than negative ones (Line 2).
(2) It computes the bias cost (both positive and negative) for each sensitive group (Lines 5-6).
(3) For each neuron weight $w^l$, it computes the gradient loss with respect to the deprived and favored groups for both misclassified and correctly classified instances, according to procedure in Section 4.6. This is done by calling the function *Compute_Gradient_Loss* (Lines 11-14).
(4) Similarly, for each neuron weight $w^l$, it computes the forward impact with respect to the deprived and favored groups for both misclassified and correctly classified instances, using the *Compute_Forward_Impact* function (Lines 16-19).
(5) The values of the gradient loss and forward impact are aggregated to create a combined score for each neuron weight in Line 15 and Line 20, respectively. This aggregation gives higher priority to neuron weights that contribute more to negative inputs, with a specific focus on the deprived community relative to the favored group.
(6) The algorithm identifies the set of neuron weights $\mathbf{w_r}$ that constitute the Pareto front based on the combined scores for gradient loss and forward impact (Line 23). The Pareto front consists of neuron weights for which no other weight has both a greater gradient loss and greater forward impact simultaneously. These neurons are the most impactful for repair, as improving them will likely lead to the greatest reduction in bias without negatively affecting model accuracy.

4.7.2 **SettDiff** *(fair-aware fault localization when Y ≠ S)*. Similarly to SettSame, $W_0$ measures bias in the underrepresented class label (Eq. 17), and $W_1$ measures bias in the more represented class label (Eq. 18). The neuron scoring function prioritizes underrepresented groups based on their class labels and sensitive attributes. The scoring function for SettDiff balances the need to address bias in the underrepresented class while maintaining fairness in the more represented class. The goal is to prioritize neuron weights contributing to biased outcomes for underrepresented classes while ensuring that the adjustment does not unfairly penalize the more represented class.

**Algorithm 1** FairFL fault localization when $Y = S$

---

**INPUT:** A DNN model to be repaired $\mathcal{M}$; set of inputs correctly and incorrectly classified data instance $\{\mathcal{D}rep_{pos}, \mathcal{D}rep_{neg}\} \in \mathcal{D}rep$; the sensitive attributes $S \in \{s_0, s_1\}$; the target layer $l$.
**OUTPUT:** a set of neural weights to target for repair, $\mathbf{w_r}$

1: **procedure** FairFL
2:     $\mathcal{D}rep_{pos} \leftarrow$ RandomSample($\mathcal{D}rep_{pos}$, sizeof=$|\mathcal{D}rep_{neg}|$)
3:     $\{\mathcal{D}rep_{pos,s_0}, \mathcal{D}rep_{pos,s_1}\} \leftarrow$ Inputs belonging to deprived and favored group in $\mathcal{D}rep_{pos}$
4:     $\{\mathcal{D}rep_{neg,s_0}, \mathcal{D}rep_{neg,s_1}\} \leftarrow$ Inputs belonging to deprived and favored group in $\mathcal{D}rep_{neg}$
5:     $W_{neg} \leftarrow$ compute_cost_of_bias($\mathcal{D}rep_{neg}, s_0, s_1$)
6:     $W_{pos} \leftarrow$ compute_cost_of_bias($\mathcal{D}rep_{pos}, s_0, s_1$)
7:     $scoredWeights \leftarrow [\,]$
8:     $W^l \leftarrow$ Set of weights in the target layer $l$
9:     $\mathcal{L} \leftarrow$ a loss function used in $\mathcal{M}$
10:     **for** $w^l \in W^l$ **do**
11:        $grad\_loss_{w^l(\mathcal{D}rep_{neg,s_0})} \leftarrow$ Compute_Gradient_Loss($w^l, \mathcal{M}, \mathcal{D}rep_{neg,s_0}, \mathcal{L}$)
12:        $grad\_loss_{w^l(\mathcal{D}rep_{neg,s_1})} \leftarrow$ Compute_Gradient_Loss($w^l, \mathcal{M}, \mathcal{D}rep_{neg,s_1}, \mathcal{L}$)
13:        $grad\_loss_{w^l(\mathcal{D}rep_{pos,s_0})} \leftarrow$ Compute_Gradient_Loss($w^l, \mathcal{M}, \mathcal{D}rep_{pos,s_0}, \mathcal{L}$)
14:        $grad\_loss_{w^l(\mathcal{D}rep_{pos,s_1})} \leftarrow$ Compute_Gradient_Loss($w^l, \mathcal{M}, \mathcal{D}rep_{pos,s_1}, \mathcal{L}$)
15:        $grad\_lossScore_{w^l} \leftarrow W_{neg} * \frac{grad\_loss_{w^l(\mathcal{D}rep_{neg,s_0})}}{1+grad\_loss_{w^l(\mathcal{D}rep_{neg,s_1})}} - W_{pos} * \frac{grad\_loss_{w^l(\mathcal{D}rep_{pos,s_0})}}{1+grad\_loss_{w^l(\mathcal{D}rep_{pos,s_1})}}$
16:        $fwd\_impact_{w^l(\mathcal{D}rep_{neg,s_0})} \leftarrow$ Compute_Forward_Impact($w^l, \mathcal{M}, \mathcal{D}rep_{neg,s_0}$)
17:        $fwd\_impact_{w^l(\mathcal{D}rep_{neg,s_1})} \leftarrow$ Compute_Forward_Impact($w^l, \mathcal{M}, \mathcal{D}rep_{neg,s_1}$)
18:        $fwd\_impact_{w^l(\mathcal{D}rep_{pos,s_0})} \leftarrow$ Compute_Forward_Impact($w^l, \mathcal{M}, \mathcal{D}rep_{pos,s_0}$)
19:        $fwd\_impact_{w^l(\mathcal{D}rep_{pos,s_1})} \leftarrow$ Compute_Forward_Impact($w^l, \mathcal{M}, \mathcal{D}rep_{pos,s_1}$)
20:        $fwd\_impact_{w^l} \leftarrow W_{neg} * \frac{fwd\_impact_{w^l(\mathcal{D}rep_{neg,s_0})}}{1+fwd\_impact_{w^l(\mathcal{D}rep_{neg,s_1})}} - W_{pos} * \frac{fwd\_impact_{w^l(\mathcal{D}rep_{pos,s_1})}}{1+fwd\_impact_{w^l(\mathcal{D}rep_{pos,s_1})}}$
21:        **Add tuple** ($w^l, grad\_loss, fwd\_impact$) **to** $scoredWeights$
22:     **end for**
23:     $\mathbf{w_r} \leftarrow$ ExtractParetoFront($scoredWeights$)
24:     **Return** $\mathbf{w_r}$
25: **end procedure**

---

Algorithm 2 presents the pseudo-code of the FairFL method given $Y \neq S$.
(1) It starts by randomly sampling equal numbers of instances from the less represented class (e.g., $Y = 0$) and the more represented class (e.g., $Y = 1$), ensuring balanced input data for the scoring process (Line 2). The algorithm assumes that the cost of misclassification for $Y = 0$ is greater than that for $Y = 1$, meaning that the underrepresented group ($Y = 0$) faces more bias.
(2) It computes the bias cost (both $Y = 1$ and $Y = 0$) for each sensitive group (Lines 5-6).
(3) For each neuron weight $w$, it computes the gradient loss with respect to the deprived and favored groups for classes, according to procedure in Section 4.6. This is done by calling the function *Compute_Gradient_Loss* (Lines 11-14).
(4) Similarly, for each neuron weight $w$, it computes the forward impact with respect to the deprived and favored groups for both classes, using the *Compute_Forward_Impact* function (Lines 16-19).

---

**Algorithm 2** FairFLRep fault localization when $Y \neq S$

---

**INPUT:** A DNN model to be repaired $\mathcal{M}$; set of inputs correctly and incorrectly classified data instance $\{\mathcal{D}rep_{pos}, \mathcal{D}rep_{neg}\} \in \mathcal{D}rep$; the sensitive attributes $S \in \{s_0, s_1\}$; the target layer $l$.
**OUTPUT:** a set of neural weights to target for repair, $\mathbf{w_r}$

1: **function** FairFL
2:     $\mathcal{D}rep_{pos} \leftarrow \text{RandomSample}(\mathcal{D}rep_{pos}, \text{sizeof}=|\mathcal{D}rep_{neg}|)$
3:     $\{\mathcal{D}rep_{1,s_0}, \mathcal{D}rep_{1,s_1}\} \leftarrow$ Inputs belonging to deprived and favored group, respectively, in $\mathcal{D}rep_1$
4:     $\{\mathcal{D}rep_{0,s_0}, \mathcal{D}rep_{0,s_1}\} \leftarrow$ Inputs belonging to deprived and favored group, respectively, in $\mathcal{D}rep_0$
5:     $W_0 \leftarrow \text{compute\_cost\_of\_bias}(\mathcal{D}rep_0, s_0, s_1)$
6:     $W_1 \leftarrow \text{compute\_cost\_of\_bias}(\mathcal{D}rep_1, s_0, s_1)$
7:     $scoredWeights \leftarrow []$
8:     $W^l \leftarrow$ Set of weights in the target layer $l$
9:     $\mathcal{L} \leftarrow$ a loss function used in $\mathcal{M}$
10:     **for** $w^l \in W^l$ **do**
11:         $grad\_loss_{w^l(\mathcal{D}rep_{0,s_0})} \leftarrow \text{Compute\_Gradient\_Loss}(w^l, \mathcal{M}, \mathcal{D}rep_{0,s_0}, \mathcal{L})$
12:         $grad\_loss_{w^l(\mathcal{D}rep_{0,s_1})} \leftarrow \text{Compute\_Gradient\_Loss}(w^l, \mathcal{M}, \mathcal{D}rep_{0,s_1}, \mathcal{L})$
13:         $grad\_loss_{w^l(\mathcal{D}rep_{1,s_0})} \leftarrow \text{Compute_Gradient\_Loss}(w^l, \mathcal{M}, \mathcal{D}rep_{1,s_0}, \mathcal{L})$
14:         $grad\_loss_{w^l(\mathcal{D}rep_{1,s_1})} \leftarrow \text{Compute\_Gradient\_Loss}(w^l, \mathcal{M}, \mathcal{D}rep_{1,s_1}, \mathcal{L})$
15:         $grad\_lossScore_{w^l} \leftarrow W_0 * \frac{grad\_loss_{w^l(\mathcal{D}rep_{0,s_0})}}{1+grad\_loss_{w^l(\mathcal{D}rep_{0,s_1})}} - W_1 * \frac{grad\_loss_{w^l(\mathcal{D}rep_{1,s_0})}}{1+grad\_loss_{w^l(\mathcal{D}rep_{1,s_1})}}$
16:         $fwd\_impact_{w^l(\mathcal{D}rep_{0,s_0})} \leftarrow \text{Compute\_Forward\_Impact}(w^l, \mathcal{M}, \mathcal{D}rep_{0,s_0})$
17:         $fwd\_impact_{w^l(\mathcal{D}rep_{0,s_1})} \leftarrow \text{Compute\_Forward\_Impact}(w^l, \mathcal{M}, \mathcal{D}rep_{0,s_1})$
18:         $fwd\_impact_{w^l(\mathcal{D}rep_{1,s_0})} \leftarrow \text{Compute\_Forward\_Impact}(w^l, \mathcal{M}, \mathcal{D}rep_{1,s_0})$
19:         $fwd\_impact_{w^l(\mathcal{D}rep_{1,s_1})} \leftarrow \text{Compute\_Forward\_Impact}(w^l, \mathcal{M}, \mathcal{D}rep_{1,s_1})$
20:         $fwd\_impactScore_{w^l} \leftarrow W_0 * \frac{fwd\_impact_{w^l(\mathcal{D}rep_{0,s_0})}}{1+fwd\_impact_{w^l(\mathcal{D}rep_{0,s_1})}} - W_1 * \frac{fwd\_impact_{w^l(\mathcal{D}rep_{1,s_0})}}{1+fwd\_impact_{w^l(\mathcal{D}rep_{1,s_1})}}$
21:         **Add tuple** $(w^l, grad\_loss, fwd\_impact)$ **to** $scoredWeights$
22:     **end for**
23:     $\mathbf{w_r} \leftarrow \text{ExtractParetoFront}(scoredWeights)$
24:     **Return** $\mathbf{w_r}$
25: **end function**

---

(5) The values of the gradient loss and forward impact are aggregated to create a combined score for each neuron weight in Lines 15 and 20, respectively. This aggregation gives higher priority to neuron weights that contribute more to class-sensitive input combination, with a specific focus on the deprived community relative to the favored group.
- The scoring function prioritizes the underrepresented class $Y = 0$ (the term on the left of the equations, in Line 15 and Line 20). The numerators in the scoring functions capture the impact of neuron weights on the deprived community, while the denominators capture their impact on the favored community. The scoring maximizes the effect on underrepresented groups (with higher bias, as measured by $W_0$) and minimizes the impact on more represented groups (measured by $W_1$).
- If the cost for $Y = 1$ is greater, then the same equations apply but reverse the roles of $W_0$ and $W_1$. Similarly, if the deprived community is $s_1$, the terms for $S = s_1$ appear in the numerator, and $s_0$ in the denominator.

(6) The algorithm identifies the set of neuron weights $\mathbf{w_r}$ that form the Pareto front based on the combined scores for gradient loss and forward impact (Line 23). The Pareto front consists of neuron weights for which no other weight has both a greater gradient loss and greater forward impact simultaneously. These neurons are the most impactful for repair, as improving them will likely lead to the greatest reduction in bias without negatively affecting model accuracy.

## 4.8 Repair of the fairness in DNNs

In the repair phase of the `FairFLRep` method, the objective is to modify the suspicious neuron weights $\mathbf{w_r}$ identified during the fault localization step to reduce bias and improve fairness in the DNN $\mathcal{M}$. This process is designed to adjust the model without causing a significant drop in overall accuracy. The method utilizes Particle Swarm Optimization (PSO) to explore possible weight changes that enhance fairness while correcting misclassifications.

*4.8.1 Initialization Strategy.* The repair process starts by initializing a population of candidate solutions, where each solution represents an alternative configuration of neuron weights. To avoid drastic changes that might degrade model performance, the search focuses on regions of the weight space near the original values.

For each candidate solution, the initial value of each neuron weight is sampled from a Gaussian distribution. The mean and standard deviation of the distribution are computed using the existing weights within the same layer. This ensures that the search begins in a localized region of the parameter space, close to the original weight values, allowing for gradual adjustments.

*4.8.2 Particle Swarm Optimization (PSO).* PSO is a population-based optimization technique inspired by the social behavior of birds flocking or fish schooling. Each individual in the population (called a "particle") represents a potential solution, which in this context is a set of neuron weights. The particles explore the search space by adjusting their positions based on their own experience and the experience of neighboring particles. The goal of PSO in this context is to find an optimal configuration of neuron weights that improves fairness in the model while maintaining or improving classification accuracy.

Each particle's position (representing a weight configuration) is updated based on two factors:
(1) *pbest* - (personal best): The best position (i.e., the set of weights in the neural network) that an individual particle has found so far during the optimization process.
(2) *gbest* - (global best): The best position found by any particle in the swarm. This serves as a reference point for all particles.

The velocity of each particle is adjusted in each iteration based on its distance from *pbest* and *gbest*, ensuring that particles move toward a combination of their own best experience and the swarm's collective best experience. The position of each particle is updated as follows:

$$pbest_i = \begin{cases} x_i^{t+1}, & \text{if Fitness}(x_i^{t+1}, \mathcal{F}) \leq \text{Fitness}(pbest_i, \mathcal{F}) \\ pbest_i, & \text{otherwise} \end{cases} \quad (21)$$

where:
- $x_i^t$ is the current position of particle $i$,
- $x_i^{t+1}$ is the updated position of particle $i$ based on its velocity,
- $pbest_i$ is the personal best position of particle $i$,
- $\mathcal{F}$ is the fairness metric e.g., *EOD*.

The global best position *gbest* is updated if any particle finds a better solution than the current *gbest*:

$$gbest = \begin{cases} x_i^{t+1}, & \text{if Fitness}(x_i^{t+1}, \mathcal{F}) \leq \text{Fitness}(gbest, \mathcal{F}) \\ gbest, & \text{otherwise} \end{cases} \quad (22)$$

*4.8.3 Fitness Function.* The fitness function plays a crucial role in guiding the PSO optimization process. It evaluates how well each candidate solution (set of neuron weights) meets the specified fairness criteria while also ensuring that the model's classification performance remains acceptable.

The fitness function is defined in terms of a fairness metric, $\mathcal{F}$, such as *SPD*, *EOD*, *FPR* or *DI*. The fairness definition is application-specific, and the user can specify which fairness metric to minimize.

The computation of the fitness for a given particle $p$ works as follows. We first build the corresponding model $\mathcal{M}_p$ by updating the neuron weights $\mathbf{w_r}$ with $p$. Then, we evaluate the repair dataset $\mathcal{D}rep$ using $\mathcal{M}_p$. Based on the results, we identify which is the deprived community (and the favored community as a consequence) of the sensitive attribute; indeed, it could be that, due to the repair action, the deprived community changes w.r.t. that of the original model. Given this information, we compute the fitness value using the fairness metric $\mathcal{F}$.

Note that the fitness function aims at improving the fairness metric, and it does not explicitly consider overall accuracy. Therefore, it could happen that some previously correct input is misclassified by the modified model. In the experiments (RQ2), we will assess how much accuracy is affected by repair.

*4.8.4 Stopping criterion.* The search continues until a predefined stopping criterion is met (e.g., a maximum number of iterations or a satisfactory reduction in unfairness).

## 5 Experiment design

In this section, we introduce the design of the experiments we conducted to assess `FairFLRep`. The approach has been implemented in Keras [34] with a TensorFlow back-end [1]. The code and all the experimental results are available at [53].

### 5.1 Research questions (RQs)

We will evaluate `FairFLRep` using these five research questions (RQs).

**RQ1**: *Can `FairFLRep` effectively improve the fairness of the DNN?*
This question is critical to validating the method's core objective: mitigating bias in DNNs. We aim to determine whether `FairFLRep` can accurately detect and reduce bias, leading to enhanced fairness across various sensitive groups.

**RQ2**: *Can `FairFLRep` effectively improve the fairness without harming the accuracy of the DNN?*
While fairness is a key concern, maintaining high predictive performance is crucial in real-world deployments. We aim to assess whether `FairFLRep` can reduce bias while preserving the model's classification accuracy, ensuring that the method is practical for application in real-world systems.

**RQ3**: *Which fairness metric is better for repairing the fairness of DNN?*
Fairness can be measured using various metrics such as *EOD*, *SPD*, *DI*, and *FPR*, each capturing different aspects of fairness. Optimizing one metric may impact others and model's predictive performance. Understanding which fairness metric leads to the most effective bias mitigation is crucial for guiding the repair process and ensuring alignment with specific fairness concerns in different applications. This RQ aims to identify the most suitable metric for fairness optimization, as well as to explore how these metrics relate with one another, to help uncover potential correlations or trade-offs.

**RQ4**: *Is repairing only the last layer more effective than repairing other layers?*

One key question in model repair is whether fairness can be improved effectively by modifying only the last layer (fully connected layer) instead of adjusting other layers in the network, which is a more costly operation, due to the higher number of weights. This RQ aims to assess whether modifications to the last layer alone are sufficient for bias mitigation, or if fairness-aware repair must target deeper layers (e.g., convolutional layers in CNNs). Understanding this trade-off is crucial for minimizing computational overhead while maximizing fairness improvements.

**RQ5**: *How efficient is `FairFLRep`?*

This question examines the computational efficiency of `FairFLRep` to ensure it can repair fairness quickly and efficiently. This is especially important for large-scale or time-sensitive deployments. By addressing this RQ, we can evaluate whether `FairFLRep` can scale to large datasets and complex models without imposing significant computational overhead.

### 5.2 Benchmark datasets and models

**Image datasets**: We use four well-known image classification datasets for fairness testing including `FairFace` [33], `UTKFace` [79], `CelebA` [46], and Labeled Faces in the Wild (`LFW`) [31]. The images exhibit a broad range of variations in pose, facial expression, illumination, occlusion, resolution, and more. All four datasets provide annotations for binary gender. `FairFace` and `UTKFace` also include annotations for race (*black*, *white*, *Asian*, *Middle Eastern*, and *Indian*); for these datasets, we only select samples from the *black* and *white* categories. The `CelebA` dataset includes a predefined binary "Young" attribute to represent age, where "1" signifies "*young*" and "0" signifies "*not-young*" (or "*old*"), as originally assigned by the dataset authors.

**Tabular datasets**: We also include four real-world well-known tabular datasets for fairness assessment: Student Performance (`Student`) [15], COMPAS Recidivism [68], Adult Income (`Adult`) [4], and Medical Expenditure Panel Survey (`MEPS`) [2]. The `Student` dataset contains academic performance data from Portuguese secondary schools, with final grades as the target variable. `COMPAS` includes over 10,000 criminal defendant records used for recidivism risk assessment, known for racial bias in false positive rates [12]. The `Adult` Income dataset is used for income classification, with protected attributes including *gender*, *race*, and *age*. `MEPS` is a healthcare dataset for predicting medical expenditure, with *gender* as the primary protected attribute.

As explained in Section 4.3, the fault localization of `FairFLRep` (i.e., FairFL) differs depending on whether the sensitive attribute is the same as attribute under classification (`SettSame`), or the sensitive attribute and the class label are different (`SettDiff`). We experiment both cases as follows:

`SettSame`: we use all image datasets `UTKFace`, `FairFace`, `LFW`, and `CelebA`, using *gender* as attribute under classification, and also as sensitive attribute. We do not use tabular datasets, as they are not applicable in this setting;

`SettDiff`: for image datasets, we use datasets `UTKFace`, `FairFace`, and `CelebA` (`LFW` is omitted due to lack of race or age group labels), using *gender* as attribute under classification for all of them; the sensitive attribute is race (*black* or *white*) for the `UTKFace` and `FairFace`, and *age* group (*young* or *old*) for the `CelebA` dataset. For tabular datasets, we use all of them (i.e., `Student`, `COMPAS`, `Adult`, and `MEPS`). The attribute under classification is: *exam result* for `Student` (*passed* or *failed*), *reoffending* for `COMPAS` (*yes* or *no*), *high income* for `Adult` (*yes* or *no*), and *utilize* for `MEPS` (*yes* or *no*). We experiment with two sensitive attributes: *gender* for all datasets, and *race* for `Adult`, `COMPAS`, and `MEPS`.

**Models.** We experiment with two DNN models of different architectures and complexities:

*Image datasets*: For the image datasets, we experiment with two DNN models of different architectures and complexities. (1) VGG16: A CNN composed of a series of convolutional and pooling layers, followed by three dense layers (fully connected layers). (2) LeNet-5: A relatively small network with seven layers, consisting of convolutional, pooling, and fully connected layers, making it a foundational model for deep learning.

Each model is trained separately on the four datasets, resulting in a total of eight models. These models are then used as inputs to evaluate the compared approaches.

*Tabular datasets*: For tabular datasets, we train a 1D CNN to process structured data effectively. Instead of traditional fully connected layers, we leverage Conv1D layers to extract feature patterns from tabular inputs, treating each sample as a 1D sequence of numerical features. This architecture enables efficient feature extraction from tabular datasets, leveraging spatial dependencies among features while maintaining computational efficiency.

### 5.3 Compared approaches

FairFLRep. It is our proposed fairness-aware method where both the fault localization and repair phases are explicitly optimized with fairness in mind. The method takes as input a fairness metric $\mathcal{F}$ that needs to be addressed (*EOD*, *SPD*, *DI*, or *FPR*); we indicate as FairFLRep$_\mathcal{F}$ the approach initialized with fairness metric $\mathcal{F}$. As explained in Section 2.1, the approach takes in input a repair dataset $\mathcal{D}rep$, split between misclassified inputs $\mathcal{D}rep_{neg}$ and correctly classified inputs $\mathcal{D}rep_{pos}$.

FairArachne. In order to assess whether considering fairness both in fault localization and repair is necessary, we experiment with an approach (named FairArachne) that does not use our proposed fault localization approach, but the one from Arachne; the repair phase, instead, uses our FairRep repair technique (which is fairness-aware). This setup helps isolate the impact of using fairness-aware fault localization. We indicate with FairArachne$_\mathcal{F}$ the approach that uses the metric $\mathcal{F}$ (e.g., *SPD*, *EOD*, etc.) during the repair process. For consistency, we use the same repair dataset $\mathcal{D}rep$ used in FairFLRep; the only difference is that, by following Arachne's approach [64], the negative samples $\mathcal{D}rep_{neg}$ used in fault localization only contain the deprived community.

FairMOON. It is a variant of FairFLRep that combines MOON's spectrum-based fault localization [27] with FairFLRep's fairness-aware repair. We indicate with FairMOON$_\mathcal{F}$ the approach that uses the fairness metric $\mathcal{F}$ during the repair process. We use the same repair data as in FairArachne.

Arachne. It is a widely recognized repair technique [64] primarily designed to address misclassification errors in DNNs. Although originally intended for general model repair, the authors have noted that Arachne can be applied to fairness-related domains. We follow the guidelines from the original paper to adapt Arachne for fairness repair. As repair dataset, we use the same used in FairArachne.

LIMI. It is a fairness testing method based on generating natural discriminatory instances through surrogate boundary approximation in a generative model's latent space [71]. To adapt LIMI for fairness repair evaluation, we follow the retraining strategy proposed by the original authors and by Zhang et al. [77]: discriminatory instances generated by LIMI are randomly selected at 30% the size of the original training dataset, then combined with the original data, and the model is retrained on this augmented dataset. We use LIMI as a baseline to assess how retraining-based bias mitigation compares to direct fault localization and neuron-level repair, as proposed in FairFLRep.

### 5.4 Experimental setup

An *experiment* is the execution of one approach (see Section 5.3), over a benchmark (i.e., a model trained over a dataset), for one of the fault localization settings. For image datasets, there are eight benchmark models (two models trained over four datasets); `SettSame` is applicable for all the eight benchmark models, while `SettDiff` for six benchmark models. For tabular datasets, there are four benchmarks (one model trained over four datasets) that can only experimented over `SettDiff`; for three datasets, we experiment with two sensitive attributes, while for a dataset we experiment with only one. So, in total, there are 21 experiments. Note that `LIMI` did not produce any discriminative input for the dataset `Student` and, therefore, cannot be used for this dataset. To account for the randomness of the search algorithm used in repair, by following [3], each experiment has been executed 30 times.

We configure the PSO algorithm used during `FairFLRep` repair using the same setting as in the original `Arachne` paper [64]. Each individual repair is represented by a swarm of 100 particles, where each particle corresponds to a candidate solution (i.e., a specific weight adjustment). The optimization runs for a maximum of 100 generations, but includes an early stopping criterion: if the best candidate solution does not improve for 10 consecutive generations, the search terminates early. Regarding `FairMOON`, for fault localization, we use the *DStar* formula with exponent 3, as recommended in the original MOON paper [27]. We select the top 15% suspicious neurons ($\lambda = 0.15$) for repair based on suspiciousness scores. Test input selection is guided by MOON's Suspicious Neuron Activation (SNA) and Test Input Diversity (TID) objectives. Regarding `LIMI` [71], we run `LIMI` using the implementation and default hyperparameters provided in the original replication package. Specifically, the candidate probing parameter $\lambda$ is set to 0.3, as recommended by the authors to balance between avoiding duplicate test cases ($\lambda$ too small) and preventing samples from deviating too far from the decision boundary ($\lambda$ too large). For generative modeling, we use CTGAN with the default settings: 300 training epochs and batch size of 500 for each tabular dataset.

## 5.5 Evaluation metrics and statistical analysis

To validate the effectiveness of our `FairFLRep` technique against baseline methods, we compare their results in terms of:

*Fairness Metrics ($\mathcal{F}$)*: We employ four group fairness metrics: *DI*, *SPD*, *EOD* and *FPR* (see Section 2.4). These metrics allow us to capture different aspects of group fairness in DNN classifiers.

*Accuracy*: The accuracy of the repaired model is crucial to assess its overall utility. `FairFLRep` aims to improve fairness while preserving the model's accuracy as much as possible.

We will report the distributions of the values of the metrics across different runs and compare them using appropriate statistical tests [3]. Specifically, given an experiment (i.e., a given benchmark model and a repair setting (`SettSame` or `SettDiff`)) and a metric $M$ (a fairness metric or accuracy), we compare the distributions of $M$ of `FairFLRep` and a baseline approach across the 30 runs. Specifically, we employ the Mann-Whitney U test to assess the statistical significance difference of the distributions. If the *p*-value is less than 0.05, we reject the null hypothesis that there is no statistically significant difference. In case of significant difference, we use Vargha and Delaney's $\hat{A}_{12}$ [3] effect size statistic to quantify the magnitude of the difference. A value of $\hat{A}_{12}$ greater than 0.5 means that then first approach produces significantly higher values of $M$; a value of $\hat{A}_{12}$ lower than 0.5 means the opposite. We further categorize the effect size as follows [36]: *negligible* if $0.5 \leq \hat{A}_{12} < 0.556$ or $0.494 \leq \hat{A}_{12} < 0.5$; *small* if $0.556 \leq \hat{A}_{12} < 0.638$ or $0.362 \leq \hat{A}_{12} < 0.494$; *medium* if $0.638 \leq \hat{A}_{12} < 0.714$ or $0.286 \leq \hat{A}_{12} < 0.286$; *large* if $\hat{A}_{12} \geq 0.714$ or $\hat{A}_{12} \leq 0.286$.

We categorize the results of the comparison of `FairFLRep` with a baseline approach into:

*Win* (W): `FairFLRep` outperforms the baseline approach with at least small $\hat{A}_{12}$ value. We will use ✓, ✓✓, and ✓✓✓ to indicate a small, medium, and large $\hat{A}_{12}$ value.

*Tied* (T): The difference between `FairFLRep` and the baseline is not significant or negligible.

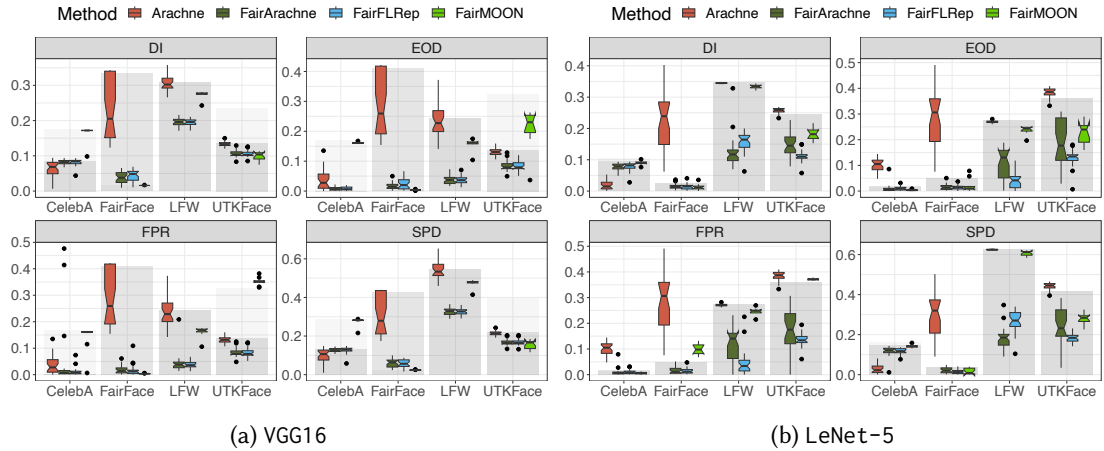

(a) VGG16    (b) LeNet-5

Fig. 3. RQ1 – `SettSame` – Image datasets – Fairness of the repaired models (the lower the value, the better). The gray-bar plot indicates the fairness of the original model.

*Loss* (L): The baseline outperforms `FairFLRep` with at least small $\hat{A}_{12}$ value. We will use ×, ××, and ××× to indicate a small, medium, and large $\hat{A}_{12}$ value.

## 6 Experimental Results

This section presents the results of our empirical evaluation, addressing the research questions outlined in Section 5.1. It is important to note that all reported results are based on the test sets.

### 6.1 RQ1 – Can `FairFLRep` effectively improve fairness in DNN?

We evaluate the effectiveness of our proposed approach `FairFLRep` in the task of repairing model fairness. The evaluation in this section is broken into two sub-experiments for two fault localization settings (`SettSame` and `SettDiff`); for image datasets, we compare `FairFLRep` with `Arachne`, `FairArachne`, and `FairMOON`; for tabular datasets, we compare it with `Arachne`, `FairArachne`, `FairMOON`, and `LIMI`.

*6.1.1 `SettSame` – Image datasets.* Here, we present the results of repairing bias and unfairness in gender prediction for the two DNN models trained on the four image datasets.

Figure 3 shows the results of the repaired models using `FairFLRep`, `FairArachne`, `Arachne`, and `FairMOON` for each benchmark. The figure groups results based on the fairness metrics used during the repair process (*DI, EOD, SPD,* and *FPR*) applied to the VGG16 and LeNet-5 models. The gray bars represent the fairness score of the original model.

Overall, `FairFLRep` outperforms `Arachne` in most of the cases. `FairFLRep` exhibits greater stability, with smaller deviations across different runs, indicating that it produces consistent results. This stability is observed across multiple datasets. The consistent results suggest that integrating fairness considerations into both the fault localization and repair stages leads to more reliable and fair models across sensitive groups. `FairArachne` shows moderate improvements in fairness, outperforming `Arachne` in several cases. However, it does not perform as well as `FairFLRep`, particularly in dataset like `UTKFace`.

`FairMOON`, similar to `FairArachne`, occasionally matches or slightly outperforms `FairFLRep` on specific cases (e.g., `FairFace` with VGG16). However, its performance is highly inconsistent across datasets and fairness metrics. For example, `FairMOON` significantly underperforms on *DI, EOD,* and *FPR* on `UTKFace` and `CelebA`, especially with the `VGG16` model. Moreover, while `FairMOON` shows competitive performance on `FairFace` (`VGG16`), this is not the case for `LeNet-5`, where it severely degrades the original model's fairness, particularly on *FPR*. These inconsistencies highlight `FairMOON`'s limited reliability as a comprehensive fairness repair method.

Unlike `FairFLRep`, `Arachne` occasionally degrades the fairness of the original model. In certain cases, the repaired model performs worse than the original, especially on metrics like *DI*, *FPR*, and *EOD* for `LeNet-5` models trained on the `FairFace` and `CelebA` datasets (see the exceptions discussed below). This degradation is likely because `Arachne` focuses more on optimizing accuracy, without sufficiently addressing fairness disparities. In contrast, `FairFLRep` preserves the original fairness of the model.

According to the 80% rule [23], which considers a *DI* value > 0.2 as unfair, some of the models repaired using `Arachne` on the `FairFace` dataset had *DI* values ranging between 0.24 and 0.4, indicating significant unfairness. These results suggest that `FairFLRep` is more effective at reducing unfairness in DNNs while preserving the overall performance, whereas `Arachne` may not provide the same level of fairness improvement.

*Consistency Across Datasets.* The results from multiple datasets highlight that `FairFLRep` provides more reliable fairness improvements compared to `Arachne`, `FairArachne`, and `FairMOON`. While `Arachne` shows some improvement in specific cases, its focus on accuracy often leads to inconsistent fairness outcomes. The following points summarize the key insights observed across the datasets:

(1) `FairFLRep` consistently provides more balanced and comprehensive fairness improvements across various metrics and datasets, demonstrating its versatility and effectiveness in addressing fairness concerns across diverse settings.

(2) Interestingly, on the `CelebA` dataset, `Arachne` performs relatively well for *DI* and *SPD*, where it achieves lower fairness scores compared to other datasets and metrics. One potential reason for this is that the original `VGG16` and `LeNet-5` models trained on `CelebA` already exhibit relatively fair behavior, which gives `Arachne` a better starting point to improve accuracy without significantly harming fairness.

However, for other metrics such as *EOD* and *FPR*, and across other datasets, `FairFLRep` consistently outperforms `Arachne`. This suggests that while `Arachne` may excel in specific scenarios—particularly when the original model is already relatively fair—it lacks robustness across a broader range of fairness measures and datasets.

(3) While `FairArachne` generally underperforms compared to `FairFLRep`, there are a few instances where `FairArachne` outperforms `FairFLRep` on specific metrics, such as *DI* and *SPD* on the `LFW` dataset. This outcome may be due to `FairArachne`'s ability to focus on fairness during the repair phase alone, which may allow it to more directly target group-level disparities (as measured by *DI* and *SPD*). Additionally, `LFW` might present simpler bias patterns, where fairness issues are more straightforward (e.g., fewer confounding factors), making `FairArachne` more effective in optimizing these specific fairness metrics.

(4) `FairMOON` demonstrates competitive results on certain datasets and metrics but fails dramatically on others, particularly `UTKFace` and `CelebA`, reflecting instability across diverse settings.

> **Answer to RQ1 (`SettSame` – Image datasets).** Across most datasets (3 out of 4), using `FairFLRep` provides a more consistent and robust solution for improving fairness in DNNs. It outperforms `Arachne` across most fairness metrics and datasets. While `FairArachne` provides some gains in fairness, it lacks the robustness and consistency achieved by `FairFLRep`,

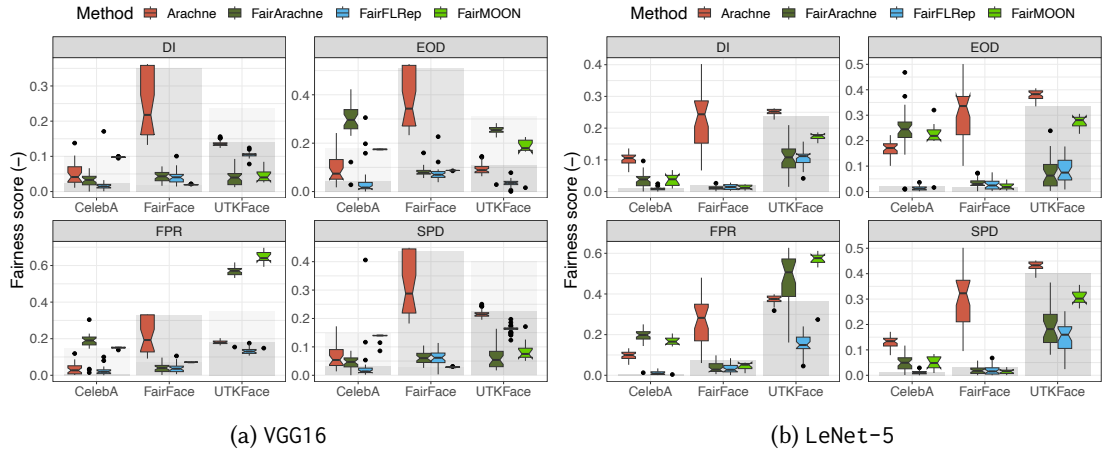

Fig. 4. RQ1 – `SettDiff` – Image datasets – Fairness of the repaired models (the lower the value, the better). The gray-bar plot indicates the fairness of the original model.

> especially in reducing bias across a broader range of fairness metrics and datasets. `FairMOON`, though often better than `Arachne`, shows variable performance across datasets and metrics. This validates the effectiveness of applying fairness-aware techniques in both stages (fault localization and repair), resulting in significantly better fairness outcomes for the models.

*6.1.2* `SettDiff` — *Image datasets.* We present the results of repairing bias in image datasets: race disparity in the `UTKFace` and `FairFace` datasets, and age bias in the `CelebA` dataset. In all the cases, the class being predicted is gender. Figure 4 reports the results. As in the previous figure (Figure 3), the boxplots report the value of the fairness metric $\mathcal{F}$ under repair in the repaired model, and the gray bars represent the fairness scores of the original model.

According to Figure 4, `Arachne` shows some improvement over the original model, but its impact remains limited across all datasets and fairness metrics. Overall, `Arachne` consistently struggles to reduce fairness scores, highlighting its limitations in fairness-related tasks.

`FairArachne`, which incorporates fairness only during the repair phase, is more effective than `Arachne` across most metrics and datasets. However, it performs worse than the original model on the `CelebA` dataset for all fairness metrics. This suggests that `FairArachne`, although it is fairness-aware during the repair phase, may not localize the correct set of faulty neurons responsible for fairness disparities, especially in a dataset like `CelebA` where the original models already show relatively fair behavior. In such cases, fairness improvements may be insufficient, or in some instances, the adjustments may degrade the original fairness of the model, as seen in both `Arachne` and `FairArachne` performing worse on `CelebA`.

In contrast, `FairFLRep` outperforms both `Arachne` and `FairArachne` across all metrics and datasets, consistently lowering fairness scores with minimal variation. `FairFLRep` demonstrates higher consistency and robustness in mitigating bias across multiple datasets and metrics. This reinforces the advantage of using `FairFLRep`, which incorporates fairness awareness in both the fault localization and repair phases, effectively reducing bias and with minimal variation. `Arachne`, by comparison, shows greater variability and, in some cases, even performs worse than the original model in terms of fairness. `Arachne`'s emphasis on optimizing accuracy may contribute to this issue, potentially leading to greater disparities in treatment across sensitive groups.

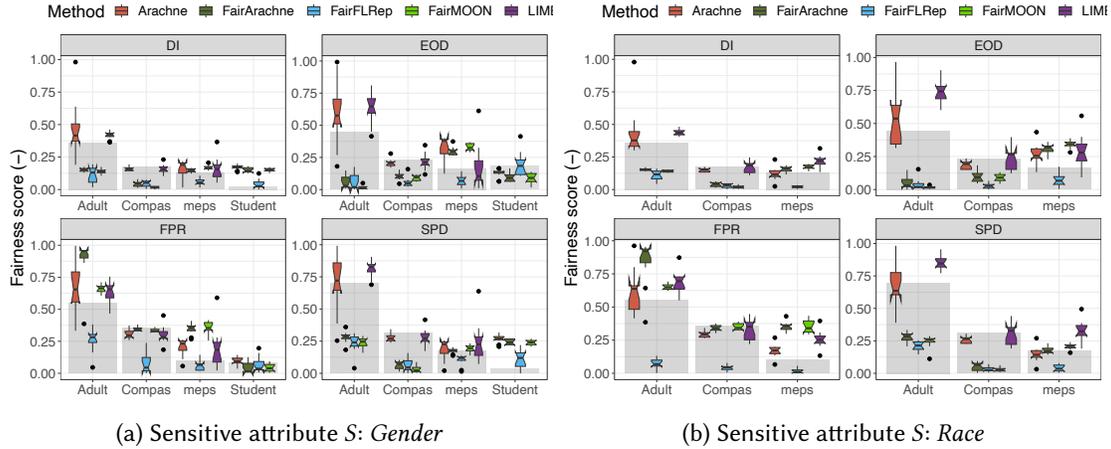

(a) Sensitive attribute S: *Gender*  (b) Sensitive attribute S: *Race*

Fig. 5. RQ1 – `SettDiff` – Tabular datasets – Fairness of the repaired models (the lower the value, the better). The gray-bar plot indicates the fairness of the original model.

`FairMOON` sometimes produces competitive results on some metrics and datasets but shows higher variability compared to `FairFLRep`. For example, on `FairFace` in VGG16, `FairMOON` outperforms in *SPD* and *DI*, but fails to match `FairFLRep`'s consistency across *EOD* and *FPR*, especially on `CelebA`. This further reinforces the importance of a methodical integration of fairness objectives across the full repair pipeline.

> **Answer to RQ1 (`SettDiff` – Image datasets).** The consistently lower fairness scores with with minimal variation across all metrics in `FairFLRep` indicate that this method is particularly effective at reducing bias, while remaining robust across datasets. For both fairness repair tasks (addressing disparities in *age* and *race*), these improvements are more pronounced in `FairFLRep`, that is fair-aware both in the fault localization and repair phases. While `FairFLRep` and `FairMOON` offers moderate improvements, they lacks the robustness and consistency of `FairFLRep`, particularly when dealing with more complex fairness issues such as race and age disparities. This observation suggests that incorporating fairness mechanisms in both the fault localization and repair stages is crucial to achieving comprehensive fairness in DNNs.

*6.1.3 `SettDiff` – Tabular datasets.* We now consider the `SettDiff` scenario using tabular data. We evaluate fairness repair for two cases: *gender* bias and *race* bias, using the four tabular datasets. For both bias cases, the predicted class is the same.

*Gender bias repair.* Figure 5a presents the fairness scores for models repaired for *gender* bias across the four tabular datasets and the four fairness metrics. Each bar group in the figure represents the results for one dataset and one fairness metric. The compared methods are `FairFLRep` against `Arachne`, `FairArachne`, `FairMOON`, and `LIMI`. According to Figure 5a, `FairFLRep` achieves the lowest fairness scores (i.e., best fairness) across most datasets and metrics, outperforming all baselines, particularly in *DI*, *EOD*, and *FPR*. `FairArachne` performs moderately better than `Arachne` and `LIMI` in some cases, though its improvements are not consistent across datasets. `FairMOON` shows competitive performance for *SPD*, especially on the `COMPAS` dataset, but tends to underperform compared to `FairFLRep` in *EOD* and *FPR*. `LIMI` performs inconsistently; while it sometimes improves fairness compared to `Arachne`, it is generally less stable and less effective overall. Similarly, `Arachne` shows some marginal improvement in certain metrics (notably on `MEPS` for *SPD*), but often falls short or even increases bias due to its lack of fairness awareness.

*Race bias repair.* Figure 5b displays the results for repairing *race* bias using the tabular datasets, except for the `Student` dataset which does not include race as a sensitive feature. The results indicate that `FairFLRep` continues to provide the most significant fairness reductions, especially in *DI* and *FPR*, across all datasets, affirming its advantage in handling race-related bias. `FairArachne` again outperforms `Arachne` in several cases but shows increased variance in its results, indicating lower stability across fairness metrics. Notably, `FairArachne` and `Arachne` perform worse than the original model on some metrics in the `COMPAS` and `MEPS` datasets, highlighting their potential to degrade fairness under distribution shifts. `FairMOON` provides competitive results on *SPD* in the `Adult` dataset but shows inconsistency in other metrics. `LIMI` struggles to consistently match the fairness gains of `FairFLRep`, particularly on datasets like `MEPS` and `Adult`.

Results of both biases reinforce the benefit of incorporating fairness considerations into both the fault localization and repair phases, as done in `FairFLRep`. The method reliably localizes fairness-related faults and performs targeted interventions that generalize across broader datasets, thereby achieving better fairness outcomes for both gender and race biases.

> **Answer to RQ1 (`SettDiff` – Tabular datasets).** Results show the consistent superiority of `FairFLRep` across all tabular datasets, demonstrating its robustness in identifying and repairing fairness issues. `FairFLRep` achieves more reliable and significant improvements across multiple fairness metrics when compared to baselines, for both gender and race.

## 6.2 RQ2 – Can `FairFLRep` effectively improve the fairness without harming the accuracy?

So far, we have observed that `FairFLRep` can effectively improve fairness in DNN models. However, it is equally important to consider the impact of these fairness improvements on the model's predictive performance. Indeed, it could happen that, in order to increase fairness, some previously correctly classified input is misclassified in the repaired model. If fairness constraints can be optimized while maintaining or minimally affecting accuracy, this would demonstrate the effectiveness of the proposed repair technique in balancing fairness and performance. Below, we assess the accuracy outcomes of `FairFLRep` compared to the baseline approaches.

*6.2.1 `SettSame` – Image datasets.* This section reports the accuracy metrics for *gender* prediction for all the models and datasets. The accuracy is broken down by sensitive categories (*female* and *male*) as well as the overall accuracy for each dataset. Figure 6 illustrates the accuracy metrics of the repaired models using the four compared methods. The figure also reports the value of the accuracy of the original model (three dashed lines showing overall accuracy, and accuracy for *male* and *female*). We observe that the original model shows varied performance across the datasets, with a notable gender gap in accuracy. In most cases, the model performs better for one gender group over the other, indicating a potential imbalance in accuracy before any repair is applied.

Regarding the results of the repaired models, `Arachne` tends to exhibit significant differences in accuracy between *female* and *male* groups across most datasets, although it improves overall accuracy in some cases. In contrast, `FairFLRep` achieves more balanced accuracy between *male* and *female* groups in the `LFW`, `CelebA`, and `FairFace` datasets. For instance, VGG16 repaired with `FairFLRep` achieves a median accuracy of 70% for *female* and 72% for *male*, significantly reducing the gender disparity. This more balanced accuracy within the sensitive attribute is due to the improvement of fairness (i.e., they are correlated).

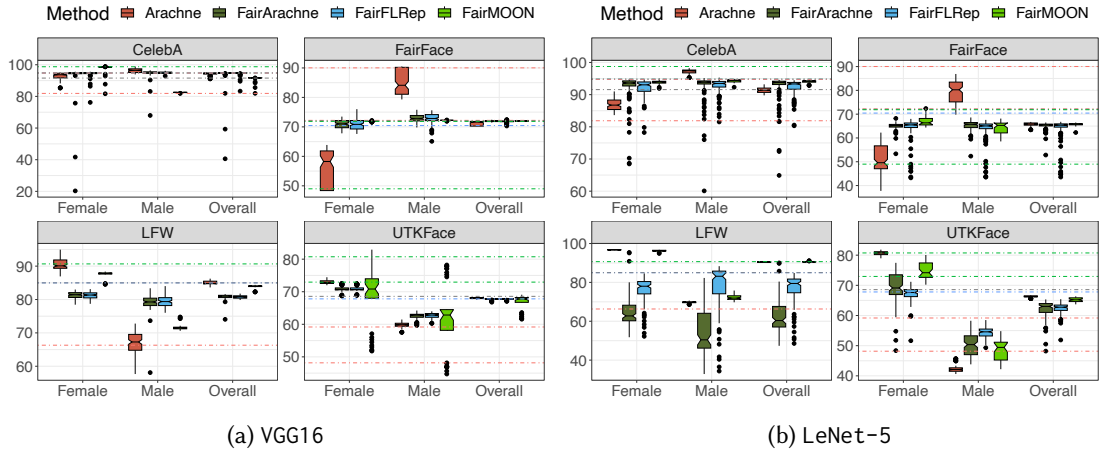

Fig. 6. RQ2 – `SettSame` – Image datasets – Accuracy of the repaired models

`FairMOON` produces more balanced accuracy than `Arachne` in some datasets (particularly `FairFace`) but struggles in others. It shows notable gender gaps in `UTKFace` and `CelebA`, especially with the `LeNet-5` model on `UTKFace`, and with both architectures on `CelebA` and `LFW`. While `FairMOON` improves overall accuracy, the results highlight its limited ability to ensure balance across sensitive groups. `FairArachne` delivers moderate improvements in balanced accuracy, generally outperforming `Arachne` and `FairMOON`, but still falling short of the performance of `FairFLRep` in most datasets. There are some datasets where the accuracy for the *male* or *female* categories is less balanced, which suggests that the combination considering fairness only in repair provides some benefits but lacks the precision and consistency of considering fairness in both stages.

In some cases, particularly for the `LFW` dataset with `LeNet-5` model, `FairArachne` resulted in a noticeable drop in accuracy when fairness was improved in some datasets (see fairness score in Figure 3b). This suggests that `FairArachne` might not be accurately identifying the problematic weights for fairness repair, causing a conflict between accuracy and bias mitigation. However, this issue is not present in the `FairFLRep` method, where fairness is explicitly considered in both the fault-localization and repair phases. This ensures that the correct weights are targeted for adjustment, allowing the model to maintain a balanced accuracy while effectively addressing bias. These results validate the effectiveness of the `FairFLRep`, confirming its superiority in achieving fairness without compromising performance.

> **Answer to RQ2 (`SettSame` – Image datasets).** The results indicate a clear pattern: (i) While `Arachne` improves accuracy, it fails to fully address fairness, as disparities often persist. (ii) `FairArachne`, which introduces fairness-awareness solely in the repair phase, improves fairness, but a significant drop in accuracy on some datasets (e.g., LFW) suggests a mismatch between `Arachne`'s fault localization and fairness-aware repair. (iii) `FairMOON`, although it improves overall accuracy, has a limited ability to achieve balance across sensitive groups, as it shows notable gender gaps in `UTKFace` and `CelebA`. (iv) In contrast, `FairFLRep` effectively reduces bias across all datasets (especially in highly biased ones like `LFW` and `UTKFace`)

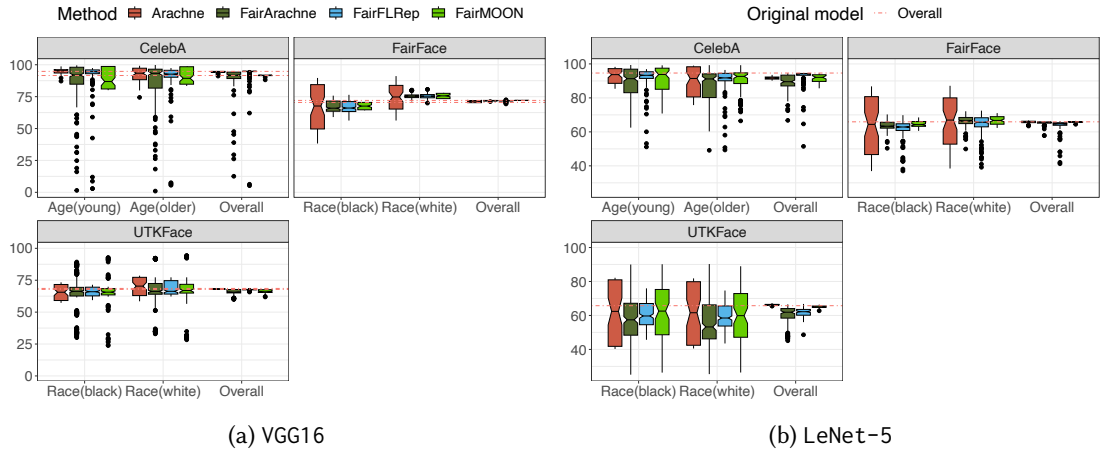

(a) VGG16  (b) LeNet-5

Fig. 7. RQ2 – `SettDiff` – Image datasets – Accuracy of the repaired models

> while maintaining or enhancing overall accuracy. This demonstrates `FairFLRep`'s ability to target the appropriate weights, achieving fairness without compromising performance.

*6.2.2 `SettDiff` – Image datasets.* Figure 7 compares the performance of `FairFLRep` against baselines in the context of `SettDiff`, where the prediction class (*gender*) is different from the sensitive attribute (*race* or *age*). For this setting, only datasets `UTKFace`, `CelebA`, and `FairFace` are used; the accuracy is presented for different sensitive groups (*race*: *black* vs. *white*; *age*: *young* vs. *old*) and overall accuracy.

Across all datasets, `FairFLRep` consistently maintains competitive accuracy compared to `Arachne`, and even outperforms `FairArachne` in some cases (e.g., in `UTKFace` with VGG16, and `CelebA` with both VGG16 and LeNet-5). These results further validate the effectiveness of `FairFLRep` in targeting the appropriate weights for repair without compromising accuracy across diverse fairness repair tasks, similarly to the observations in `SettSame`. The `FairFLRep` method successfully minimizes disparities between sensitive groups (*race*, *age*), ensuring more balanced accuracy while also maintaining or improving overall model performance. `FairMOON` shows mixed results. While it demonstrates relatively strong overall accuracy in some cases—such as `FairFace` with VGG16–it frequently shows wider disparities across some sensitive groups. On the `UTKFace` dataset, `FairMOON` presents noticeable accuracy fluctuations between sensitive groups, suggesting instability and a less reliable group-level performance. On `CelebA`, `FairFLRep` slightly outperforms `FairMOON` both in terms of group-level and overall accuracy, indicating that `FairFLRep` achieves better trade-offs between fairness and model performance, even in competitive scenarios.

> **Answer to RQ2 (`SettDiff` – Image datasets).** Our evaluation indicates that `FairFLRep` can effectively minimize disparities between *race* and *age* sub-groups. It ensures more balanced accuracy while maintaining or improving overall model performance.

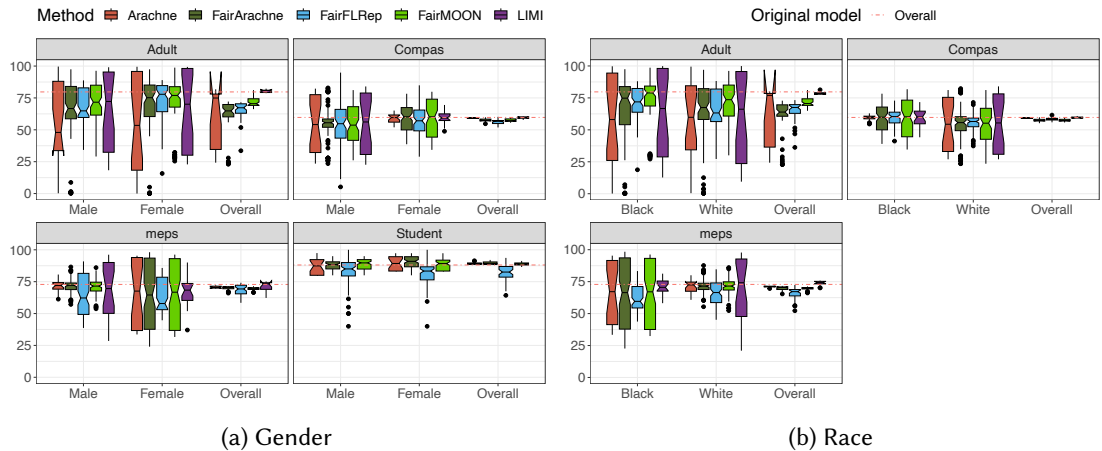

Fig. 8. RQ2 – `SettDiff` – Tabular datasets – Accuracy of the repaired models

*6.2.3 `SettDiff`– Tabular datasets.* Figure 8 compares the accuracy performance of `FairFLRep` against the baseline methods under `SettDiff` settings across tabular datasets, with a focus on sub-groups disparities (Gender and Race) and overall accuracy.

According to the results, `Arachne` and `LIMI` exhibit pronounced Gender and Race accuracy gaps, revealing a contrasting performance pattern compared to `FairFLRep`. While `Arachne` and `LIMI` improves the overall accuracy, they often introduce or exacerbate disparities between subgroups across most datasets. This suggest that their emphasis on maximizing the global accuracy may come at the cost of individual accuracies, especially for minority groups.

In contrast, `FairFLRep` consistently demonstrates a more balanced accuracy between subgroups across all datasets. It achieves this without significantly sacrificing overall accuracy, underscoring its strength in reducing subgroups disparities while maintaining competitive model performance.

When compared to `FairArachne`, the results show that `FairArachne` achieves relatively more balanced subgroup accuracy than `Arachne` or `LIMI`. However, `FairArachne` also exhibits instability in certain cases– for example, large accuracy variations are observed in the *Female* and *Black* subgroups for the MEPS dataset–indicating a lack of robustness.

`FairMOON` performs competitively in terms of overall accuracy, but shows noticeable variability within minority groups, particularly for *Female* and *Black* subgroups in the MEPS and COMPAS datasets. This subgroups disparities suggest that `FairMOON`, while promising, may not consistently deliver equitable outcomes across sensitive attributes.

> **Answer to RQ2 (`SettDiff` – Tabular datasets).** `FairFLRep` demonstrates superior subgroups balance and competitive overall accuracy across the tabular datasets. Unlike `Arachne` and `LIMI`, which often trade subgroup fairness for overall accuracy, `FairFLRep` achieves both. While `FairArachne` and `FairMOON` show some promise, their inconsistency across minority subgroups limits their reliability. `FairFLRep` remains the most stable and effective solution for mitigating Gender and Race disparities in tabular classification tasks.

## 6.3 Statistical analysis of the model fairness and model accuracy

In this section, we present the statistical analysis of the fairness and accuracy of models repaired using different techniques, to summarize the observations of RQ1 and RQ2 presented in Section 6.1 and Section 6.2. As before, the analysis is split for the two fault localization settings.

*6.3.1 `SettSame` – Image datasets.* In Table 2, we report the statistical comparison between the different repair methods in terms of fairness and overall accuracy of the repaired models. As explained in Section 5.5, we use the categorization W to indicate where `FairFLRep` outperforms (i.e., wins) the baseline (columns); T where the comparison of the two methods is tied; and L when the baseline outperforms `FairFLRep`. Each subtable reports the results for the different fairness metrics $\mathcal{F}$ used as fitness functions. For example, `FairFLRep`$_{EOD}$ indicates the case where `FairFLRep` used the *EOD* as the fairness definition. Each subtable reports the results for the fairness metric and for the accuracy. At the bottom of the table, the summary provides an overall count of wins, ties, and losses across different datasets and models, giving a comprehensive view of the effectiveness of `FairFLRep`.

According to results in Table 2, `FairFLRep` achieved a high number of wins, often by large margins, when compared to `Arachne` in terms of fairness results (29 out of 32 cases). There were only 3 cases where the `FairFLRep` approach lost to `Arachne`, and no ties were observed in the fairness metrics. However, this trend does not hold when comparing the overall accuracy of models repaired by `FairFLRep` to those repaired by `Arachne`. In this case, `FairFLRep` only won in 10 out of 32 cases, while losing in 19 cases. This outcome aligns with the focus of `Arachne` on correcting misclassification problems. Conversely, our results aim to address both fairness and model predictive performance without significantly sacrificing either.

The comparison between `FairFLRep` and `FairArachne` further supports this claim: in terms of fairness metrics, `FairFLRep` won in 7 out of 32 cases, and tied in 23; in terms of overall accuracy, it won in 4 cases, and tied in 26. This demonstrates that `FairFLRep` can achieve favorable fairness outcomes while maintaining competitive model accuracy, highlighting its effectiveness in balancing these critical aspects.

When comparing with `FairMOON`, the results are more mixed. In terms of fairness, `FairFLRep` achieved 15 wins, 7 ties, and 4 losses out of 32 comparisons, indicating that `FairMOON` underperforms in fairness compared to `FairFLRep`. Notably, 3 our 4 fairness loses occurred on the VGG16 model repaired on the `FairFace` dataset, highlighting cases where `FairMOON` may outperform `FairFLRep` on specific datasets or metrics. However, `FairFLRep` maintains overall superiority in fairness, as reflected in the higher total number of wins, while also remaining competitive in accuracy, with 7 ties recorded in accuracy comparisons. These results suggest that while `FairMOON` may excel in certain configurations, `FairFLRep` offers more reliable fairness improvements across diverse datasets and model architectures.

> **Answer to RQ1 and RQ2 (`SettSame` – Image datasets).** In summary, `FairFLRep` can improve model fairness across various metrics, often by large margins, in a statistical significant way. This improvement is achieved with minimal impact on overall accuracy when compared to the baselines `Arachne`, `FairArachne` and `FairMOON`. While `FairFLRep` tends to sacrifice some accuracy, it remains a highly effective for balancing fairness with predictive performance.

*6.3.2 `SettDiff` – Image datasets.* In Table 3, we report the statistical comparison of the repair approaches using the image datasets in `SettDiff` setting. The results highlight `FairFLRep`'s strong performance, particularly in fairness outcomes across different datasets and model architectures. FairFLRep demonstrates significant wins across all fairness metrics, consistently outperforming Arachne, FairArachne, and FairMOON in fairness metrics. This confirms FairFLRep's effectiveness in reducing group-level bias under complex fairness scenarios.

Table 2. RQ2 – SettSame – Image datasets – Statistical comparison between FairFLRep$_\mathcal{F}$ and the baselines. W: FairFLRep$_\mathcal{F}$ wins against the baseline; T: no statistical significant difference; L: FairFLRep$_\mathcal{F}$ loses to the baseline. The number of symbols ✓ and × represents the difference magnitude (1: small, 2: medium, 3: large).

| Dataset | Model | FairFLRep$_{EOD}$ vs Arachne | | FairFLRep$_{EOD}$ vs FairArachne$_{EOD}$ | | FairFLRep$_{EOD}$ vs FairMOON$_{EOD}$ | |
|---|---|---|---|---|---|---|---|
| | | EOD | Accuracy | EOD | Accuracy | EOD | Accuracy |
| UTKFace | VGG16 | W (✓✓) | L (×××) | T | T | W (✓✓✓) | W (✓✓✓) |
| | LeNet-5 | W (✓✓) | L (×××) | W (✓✓) | T | W (✓✓✓) | L (×××) |
| Fair Face | VGG16 | W (✓✓✓) | W (✓✓✓) | T | T | L (×××) | L (×××) |
| | LeNet-5 | W (✓✓✓) | T | T | T | T | T |
| LFW | VGG16 | W (✓✓✓) | L (×××) | T | T | W (✓✓✓) | L (×××) |
| | LeNet-5 | W (✓✓✓) | L (×××) | W (✓✓✓) | W (✓✓✓) | W (✓✓✓) | L (×××) |
| CelebA | VGG16 | W (✓✓✓) | W (✓✓✓) | T | T | W (✓✓✓) | W (✓✓✓) |
| | LeNet-5 | W (✓✓✓) | T | T | L (×××) | L (×××) | L (×××) |

| Dataset | Model | FairFLRep$_{SPD}$ vs Arachne | | FairFLRep$_{SPD}$ vs FairArachne$_{SPD}$ | | FairFLRep$_{SPD}$ vs FairMOON$_{SPD}$ | |
|---|---|---|---|---|---|---|---|
| | | SPD | Accuracy | SPD | Accuracy | SPD | Accuracy |
| UTKFace | VGG16 | W (✓✓✓) | L (×××) | T | T | T | W (✓✓✓) |
| | LeNet-5 | W (✓✓✓) | L (×××) | W (✓✓✓) | T | W (✓✓✓) | L (×××) |
| Fair Face | VGG16 | W (✓✓✓) | W (✓✓) | T | T | L (×××) | L (×××) |
| | LeNet-5 | W (✓✓✓) | L (×××) | W (✓) | T | T | L (×××) |
| LFW | VGG16 | W (✓✓✓) | L (×××) | T | T | W (✓✓✓) | L (×××) |
| | LeNet-5 | W (✓✓✓) | L (×××) | L (×××) | W (✓✓✓) | W (✓✓✓) | L (×××) |
| CelebA | VGG16 | W (✓✓) | T | T | T | W (✓✓✓) | W (✓✓✓) |
| | LeNet-5 | L (×××) | W (✓✓✓) | T | T | W (✓✓✓) | L (×××) |

| Dataset | Model | FairFLRep$_{DI}$ vs Arachne | | FairFLRep$_{DI}$ vs FairArachne$_{DI}$ | | FairFLRep$_{DI}$ vs FairMOON$_{DI}$ | |
|---|---|---|---|---|---|---|---|
| | | DI | Accuracy | DI | Accuracy | DI | Accuracy |
| UTKFace | VGG16 | W (✓✓✓) | L (×××) | T | T | T | W (✓✓✓) |
| | LeNet-5 | W (✓✓✓) | L (×××) | W (✓✓✓) | T | W (✓✓✓) | L (×××) |
| Fair Face | VGG16 | W (✓✓✓) | W (✓✓) | T | T | L (×××) | L (×××) |
| | LeNet-5 | W (✓✓✓) | L (××) | T | T | T | T |
| LFW | VGG16 | W (✓✓✓) | L (×××) | T | T | W (✓✓✓) | L (×××) |
| | LeNet-5 | W (✓✓✓) | L (×××) | L (×××) | W (✓✓✓) | W (✓✓✓) | L (×××) |
| CelebA | VGG16 | L (×××) | W (✓✓) | T | T | W (✓✓✓) | W (✓✓✓) |
| | LeNet-5 | L (×××) | W (✓✓✓) | T | T | W (✓✓✓) | T |

| Dataset | Model | FairFLRep$_{FPR}$ vs Arachne | | FairFLRep$_{FPR}$ vs FairArachne$_{FPR}$ | | FairFLRep$_{FPR}$ vs FairMOON$_{FPR}$ | |
|---|---|---|---|---|---|---|---|
| | | FPR | Accuracy | FPR | Accuracy | FPR | Accuracy |
| UTKFace | VGG16 | W (✓✓✓) | L (×××) | T | T | W (✓✓✓) | T |
| | LeNet-5 | W (✓✓✓) | L (×××) | W (✓) | T | W (✓✓✓) | L (×××) |
| Fair Face | VGG16 | W (✓✓✓) | W (✓✓✓) | T | T | T | T |
| | LeNet-5 | W (✓✓✓) | L (××) | T | T | W (✓✓✓) | T |
| LFW | VGG16 | W (✓✓✓) | L (×××) | T | L (××) | W (✓✓✓) | L (×××) |
| | LeNet-5 | W (✓✓✓) | L (×××) | W (✓✓✓) | W (✓✓✓) | W (✓✓✓) | L (×××) |
| CelebA | VGG16 | W (✓✓✓) | W (✓✓) | T | T | W (✓✓✓) | W (✓✓✓) |
| | LeNet-5 | W (✓✓✓) | W (✓✓✓) | T | T | T | T |
| TOTAL | Win | 29 | 10 | 7 | 4 | 15 | 7 |
| | Tie | 0 | 3 | 23 | 26 | 7 | 7 |
| | Loss | 3 | 19 | 2 | 2 | 4 | 18 |

Table 3. RQ1 and RQ2 – SettDiff – Image datasets – Statistical comparison between FairFLRep$_\mathcal{F}$ and the baselines (Same notation used in Table 2)

| Dataset | Model | FairFLRep$_{EOD}$ vs Arachne EOD | Accuracy | FairFLRep$_{EOD}$ vs FairArachne$_{EOD}$ EOD | Accuracy | FairFLRep$_{EOD}$ vs FairMOON$_{EOD}$ EOD | Accuracy |
|---|---|---|---|---|---|---|---|
| UTKFace | VGG16 | W (✓✓✓) | L (✗✗✗) | W (✓✓✓) | W (✓✓✓) | W (✓✓✓) | W (✓✓✓) |
| | LeNet-5 | W (✓✓✓) | L (✗✗✗) | T | W (✓✓✓) | W (✓✓✓) | L (✗✗✗) |
| FairFace | VGG16 | W (✓✓✓) | W (✓✓) | T | T | T | T |
| | LeNet-5 | W (✓✓✓) | W (✓✓✓) | T | T | T | L (✗✗✗) |
| CelebA | VGG16 | W (✓✓✓) | T | W (✓✓✓) | W (✓✓✓) | W (✓✓✓) | W (✓✓✓) |
| | LeNet-5 | W (✓✓✓) | W (✓✓✓) | W (✓✓✓) | W (✓✓✓) | W (✓✓✓) | W (✓✓✓) |

| Dataset | Model | FairFLRep$_{SPD}$ vs Arachne SPD | Accuracy | FairFLRep$_{SPD}$ vs FairArachne$_{SPD}$ SPD | Accuracy | FairFLRep$_{SPD}$ vs FairMOON$_{SPD}$ SPD | Accuracy |
|---|---|---|---|---|---|---|---|
| UTKFace | VGG16 | W (✓✓✓) | L (✗✗✗) | L (✗✗✗) | W (✓✓✓) | L (✗✗✗) | W (✓✓✓) |
| | LeNet-5 | W (✓✓✓) | L (✗✗✗) | T | T | W (✓✓✓) | L (✗✗✗) |
| FairFace | VGG16 | W (✓✓✓) | W (✓✓) | T | T | L (✗✗✗) | T |
| | LeNet-5 | W (✓✓✓) | L (✗✗✗) | T | L (✗✗✗) | T | L (✗✗✗) |
| CelebA | VGG16 | W (✓✓✓) | T | W (✓✓✓) | W (✓✓✓) | W (✓✓✓) | W (✓✓✓) |
| | LeNet-5 | W (✓✓✓) | W (✓✓✓) | W (✓✓✓) | W (✓✓) | W (✓✓✓) | W (✓✓✓) |

| Dataset | Model | FairFLRep$_{DI}$ vs Arachne DI | Accuracy | FairFLRep$_{DI}$ vs FairArachne$_{DI}$ DI | Accuracy | FairFLRep$_{DI}$ vs FairMOON$_{DI}$ DI | Accuracy |
|---|---|---|---|---|---|---|---|
| UTKFace | VGG16 | W (✓✓✓) | L (✗✗✗) | L (✗✗✗) | W (✓✓✓) | L (✗✗✗) | W (✓✓✓) |
| | LeNet-5 | W (✓✓✓) | L (✗✗✗) | T | T | W (✓✓✓) | L (✗✗✗) |
| FairFace | VGG16 | W (✓✓✓) | W (✓✓) | T | T | L (✗✗✗) | T |
| | LeNet-5 | W (✓✓✓) | L (✗✗✗) | T | L (✗✗) | T | L (✗✗✗) |
| CelebA | VGG16 | W (✓✓✓) | W (✓✓) | W (✓✓✓) | W (✓✓✓) | W (✓✓✓) | W (✓✓✓) |
| | LeNet-5 | W (✓✓✓) | W (✓✓✓) | W (✓✓✓) | W (✓✓) | W (✓✓✓) | T |

| Dataset | Model | FairFLRep$_{FPR}$ vs Arachne FPR | Accuracy | FairFLRep$_{FPR}$ vs FairArachne$_{FPR}$ FPR | Accuracy | FairFLRep$_{FPR}$ vs FairMOON$_{FPR}$ FPR | Accuracy |
|---|---|---|---|---|---|---|---|
| UTKFace | VGG16 | W (✓✓✓) | L (✗✗✗) | W (✓✓✓) | L (✗✗✗) | W (✓✓✓) | T |
| | LeNet-5 | W (✓✓✓) | L (✗✗✗) | W (✓✓✓) | T | W (✓✓✓) | L (✗✗✗) |
| FairFace | VGG16 | W (✓✓✓) | W (✓✓) | T | T | W (✓✓✓) | T |
| | LeNet-5 | W (✓✓✓) | L (✗✗✗) | T | L (✗✗) | T | L (✗✗✗) |
| CelebA | VGG16 | T | T | W (✓✓✓) | W (✓✓✓) | W (✓✓✓) | W (✓✓✓) |
| | LeNet-5 | W (✓✓✓) | W (✓✓✓) | W (✓✓✓) | W (✓✓✓) | W (✓✓✓) | W (✓✓✓) |

| TOTAL | Win | 23 | 10 | 11 | 12 | 15 | 10 |
| | Tie | 1 | 3 | 11 | 8 | 5 | 6 |
| | Loss | 0 | 11 | 2 | 4 | 4 | 8 |

Compared to Arachne, FairFLRep achieved 23 fairness comparisons across the three datasets and both models, without a single loss. However, FairFLRep falls short in overall accuracy in 11 cases, compared to 10 wins, reflecting Arachne's focus on accuracy rather than fairness.

Against FairArachne, FairFLRep achieves 11 wins in fairness, 11 ties, and only 2 losses. For accuracy, FairFLRep also wins more often (12 wins vs. 4 losses), showing it not only improves fairness more effectively but also retains or improves model accuracy in many cases. The higher number of fairness ties (11) compared to accuracy ties (8) highlights the consistent fairness performance of FairFLRep across models and metrics.

Compared to FairMOON, FairFLRep achieves 15 fairness wins, 5 ties, and 4 losses. Most losses occur on *DI* and *SPD* for the VGG16 model on the UTKFace and FairFace datasets, indicating FairMOON's effectiveness is limited to specific dataset–model configurations. In terms of accuracy, FairMOON outperforms FairFLRep in 8 out of 32 comparisons. However, these gains are offset by sharp fairness degradations—especially on `UTKFace` and `CelebA` —where `FairMOON` severely worsens the original model's fairness. These trade-offs, which are not observed in `FairFLRep`, suggest that `FairMOON` may prioritize accuracy at the expense of fairness, leading to inconsistent and potentially unstable repair outcomes.

> **Answer to RQ1 and RQ2 (`SettDiff` – Image datasets).** `FairFLRep` excels at reducing bias across a range of datasets and models, although it occasionally sacrifices some accuracy, especially in comparison to `Arachne` and `FairMOON`. The high number of fairness ties and wins, along with competitive accuracy results, suggests that `FairFLRep` effectively balances fairness and accuracy, confirming its value as a robust fairness repair strategy in complex image-based DNNs.

*6.3.3 `SettDiff` – Tabular datasets.* Table 4 summarizes how `FairFLRep` performs on tabular datasets compared to four baselines under the `SettDiff` setting. The notation is as in Table 2.

According to the results, `Arachne` and `LIMI` exhibit performance patterns that contrast with `FairFLRep` in complementary ways. Compared to `Arachne`, `FairFLRep` shows a stronger ability to reduce bias, achieving 26 fairness wins and no fairness losses. However, this improvement in fairness is accompanied by a slightly weaker or more variable performance in accuracy, with 10 losses to `Arachne` in the accuracy metric.

In comparison to `LIMI`, `FairFLRep` again demonstrates clear superiority in fairness, securing 23 fairness wins and zero losses. However, `FairFLRep` consistently underperforms in accuracy against `LIMI`, losing all 22 comparisons in accuracy. This tradeoff further illustrates `LIMI`'s emphasis on preserving classification performance, whereas `FairFLRep` prioritizes fairness repair.

When evaluated against `FairArachne`, `FairFLRep` continues to outperform across fairness metrics, achieving 18 fairness wins with only 1 loss, showcasing its effectiveness in reducing bias. Accuracy-wise, the performance is largely balanced, with 11 tie cases, reflecting `FairFLRep`'s ability to enhance fairness without drastically compromising model performance.

In comparison to `FairMOON`, the results show that while `FairFLRep` achieves 17 fairness wins and only 2 losses, `FairMOON` maintains a competitive stance in terms of accuracy, also resulting in 10 tie cases. This suggests that `FairMOON`, while less consistent in fairness outcomes, may retain or improve model accuracy in certain scenarios.

> **Answer to RQ1 and RQ2 (`SettDiff` – Tabular datasets).** Overall, `FairFLRep` demonstrates a robust advantage in fairness repair across tabular datasets, while maintaining competitive performance in accuracy. These results underscore `FairFLRep`'s generalizability across both fairness metrics and datasets.

## 6.4 RQ3 – Which is the "best" fairness metric for repair?

The different fairness metrics capture different aspects of fairness, and several recent studies [6, 12–14, 57] have shown that it is almost impossible to satisfy multiple notions of fairness simultaneously. Some fairness mitigation techniques may excel at improving one metric, such as *SPD*, while being less effective at improving another, like *DI*. This suggests that selecting the appropriate fairness metric is crucial, depending on the specific fairness concerns of the application under test. Moreover, improving one fairness metric may not necessarily lead to proportional improvements across all metrics. For instance, a method that significantly reduces *SPD* might show limited improvement in

Table 4. RQ1 and RQ2 – SettDiff – Tabular datasets – Statistical comparison between FairFLRep$_\mathcal{F}$ and the baselines (Same notation used in Table 2)

| Dataset | Model | FairFLRep$_{EOD}$ vs Arachne EOD | Accuracy | FairFLRep$_{EOD}$ vs FairArachne$_{EOD}$ EOD | Accuracy | FairFLRep$_{EOD}$ vs FairMOON$_{EOD}$ EOD | Accuracy | FairFLRep$_{EOD}$ vs LIMI EOD | Accuracy |
|---|---|---|---|---|---|---|---|---|---|
| Adult | Gender | W(✓✓✓) | T | T | T | T | T | W(✓✓✓) | L(×××) |
| | Race | W(✓✓✓) | T | T | T | T | T | W(✓✓✓) | L(×××) |
| COMPAS | Gender | W(✓✓✓) | L(×××) | W(✓✓✓) | T | W(✓✓✓) | T | W(✓✓✓) | L(×××) |
| | Race | W(✓✓✓) | L(×××) | W(✓✓✓) | W(✓✓✓) | W(✓✓✓) | W(✓✓✓) | W(✓✓✓) | L(×××) |
| Student | Gender | T | T | L(×××) | T | L(×××) | T | | |
| MEPS | Gender | W(✓✓✓) | T | W(✓✓✓) | T | W(✓✓✓) | T | T | T |
| | Race | W(✓✓✓) | L(×××) | W(✓✓✓) | L(×××) | W(✓✓✓) | L(×××) | W(✓✓✓) | L(×××) |

| Dataset | Model | FairFLRep$_{SPD}$ vs Arachne SPD | Accuracy | FairFLRep$_{SPD}$ vs FairArachne$_{SPD}$ SPD | Accuracy | FairFLRep$_{SPD}$ vs FairMOON$_{SPD}$ SPD | Accuracy | FairFLRep$_{SPD}$ vs LIMI SPD | Accuracy |
|---|---|---|---|---|---|---|---|---|---|
| Adult | Gender | W(✓✓✓) | T | T | T | T | T | W(✓✓✓) | L(×××) |
| | Race | W(✓✓✓) | T | W(✓✓✓) | T | W(✓✓✓) | L(×××) | W(✓✓✓) | L(×××) |
| COMPAS | Gender | W(✓✓✓) | L(×××) | T | T | T | T | W(✓✓✓) | L(×××) |
| | Race | W(✓✓✓) | L(×××) | T | W(✓✓✓) | T | W(✓✓✓) | W(✓✓✓) | L(×××) |
| Student | Gender | W(✓✓✓) | L(×××) | W(✓✓✓) | L(×××) | W(✓✓✓) | L(×××) | | |
| MEPS | Gender | W(✓✓✓) | T | W(✓✓✓) | T | W(✓✓✓) | T | W(✓✓✓) | L(×××) |
| | Race | W(✓✓✓) | L(×××) | W(✓✓✓) | L(×××) | W(✓✓✓) | L(×××) | W(✓✓✓) | L(×××) |

| Dataset | Model | FairFLRep$_{DI}$ vs Arachne DI | Accuracy | FairFLRep$_{DI}$ vs FairArachne$_{DI}$ DI | Accuracy | FairFLRep$_{DI}$ vs FairMOON$_{DI}$ DI | Accuracy | FairFLRep$_{DI}$ vs LIMI DI | Accuracy |
|---|---|---|---|---|---|---|---|---|---|
| Adult | Gender | W(✓✓✓) | T | T | T | T | L(×××) | W(✓✓✓) | L(×××) |
| | Race | W(✓✓✓) | T | W(✓✓✓) | T | T | L(×××) | W(✓✓✓) | L(×××) |
| COMPAS | Gender | W(✓✓✓) | L(×××) | T | L(×××) | L(×××) | T | W(✓✓✓) | L(×××) |
| | Race | W(✓✓✓) | L(×××) | T | W(✓✓✓) | T | W(✓✓✓) | W(✓✓✓) | L(×××) |
| Student | Gender | W(✓✓✓) | L(×××) | W(✓✓✓) | L(×××) | W(✓✓✓) | L(×××) | | |
| MEPS | Gender | W(✓✓✓) | L(×××) | W(✓✓✓) | L(×××) | W(✓✓✓) | L(×××) | W(✓✓✓) | L(×××) |
| | Race | W(✓✓✓) | L(×××) | W(✓✓✓) | L(×××) | W(✓✓✓) | L(×××) | W(✓✓✓) | L(×××) |

| Dataset | Model | FairFLRep$_{FPR}$ vs Arachne FPR | Accuracy | FairFLRep$_{FPR}$ vs FairArachne$_{FPR}$ FPR | Accuracy | FairFLRep$_{FPR}$ vs FairMOON$_{FPR}$ FPR | Accuracy | FairFLRep$_{FPR}$ vs LIMI FPR | Accuracy |
|---|---|---|---|---|---|---|---|---|---|
| Adult | Gender | W(✓✓✓) | T | W(✓✓✓) | W(✓✓✓) | W(✓✓✓) | L(×××) | W(✓✓✓) | L(×××) |
| | Race | W(✓✓✓) | T | W(✓✓✓) | W(✓✓✓) | W(✓✓✓) | L(×××) | W(✓✓✓) | L(×××) |
| COMPAS | Gender | W(✓✓✓) | L(×××) | W(✓✓✓) | L(×××) | W(✓✓✓) | L(×××) | W(✓✓✓) | L(×××) |
| | Race | W(✓✓✓) | L(×××) | W(✓✓✓) | L(×××) | W(✓✓✓) | L(×××) | W(✓✓✓) | L(×××) |
| Student | Gender | T | L(×××) | T | L(×××) | T | L(×××) | | |
| MEPS | Gender | W(✓✓✓) | T | W(✓✓✓) | W(✓✓✓) | W(✓✓✓) | W(✓✓✓) | W(✓✓✓) | T |
| | Race | W(✓✓✓) | L(×××) | W(✓✓✓) | T | W(✓✓✓) | T | W(✓✓✓) | L(×××) |
| TOTAL | Win | 26 | 0 | 18 | 6 | 17 | 4 | 23 | 0 |
| | Tie | 2 | 3 | 7 | 11 | 9 | 10 | 1 | 2 |
| | Loss | 0 | 16 | 1 | 10 | 2 | 14 | 0 | 22 |

*DI* or *EOD*. This analysis will highlight the potential trade-offs between different fairness objectives when applying these techniques.

The intuition is that fairness-aware repair techniques should be designed to optimize the most appropriate fairness metric for the specific fairness problem, considering the context and trade-offs inherent in the decision-making domain. In this RQ, we use the term "best" to describe scenarios where optimizing one fairness metric also leads to improvements in another. Specifically, we investigate the impact of optimizing one fairness metric on others based on FairFLRep technique (e.g., how optimizing for *DI* affects *SPD*, *EOD*, and *FPR*). Consistently with RQ1 and RQ2, we evaluate performance in terms of both accuracy and fairness across 30 runs.

*6.4.1 SettSame – Image datasets.* In Figure 9, we visualize the results of this analysis, showing how optimizing for one fairness metric impacts the performance of other metrics. Each plot in a cell shows the distribution of fairness scores across the fairness metrics after repair.

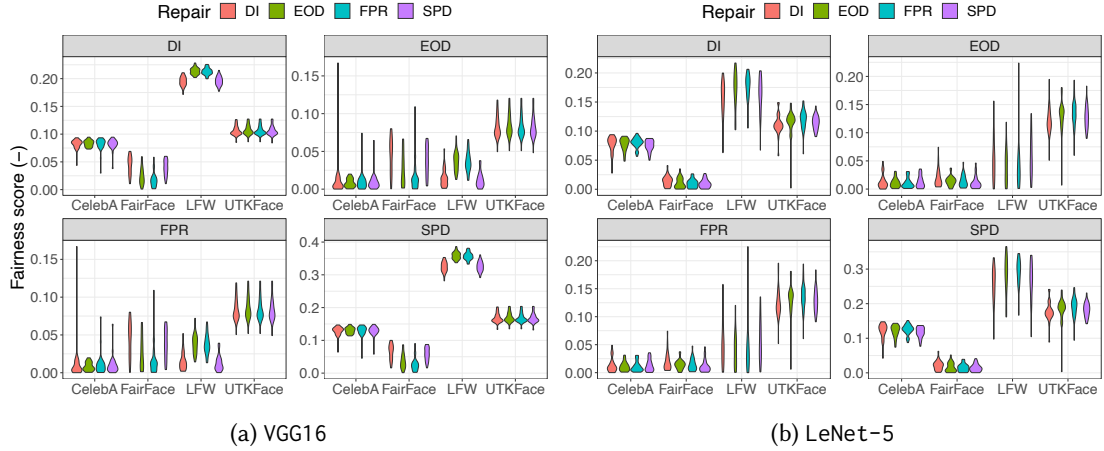

Fig. 9. RQ3 – `SettSame` – Image datasets – The violin plot visualizes the relationship between optimizing one fairness metric and its effect on other fairness metrics. Each plot in a cell shows the distribution of fairness scores across different fairness metrics after the repair process is applied.

- The vertical axis represents the fairness score, where lower values indicate a more favorable fairness outcome (i.e., less bias).
- Each cell corresponds to the distribution of a fairness score (shown in the header of the cell) after optimizing for a specific fairness metric (identified by the color of the violin plot), across different datasets. The width of the violin plot shows the density of scores at that level, with wider sections indicating more frequent occurrences.

Based on the patterns and distributions of the violin plots in Figure 9, we can gain insights into the relationships between various fairness metrics.

*Relationships Between Fairness Metrics:* The results in Figure 9 indicate that fairness metrics like *DI* and *SPD*, or *FPR* and *EOD*, often have similarly shaped distributions across datasets due to their conceptual relationships. For instance, *DI* and *SPD* both address disparities in positive outcomes across groups, which explains why their violin plots often show similar distributions. Likewise, *EOD* requires equalization of *FPR* and false negative rates (*FNR*) across groups, so changes in *EOD* naturally impact *FPR*. This conceptual overlap between these metrics suggests that optimizing one fairness metric may lead to improvements in others, as verified by the statistical test in Table 5.

*Consistency Across Datasets:* Comparing the results across different datasets, the distributions of *DI* and *SPD*, as well as *EOD* and *FPR*, are consistently similar. This suggests that the relationship between these metrics remains stable across datasets, with no substantial variability. Therefore, improvements in one fairness metric, such as *DI*, are consistently reflected in the other (*SPD*), and the same holds true for *EOD* and *FPR* across all datasets. This consistency reinforces the conceptual overlap between these fairness metrics across various datasets.

*Statistical analysis.* Table 5 provides a detailed analysis of the statistical comparison between different fairness metrics and model accuracy when the `FairFLRep` repair technique is applied across the four datasets and the two models. Each row reports the four versions of `FairFLRep` instantiated with the four fairness metrics (`FairFLRep`$_{EOD}$, `FairFLRep`$_{SPD}$, `FairFLRep`$_{DI}$, `FairFLRep`$_{FPR}$); similarly, the last four columns also report the four versions of `FairFLRep`. Each cell reports the statistical comparison between the repair approach on the row and the one on the column (e.g., `FairFLRep`$_{EOD}$ vs. `FairFLRep`$_{SPD}$); the comparison is performed in terms of the fairness metric

Table 5. RQ3 – SettSame – Image datasets – Statistical comparison between the performance of fairness metrics used in FairFLRep$_\mathcal{F}$. W: the fairness metric on the row wins against the fairness metric on the column; T: no statistical significant difference; L: the fairness metric on the row loses against the fairness metric on the column. The number of symbols ✓ and × represents the difference magnitude (1: small, 2: medium, 3: large).

| Dataset | Model | Repair | FairFLRep$_{EOD}$ EOD | FairFLRep$_{EOD}$ Accuracy | FairFLRep$_{SPD}$ SPD | FairFLRep$_{SPD}$ Accuracy | FairFLRep$_{DI}$ DI | FairFLRep$_{DI}$ Accuracy | FairFLRep$_{FPR}$ FPR | FairFLRep$_{FPR}$ Accuracy |
|---|---|---|---|---|---|---|---|---|---|---|
| UTKFace | VGG16 | FairFLRep$_{EOD}$ | - | - | T | T | T | T | T | T |
| | | FairFLRep$_{SPD}$ | T | T | - | - | T | T | T | T |
| | | FairFLRep$_{DI}$ | T | T | T | T | - | - | T | T |
| | | FairFLRep$_{FPR}$ | T | T | T | T | T | T | - | - |
| | LeNet-5 | FairFLRep$_{EOD}$ | - | - | T | T | T | T | T | T |
| | | FairFLRep$_{SPD}$ | T | T | - | - | T | T | T | T |
| | | FairFLRep$_{DI}$ | T | T | T | T | - | - | W (✓✓) | T |
| | | FairFLRep$_{FPR}$ | T | T | T | T | L (××) | T | - | - |
| FairFace | VGG16 | FairFLRep$_{EOD}$ | - | - | W (✓✓✓) | T | W (✓✓✓) | T | T | T |
| | | FairFLRep$_{SPD}$ | L (××) | T | - | - | T | T | L (×××) | T |
| | | FairFLRep$_{DI}$ | L (×××) | T | T | T | - | - | L (×××) | L (××) |
| | | FairFLRep$_{FPR}$ | T | T | W (✓✓✓) | T | W (✓✓✓) | W (✓✓) | - | - |
| | LeNet-5 | FairFLRep$_{EOD}$ | - | - | T | T | T | T | T | T |
| | | FairFLRep$_{SPD}$ | T | T | - | - | W (✓✓) | T | T | T |
| | | FairFLRep$_{DI}$ | T | T | L (××) | T | - | - | T | T |
| | | FairFLRep$_{FPR}$ | T | T | T | T | W (✓✓) | T | - | - |
| LFW | VGG16 | FairFLRep$_{EOD}$ | - | - | L (×××) | W (✓✓✓) | L (×××) | W (✓✓✓) | T | T |
| | | FairFLRep$_{SPD}$ | W (✓✓✓) | L (×××) | - | - | T | T | W (✓✓✓) | L (×××) |
| | | FairFLRep$_{DI}$ | W (✓✓✓) | L (×××) | T | T | - | - | W (✓✓✓) | L (×××) |
| | | FairFLRep$_{FPR}$ | T | T | L (×××) | W (✓✓✓) | L (×××) | W (✓✓✓) | - | - |
| | LeNet-5 | FairFLRep$_{EOD}$ | - | - | T | T | T | T | T | T |
| | | FairFLRep$_{SPD}$ | T | T | - | - | T | T | L (××) | T |
| | | FairFLRep$_{DI}$ | T | T | T | T | - | - | T | L (××) |
| | | FairFLRep$_{FPR}$ | T | T | T | T | L (××) | W (✓✓) | - | - |
| CelebA | VGG16 | FairFLRep$_{EOD}$ | - | - | T | T | T | T | T | T |
| | | FairFLRep$_{SPD}$ | T | T | - | - | T | T | T | T |
| | | FairFLRep$_{DI}$ | T | T | T | T | - | - | T | T |
| | | FairFLRep$_{FPR}$ | T | T | T | T | T | T | - | - |
| | LeNet-5 | FairFLRep$_{EOD}$ | - | - | T | T | T | T | T | T |
| | | FairFLRep$_{SPD}$ | T | T | - | - | T | T | T | T |
| | | FairFLRep$_{DI}$ | T | T | T | T | - | - | T | T |
| | | FairFLRep$_{FPR}$ | T | T | T | T | T | T | - | - |
| | | | W/T/L | W/T/L | W/T/L | W/T/L | W/T/L | W/T/L | W/T/L | W/T/L |
| Summary | | FairFLRep$_{EOD}$ | - | - | 1/6/1 | 1/7/0 | 1/6/1 | 1/7/0 | 0/8/0 | 0/8/0 |
| | | FairFLRep$_{SPD}$ | 1/6/1 | 0/7/1 | - | - | 1/7/0 | 0/8/0 | 1/5/2 | 0/7/1 |
| | | FairFLRep$_{DI}$ | 1/6/1 | 0/7/1 | 0/7/1 | 0/8/0 | - | - | 2/5/1 | 0/5/3 |
| | | FairFLRep$_{FPR}$ | 0/8/0 | 0/8/0 | 1/6/1 | 1/7/0 | 1/4/3 | 3/5/0 | - | - |
| TOTAL | | Win | 2 | 0 | 2 | 2 | 4 | 4 | 3 | 0 |
| | | Tie | 20 | 22 | 19 | 22 | 16 | 20 | 18 | 20 |
| | | Loss | 2 | 2 | 3 | 0 | 4 | 0 | 3 | 4 |

of the approach on the column (e.g., *SPD*) and in terms of accuracy. The possible results of the comparison of FairFLRep$_{\mathcal{F}_1}$ and FairFLRep$_{\mathcal{F}_2}$ in terms of $\mathcal{F}_2$ are as follows:

W (*Win*): it indicates that FairFLRep$_{\mathcal{F}_1}$ significantly improves $\mathcal{F}_2$ significantly better than FairFLRep$_{\mathcal{F}_2}$.
L (*Loss*): it indicates that FairFLRep$_{\mathcal{F}_1}$ is significantly worse at improving $\mathcal{F}_2$ compared to FairFLRep$_{\mathcal{F}_2}$.
T (*Tie*): it indicates that FairFLRep$_{\mathcal{F}_1}$ and FairFLRep$_{\mathcal{F}_2}$ achieve similar results in terms of $\mathcal{F}_2$.

The comparison in terms of accuracy, instead, tells which approach better maintains or improves model accuracy. A W (Win) or L (Loss) here suggests a substantial impact on accuracy.

At the bottom, the table provides a summary of the total wins, ties, and losses across each fairness metric used during repair. Based on the summary in Table 5, we can derive several key insights:

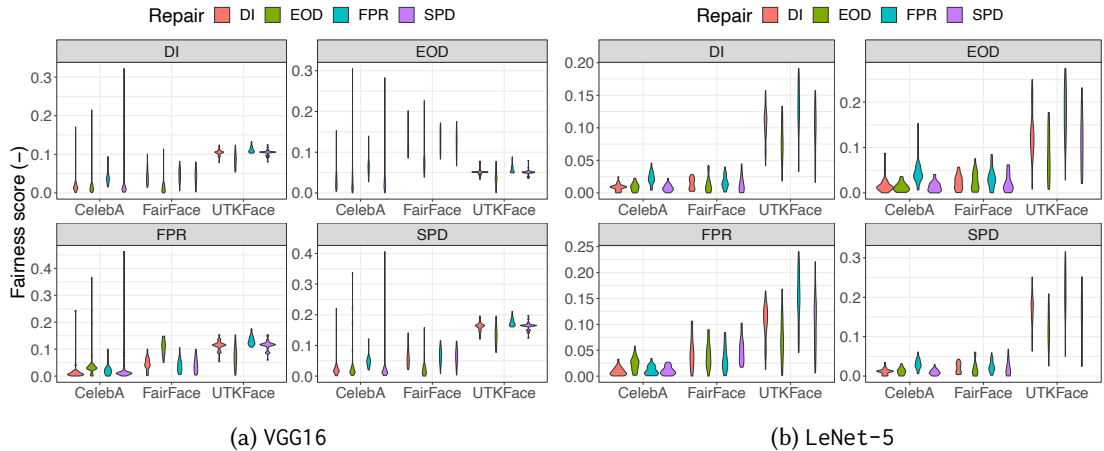

Fig. 10. RQ3 – `SettDiff` – Image datasets – Visualizing the relationship between optimizing for one fairness metrics on another based on the proposed `FairFLRep` repair technique.

*Fairness metrics are not independent:* Improving one fairness metric often impacts others, either positively or negatively. For example, optimizing *SPD* can result in improvements in *EOD* and *DI*, but may negatively affect *FPR*. This highlights that fairness metrics are interrelated.

*Trade-offs:* There are inherent trade-offs in fairness optimization. Some metrics may improve at the expense of others, as seen when optimizing *FPR* frequently results in losses in *DI* and *SPD*, especially in challenging datasets like `LFW`.

*Consistency across the dataset and model architectures:* Across the datasets, `UTKFace` and `CelebA` exhibit consistent patterns, where improvements in one fairness metric (like *SPD* or *DI*) often yield similar effects across others. In contrast, `FairFace` and `LFW` datasets show stronger trade-offs, particularly when optimizing *FPR*, with frequent losses in *DI* and *SPD*. Across the model architectures, both the `VGG16` and `LeNet-5` models exhibit trade-offs. However, `VGG16` tends to produce larger improvements in fairness but also larger trade-offs, while `LeNet-5` shows more stability, with more tied outcomes across fairness metrics. This suggests that complex models like `VGG16` may be more flexible but also more prone to variability in fairness gains and losses.

> **Answer to RQ3 (`SettSame` – Image datasets).** Our analysis confirms the conceptual overlap between metrics like *DI* and *SPD* or *EOD* and *FPR*, suggesting that optimizing one can lead to similar outcomes. However, optimizing one metric can also have both positive and negative effects on others, and the impact of fairness optimization depends heavily on the model architecture. There is no universally optimal fairness metric. The "best" metric depends on the specific fairness concerns in the given context. A "one-size-fits-all" approach to fairness is ineffective. Careful selection of the most relevant fairness metric is crucial for achieving the desired improvements, with trade-offs, especially in challenging metrics like *DI* and *FPR*.

*6.4.2* `SettDiff` *– Image datasets.* The violin plots shown in Figure 10 illustrate the relationship between using one fairness metric for repair and its effects on other fairness metrics across the three datasets (`CelebA`, `FairFace`, `UTKFace`) for the task of addressing race disparities (in `UTKFace` and `FairFace`) and age disparities (`CelebA`). By focusing on one fairness metric for repair (e.g., *DI* or *SPD*), the figure reveals how that repair impacts other metrics. For example, repairing *DI* may improve fairness measured by *DI* but have mixed effects on other metrics like *FPR* or *EOD*. This is demonstrated by the spread and shape of the violins for each dataset and fairness metric. The effect of the repair slightly differs across datasets and model being repaired, with some datasets showing more variability in fairness scores for certain metrics. The patterns and distributions of the violin plots in Figure 10 provide valuable insights into the relationships between various fairness metrics.

*Relationships Between Fairness Metrics:* The following relationships can be inferred from Figure 10: (1) Similarly to the observation for `SettSame`, the violin plots for *DI* and *SPD* across datasets tend to have similar shapes and distributions, indicating a potential positive correlation between these two metrics. This is expected, as both metrics focus on measuring group-level disparities. Since both address the rate at which groups receive favorable outcomes, improvements or declines in one metric are likely to be reflected in changes to the other. As seen by the relative consistency in their violin plot shapes, a fairness repair method that specifically targets *DI* tends to also improve *SPD* scores. (2) The violin plots for *FPR* and *EOD* show partial alignment but also notable differences across datasets. For instance, in some datasets like `CelebA`, there is closer alignment between *FPR* and *EOD*, indicating that reductions in false positive rates for one group can sometimes reflect as equal opportunity. However, in more challenging datasets like `FairFace`, the distributions for *FPR* and *EOD* diverge more, reflecting that improvements in one metric may not necessarily translate to improvements in the other. This variability was not as pronounced in `SettSame`, where the sensitive attribute is the same as the class label.

*Model complexity and fairness interactions:* The VGG16 model exhibits greater variability across datasets, reflected in the violin plot shapes. For example, repairing *DI* and *SPD* in VGG16 may cause larger fluctuations in *FPR* or *EOD*, reflecting the complex relationship between accuracy and group-level fairness. By contrast, `LeNet-5`'s simpler architecture results in more predictable fairness outcomes, where conceptually related metrics (*DI* and *SPD*, or *EOD* and *FPR*) show closer alignment. This makes `LeNet-5` less prone to large swings in fairness scores when one metric is optimized, unlike VGG16, which may exhibit more dramatic variations across metrics.

*Statistical analysis.* Table 6 reports the statistical analysis of how repairing a model with different fairness metrics affects other fairness metrics and the model's overall accuracy. Table 6 uses the same notation as Table 5. Based on the summary in Table 6, several key insights can be derived regarding the performance of `FairFLRep`.

*Impact of EOD on Other Metrics. EOD* performs relatively better in improving other fairness metrics during the repair process, as it often leads to wins on fairness metrics such as *SPD* and *DI*. However, it also shows significant losses in accuracy, indicating that while it improves fairness, there is a trade-off with model accuracy when *EOD* is used as the primary repair metric. This underscores the balance needed between *EOD* improvement and model accuracy.

*Relationships Between Fairness Metrics.* The consistent ties between *SPD* and *DI* across different datasets and models suggest a positive correlation or relationship between these two metrics. Both metrics are focused on group fairness and aim to reduce disparities in positive predictions across groups. This alignment in their goals explains why they tend to show similar performance, leading to frequent ties in the repair process. Optimizing for one could naturally improve the other, but their differences in measurement (*SPD* as a difference and *DI* as a ratio) may still lead to slight variations in certain scenarios.

Similarly, the ties between *EOD* and *FPR* suggest a certain degree of similarity or equivalence when it comes to their effects during the repair process. This makes sense, as both metrics aim to balance errors across groups—with *EOD* accounting for both *FPR* and *FNR*, and *FPR* specifically targeting false positives. While they often tie, *EOD* generally performs better than *FPR*, with *FPR* showing more losses. This indicates that *EOD* is a more effective overall fairness repair metric, likely due to its broader scope of balancing both false positive and false negative rates.

Table 6. RQ3 – `SettDiff` – Image datasets – Statistical comparison between the performance of fairness metrics used in `FairFLRep`$_\mathcal{F}$. Similar notation is used as in Table 5.

| Dataset | Model | Repair | FairFLRep$_{EOD}$ | | FairFLRep$_{SPD}$ | | FairFLRep$_{DI}$ | | FairFLRep$_{FPR}$ | |
|---|---|---|---|---|---|---|---|---|---|---|
| | | | EOD | Accuracy | SPD | Accuracy | DI | Accuracy | FPR | Accuracy |
| UTKFace | VGG16 | FairFLRep$_{EOD}$ | - | - | W(✓✓✓) | L(✗✗✗) | W(✓✓✓) | L(✗✗✗) | W(✓✓✓) | L(✗✗✗) |
| | | FairFLRep$_{SPD}$ | L(✗✗✗) | W(✓✓✓) | - | - | T | T | W(✓✓✓) | L(✗) |
| | | FairFLRep$_{DI}$ | L(✗✗✗) | W(✓✓✓) | T | T | - | - | W(✓✓✓) | T |
| | | FairFLRep$_{FPR}$ | L(✗✗✗) | W(✓✓✓) | L(✗✗✗) | W(✓) | L(✗✗✗) | T | - | - |
| | LeNet-5 | FairFLRep$_{EOD}$ | - | - | W(✓) | T | W(✓✓✓) | L(✗✗) | W(✓✓✓) | L(✗✗✗) |
| | | FairFLRep$_{SPD}$ | T | T | - | - | T | T | W(✓✓✓) | L(✗✗) |
| | | FairFLRep$_{DI}$ | L(✗✗) | W(✓✓) | T | T | - | - | W(✓✓✓) | T |
| | | FairFLRep$_{FPR}$ | L(✗✗✗) | W(✓✓✓) | L(✗✗✗) | W(✓✓) | L(✗✗✗) | T | - | - |
| FairFace | VGG16 | FairFLRep$_{EOD}$ | - | - | W(✓✓✓) | T | W(✓✓✓) | T | L(✗✗✗) | T |
| | | FairFLRep$_{SPD}$ | L(✗✗✗) | T | - | - | T | T | T | T |
| | | FairFLRep$_{DI}$ | L(✗✗✗) | T | T | T | - | - | T | T |
| | | FairFLRep$_{FPR}$ | L(✗✗✗) | T | T | T | T | T | - | - |
| | LeNet-5 | FairFLRep$_{EOD}$ | - | - | T | T | T | T | T | T |
| | | FairFLRep$_{SPD}$ | T | T | - | - | T | T | L(✗✗) | T |
| | | FairFLRep$_{DI}$ | T | T | T | T | - | - | T | T |
| | | FairFLRep$_{FPR}$ | T | T | T | T | T | T | - | - |
| CelebA | VGG16 | FairFLRep$_{EOD}$ | - | - | T | L(✗✗) | T | L(✗✗✗) | L(✗✗) | T |
| | | FairFLRep$_{SPD}$ | T | W(✓✓) | - | - | T | T | T | T |
| | | FairFLRep$_{DI}$ | T | W(✓✓✓) | T | T | - | - | W(✓✓) | T |
| | | FairFLRep$_{FPR}$ | L(✗✗✗) | T | L(✗✗✗) | T | L(✗✗✗) | T | - | - |
| | LeNet-5 | FairFLRep$_{EOD}$ | - | - | T | T | T | T | L(✗✗✗) | T |
| | | FairFLRep$_{SPD}$ | T | T | - | - | T | T | T | T |
| | | FairFLRep$_{DI}$ | T | T | T | T | - | - | T | T |
| | | FairFLRep$_{FPR}$ | L(✗✗✗) | T | L(✗✗✗) | T | L(✗✗✗) | T | - | - |
| Summary | | | W/T/L | W/T/L | W/T/L | W/T/L | W/T/L | W/T/L | W/T/L | W/T/L |
| | | FairFLRep$_{EOD}$ | - | - | 3/3/0 | 0/4/2 | 3/3/0 | 0/3/3 | 2/0/3 | 0/4/2 |
| | | FairFLRep$_{SPD}$ | 0/4/2 | 2/4/0 | - | - | 0/6/0 | 0/6/0 | 2/3/0 | 0/4/2 |
| | | FairFLRep$_{DI}$ | 0/3/3 | 3/3/0 | 0/6/0 | 0/6/0 | - | - | 3/3/0 | 0/6/0 |
| | | FairFLRep$_{FPR}$ | 0/1/5 | 2/4/0 | 0/2/4 | 2/4/0 | 0/2/4 | 0/6/0 | - | - |
| TOTAL | | Win | 0 | 7 | 3 | 2 | 2 | 0 | 7 | 0 |
| | | Tie | 8 | 11 | 11 | 14 | 11 | 15 | 6 | 14 |
| | | Loss | 10 | 0 | 4 | 2 | 4 | 3 | 3 | 4 |

> **Answer to RQ3 (`SettDiff` – Imaga datasets).** On the one hand, *EOD* seems to offer a strong ability to repair other fairness metrics, but the trade-off comes at the cost of accuracy loss. On the other hand, the relationship between *EOD* and *FPR* suggests that these two metrics might overlap in the fairness aspects they address, though *EOD* is generally a better performer. Moreover, the strong ties between *SPD* and *DI* indicate a positive relationship, where improving one metric might naturally lead to improvements in the other.

*6.4.3 `SettDiff` – Tabular datasets.* The violin plots in Figure 11 illustrate the relationship between optimizing a single fairness metric during repair and its downstream effects on other fairness metrics across the four tabular datasets under the `SettDiff` configuration. Specifically, Figure 11a focuses on *gender* disparities, while Figure 11b presents results for *race* disparities. Each violin plot represents the distribution of fairness scores when a specific metric is used as the target for fairness repair using `FairFLRep`.

Across both *gender* and *race* settings, the violin plots for *DI* and *SPD* tend to have similar shapes and overlapping ranges across all datasets, further reflecting the positive correlation, observed on repairing image datasets. The relationship between *EOD* and *FPR* is more nuanced. On some datasets (e.g., COMPAS and MEPS), repairing for *EOD* appears to also reduce *FPR* to some extent, indicating some alignment. However, in other datasets, such as Adult and Student, the plots diverge, reflecting dataset-specific trade-offs. This suggests that while both metrics relate to classification errors across groups, optimizing for one does not always guarantee improvement in the other.

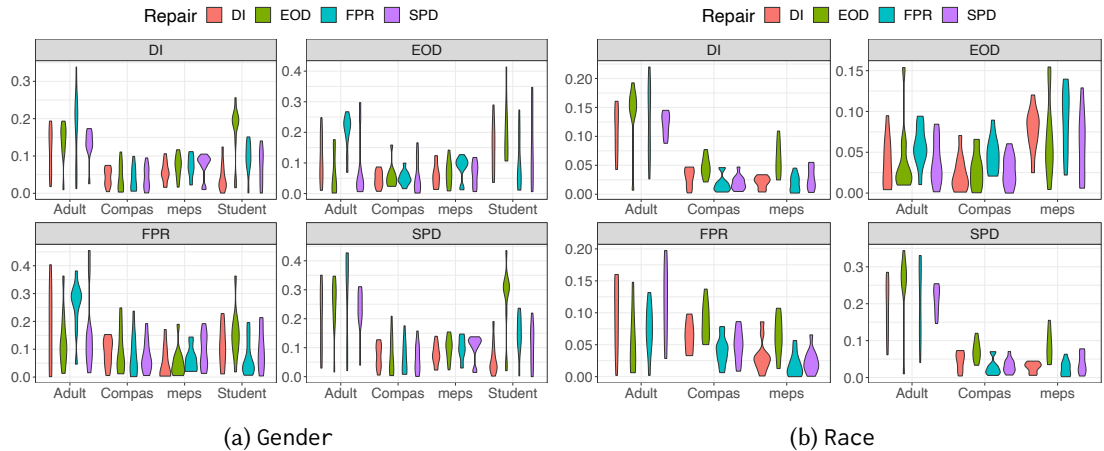

(a) Gender  (b) Race

Fig. 11. RQ3 – `SettDiff` – Tabular datasets – Visualizing the relationship between optimizing for one fairness metrics on another based on the proposed `FairFLRep` repair technique.

The plots also highlight that some datasets respond more stably to fairness repair. For instance, MEPS and Student show narrower violin plots, suggesting more predictable fairness outcomes. In contrast, datasets like COMPAS and Adult often exhibit wider distributions, indicating greater variability and thus more sensitivity to the choice of fairness metric.

*Statistical analysis.* Table 7 presents the statistical outcomes of using different fairness metrics in `FairFLRep` to repair models on the four tabular datasets under `SettDiff`. It evaluates each metric's generalizability, i.e., how repairing for one fairness objective affects other metrics and accuracy. Key insights from the table are as follows.

*Impact of EOD on other metrics.* `FairFLRep`$_{EOD}$ achieves 1 win, 16 ties, and 11 losses in fairness comparisons. The high number of ties reflect *EOD*'s relative neutrality. However, it loses four times to both *SPD* and *DI* and three times to *FPR*, indicating that `FairFLRep`$_{EOD}$ may not generalize well to other fairness metrics on tabular datasets. Regarding accuracy, `FairFLRep`$_{EOD}$ records 19 ties, with no wins or losses, further confirming its conservative behavior—it preserves accuracy but rarely outperforms in other dimensions.

`FairFLRep`$_{SPD}$ registers 1 fairness win (against *FPR*), 20 ties, and 0 losses, indicating stability across fairness dimensions. However, for accuracy, it shows 18 ties and 3 losses, suggesting that while stable in fairness, it may occasionally impact predictive performance.

`FairFLRep`$_{DI}$ Demonstrates the strongest overall performance. It achieves 19 ties, 2 fairness losses (to *EOD* and *FPR*), and 1 accuracy win (against *SPD*), with 19 ties and 1 loss (to *EOD*) in accuracy. These results indicate that `FairFLRep`$_{DI}$ generalizes well across fairness metrics and preserves model accuracy.

`FairFLRep`$_{FPR}$ shows 1 fairness win (against *EOD*), 17 ties, and 3 loses (2 to *DI* and 1 to *EOD*). In terms of accuracy, `FairFLRep`$_{FPR}$ achieves 1 win (against *SPD*), `FairFLRep`$_{FPR}$ ties, and no losses, indicating it is accuracy-preserving, though less effective in fairness generalization.

50  Moses Openja, Paolo Arcaini, Foutse Khomh, and Fuyuki IshikawaTable 7. RQ3 – SettDiff – Tabular datasets – Statistical comparison between the performance of fairness metrics used in FairFLRep$_\mathcal{F}$. Similar notation is used as in Table 5.

| Dataset | Model | Repair | FairFLRep$_{EOD}$ | | FairFLRep$_{SPD}$ | | FairFLRep$_{DI}$ | | FairFLRep$_{FPR}$ | |
|---|---|---|---|---|---|---|---|---|---|---|
| | | | EOD | Accuracy | SPD | Accuracy | DI | Accuracy | FPR | Accuracy |
| Adult | Gender | FairFLRep$_{EOD}$ | - | - | T | T | T | W(✓✓✓) | W(✓✓✓) | T |
| | | FairFLRep$_{SPD}$ | T | T | - | - | T | T | W(✓✓✓) | T |
| | | FairFLRep$_{DI}$ | L(×××) | L(×××) | T | T | - | - | T | T |
| | | FairFLRep$_{FPR}$ | L(×××) | T | T | T | L(×××) | T | - | - |
| | Race | FairFLRep$_{EOD}$ | - | - | L(×××) | W(✓✓✓) | L(×××) | T | T | T |
| | | FairFLRep$_{SPD}$ | T | L(×××) | - | - | T | T | T | T |
| | | FairFLRep$_{DI}$ | T | T | T | T | - | - | T | T |
| | | FairFLRep$_{FPR}$ | T | T | T | T | T | T | - | - |
| COMPAS | Gender | FairFLRep$_{EOD}$ | - | - | T | T | T | T | T | T |
| | | FairFLRep$_{SPD}$ | T | T | - | - | T | T | T | T |
| | | FairFLRep$_{DI}$ | T | T | T | T | - | - | T | T |
| | | FairFLRep$_{FPR}$ | T | T | T | T | T | T | - | - |
| | Race | FairFLRep$_{EOD}$ | - | - | L(×××) | T | L(×××) | T | L(×××) | T |
| | | FairFLRep$_{SPD}$ | T | T | - | - | T | L(×××) | T | L(×××) |
| | | FairFLRep$_{DI}$ | T | T | T | W(✓✓✓) | - | - | L(×××) | T |
| | | FairFLRep$_{FPR}$ | T | T | T | W(✓✓✓) | T | T | - | - |
| MEPS | Gender | FairFLRep$_{EOD}$ | - | - | T | T | T | T | T | T |
| | | FairFLRep$_{SPD}$ | T | T | - | - | T | T | T | T |
| | | FairFLRep$_{DI}$ | T | T | T | T | - | - | T | T |
| | | FairFLRep$_{FPR}$ | T | T | T | T | T | T | - | - |
| | Race | FairFLRep$_{EOD}$ | - | - | L(×××) | T | L(×××) | T | L(×××) | T |
| | | FairFLRep$_{SPD}$ | T | T | - | - | T | T | T | T |
| | | FairFLRep$_{DI}$ | T | T | T | T | - | - | T | T |
| | | FairFLRep$_{FPR}$ | T | T | T | T | T | T | - | - |
| Student | Gender | FairFLRep$_{EOD}$ | - | - | L(×××) | T | L(×××) | T | L(×××) | T |
| | | FairFLRep$_{SPD}$ | T | T | - | - | T | T | T | T |
| | | FairFLRep$_{DI}$ | T | T | T | T | - | - | T | T |
| | | FairFLRep$_{FPR}$ | W(✓✓✓) | T | T | T | L(×××) | T | - | - |
| Summary | | | W/T/L | W/T/L | W/T/L | W/T/L | W/T/L | W/T/L | W/T/L | W/T/L |
| | | FairFLRep$_{EOD}$ | - | - | 0/3/4 | 1/6/0 | 0/3/4 | 1/6/0 | 1/3/3 | 0/7/0 |
| | | FairFLRep$_{SPD}$ | 0/7/0 | 0/6/1 | - | - | 0/7/0 | 0/6/1 | 1/6/0 | 0/6/1 |
| | | FairFLRep$_{DI}$ | 0/6/1 | 0/6/1 | 0/7/0 | 1/6/0 | - | - | 0/6/1 | 0/7/0 |
| | | FairFLRep$_{FPR}$ | 1/5/1 | 0/7/0 | 0/7/0 | 1/6/0 | 0/5/2 | 0/7/0 | - | - |
| TOTAL | | Win | 1 | 0 | 0 | 3 | 0 | 1 | 2 | 0 |
| | | Tie | 18 | 19 | 17 | 18 | 15 | 19 | 15 | 20 |
| | | Loss | 2 | 2 | 4 | 0 | 6 | 1 | 4 | 1 |

*Relationships Between Fairness Metrics.* DI and SPD emerge as the most effective fairness metrics overall. Both demonstrate high stability across fairness comparisons and minimal impact on model accuracy, making them strong candidates for general-purpose fairness repair. *FPR* also shows promise, particularly in preserving accuracy, though its fairness performance is slightly less consistent. In contrast, *EOD*, which performed well on image datasets, shows reduced generalizability here–often producing fairness ties but failing to outperform other metrics. This contrast likely reflects the structured and lower-dimensional nature of tabular data, where group-level disparities are more directly addressed by metrics like *DI* and *SPD*. This demonstrates the highest overall stability and minimal trade-offs, especially in tabular data, making them good default choices for fairness repair when general fairness and accuracy preservation are both priorities.

> **Answer to RQ3 (SettDiff – Tabular datasets).** Interestingly, our statistical analysis reveals a contrast in which fairness metrics are most effective for fairness repair in tabular

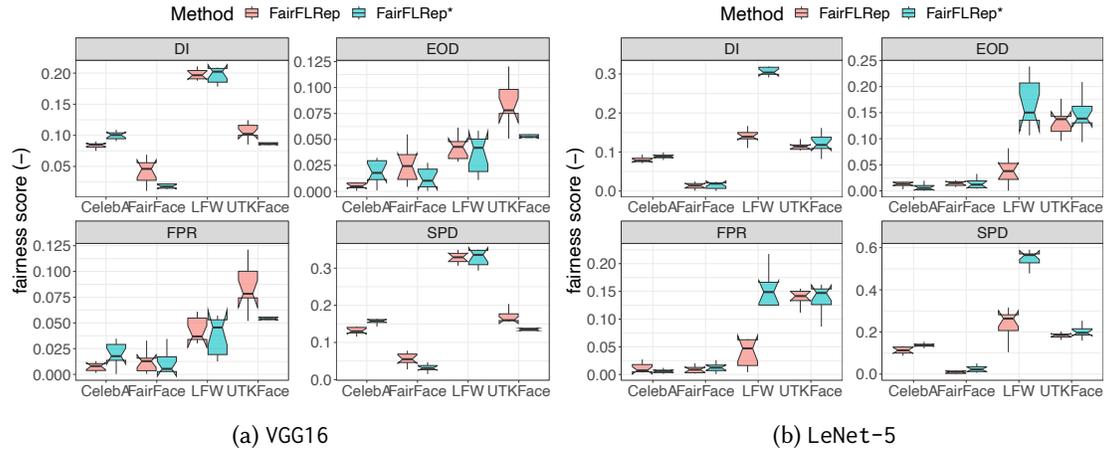

Fig. 12. RQ4 – SettSame – Image datasets – Fairness outcomes when repairing only the last layer (FairFLRep) versus deeper layers (FairFLRep∗)

> settings. In particular, FairFLRep$_{DI}$ and FairFLRep$_{SPD}$ demonstrate the strongest generalizability across other fairness metrics and minimal disruption to model accuracy. These metrics yield the highest number of fairness ties and fewest losses, making them robust and reliable choices for fairness repair in structured, lower-dimensional tabular data. FairFLRep$_{FPR}$ also performs reasonably well, especially in preserving accuracy. FairFLRep$_{EOD}$, which worked well in image data, is less effective here.

## 6.5 RQ4 – Is repairing only the last layer more effective than repairing other layers?

*6.5.1 SettSame – Image datasets.* Figure 12 compares the fairness outcomes of repairing only the last layer (FairFLRep) versus a deeper or intermediate layer (FairFLRep∗), across the four image datasets, in the SettSame setting. The corresponding classification accuracy is shown in Figure 13.

*Repairing LeNet-5.* Across all the four datasets, FairFLRep consistently outperforms FairFLRep∗ or performs comparably on most fairness metrics (see Figure 12b). Most notably, on the LFW dataset, FairFLRep significantly outperforms FairFLRep∗ across all four fairness metrics, demonstrating a clear advantage of last-layer repair in this case. For the other datasets (CelebA, FairFace, and UTKFace), FairFLRep achieves slightly better or equal fairness, particularly on *DI* and *FPR*, while differences in *SPD* and *EOD* remain minimal. These results suggest that for simpler architectures like LeNet-5, repairing only the last layer is not only sufficient, but can yield superior fairness outcomes, especially in datasets like LFW, where fairness disparities are more easily correctable at the output level.

In terms of accuracy (see Figure 13b), both methods perform similarly overall. However, on LFW, FairFLRep∗ slightly outperforms FairFLRep, aligning with earlier findings where FairFLRep showed a notable fairness gain. This reflects a rare trade-off where last-layer repair may overcorrect for fairness at a slight cost to accuracy—though this is limited to specific datasets.

*Repairing VGG16.* The fairness results for VGG16 (see Figure 12a) reveal more variation between FairFLRep and FairFLRep∗, especially in complex datasets. On UTKFace and FairFace, FairFLRep∗ slightly outperforms FairFLRep on certain metrics, indicating that deeper-layer repairs can offer marginal benefits in these configurations. However, FairFLRep remains competitive or better on metrics like *FPR* and *DI*, especially in CelebA and LFW, where its fairness scores are consistently lower (i.e., better). These results suggest that deeper repairs may offer some advantages in complex datasets, FairFLRep (last-layer repair) continues to be a strong and stable choice, particularly when minimal interventions are desired.

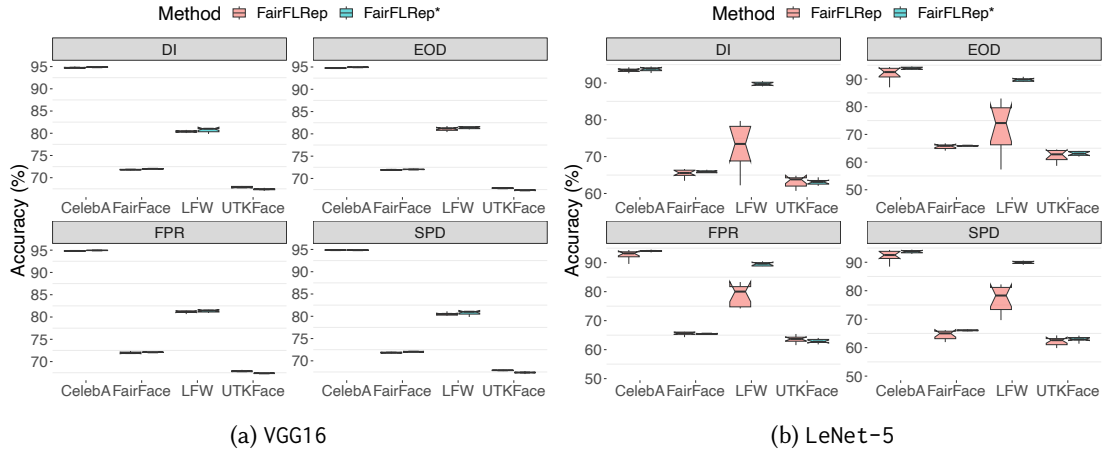

Fig. 13. RQ4 – SettSame – Image datasets – Classification accuracy of models repaired using FairFLRep versus FairFLRep∗

Comparing the accuracy for VGG16 (see Figure 13a), both FairFLRep and FairFLRep∗ yield nearly identical results across all the four datasets, with FairFLRep often matching or slightly exceeding FairFLRep∗. This result further confirms the effectiveness on FairFLRep as a practical and balanced fairness repair strategy.

> **Answer to RQ4 (SettSame – Image datasets).** Our analysis show that repairing only the last layer is often sufficient, and in some cases, yields superior fairness outcomes—particularly in simpler DNN architectures like LeNet-5. However, on LFW, this comes with a slight accuracy drop, suggesting a possible trade-off. For more complex models and datasets, such as VGG16, deeper-layer repairs may offer marginal advantages, but the overall improvements are dataset- and metric-dependent.

*6.5.2 SettDiff – Image datasets.* Figure 14 compares the fairness outcomes of repairing the last layer versus a intermediate layer under the SettDiff settings. Figure 15 reports the corresponding accuracy. As shown in the figures, across all datasets and metrics, FairFLRep performs comparably or better than FairFLRep∗.

*Repairing* LeNet-5. Particularly on CelebA and UTKFace, FairFLRep consistently achieves lower fairness scores across most metrics. On FairFace, FairFLRep and FairFLRep∗ perform similarly on *SPD* and *FPR*, while FairFLRep tends to do slightly better on *DI* and *EOD*. Notably, CelebA and UTKFace show the clearest advantage for FairFLRep, with visible reductions in disparity across all four fairness metrics. Overall, no significant gains are observed from deeper-layer repair (FairFLRep∗) in this setting, suggesting that bias in these configurations can often be corrected effectively at the output layer. Accuracy-wise (Figure 15b), both methods perform comparably across all datasets, with `FairFLRep` often matching `FairFLRep∗`.

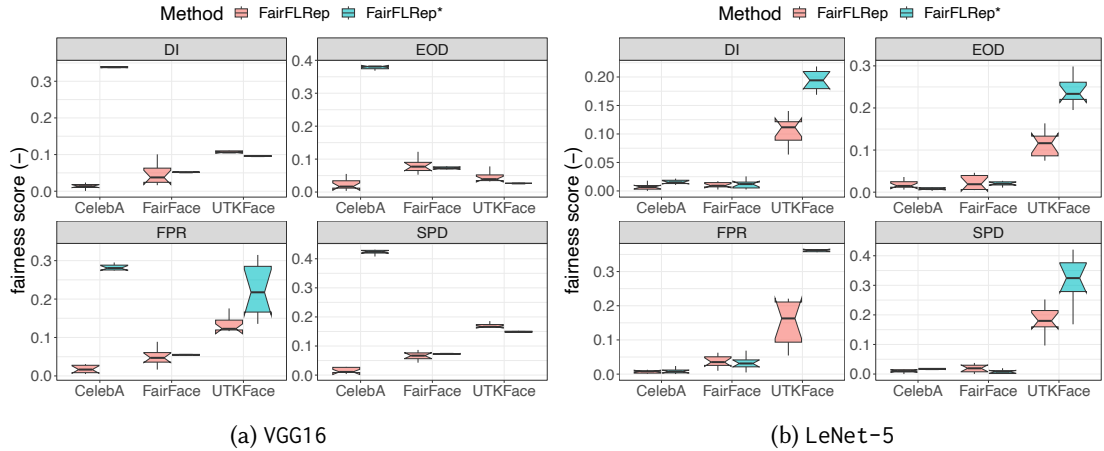

Fig. 14. RQ4 – `SettDiff` – Image datasets – Fairness outcomes when repairing only the last layer (`FairFLRep`) versus deeper layers (`FairFLRep∗`)

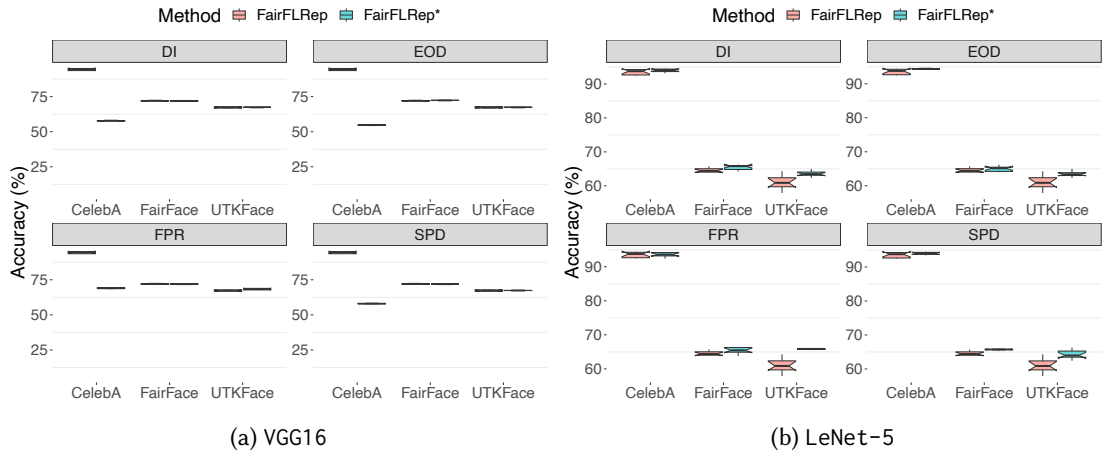

Fig. 15. RQ4 – `SettDiff` – Image datasets – Classification accuracy of models repaired using `FairFLRep` versus `FairFLRep∗`

*Repairing* VGG16. The same trend holds for VGG16: `FairFLRep` generally outperforms `FairFLRep∗` on fairness, particularly on `CelebA` and `UTKFace`. Accuracy-wise, according to Figure 15a, the two methods yield nearly identical accuracy overall. Notably, for `CelebA`, `FairFLRep` achieves visible gains in both fairness and accuracy, further demonstrating its effectiveness as a practical and balanced repair strategy.

> **Answer to RQ4 (`SettDiff` – Image datasets).** Under `SettDiff`, repairing the last layer is not only sufficient but often more effective in improving fairness without sacrificing accuracy. This reinforces `FairFLRep`'s suitability as a lightweight, reliable repair solution.

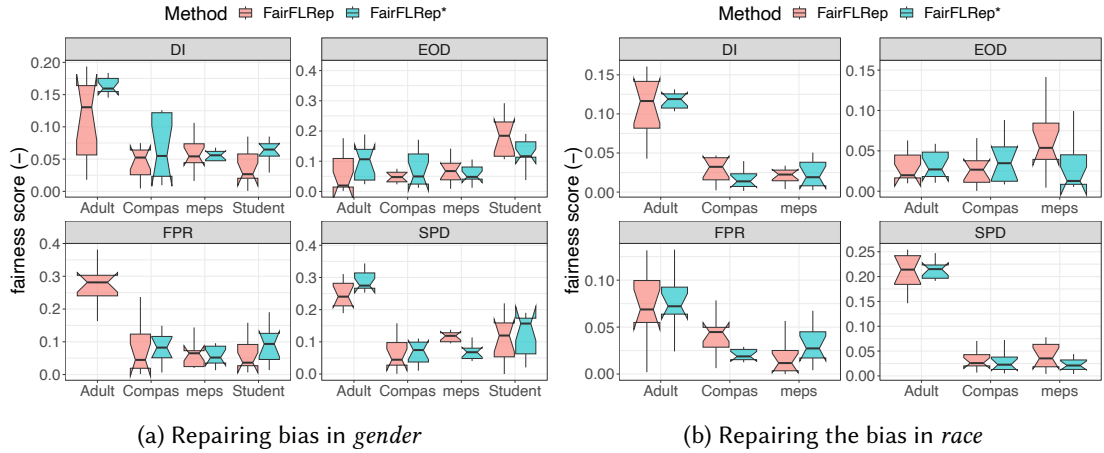

Fig. 16. R4 – `SettDiff` – Tabular datasets – Fairness outcomes when repairing only the last layer (`FairFLRep`) versus deeper layers (`FairFLRep∗`)

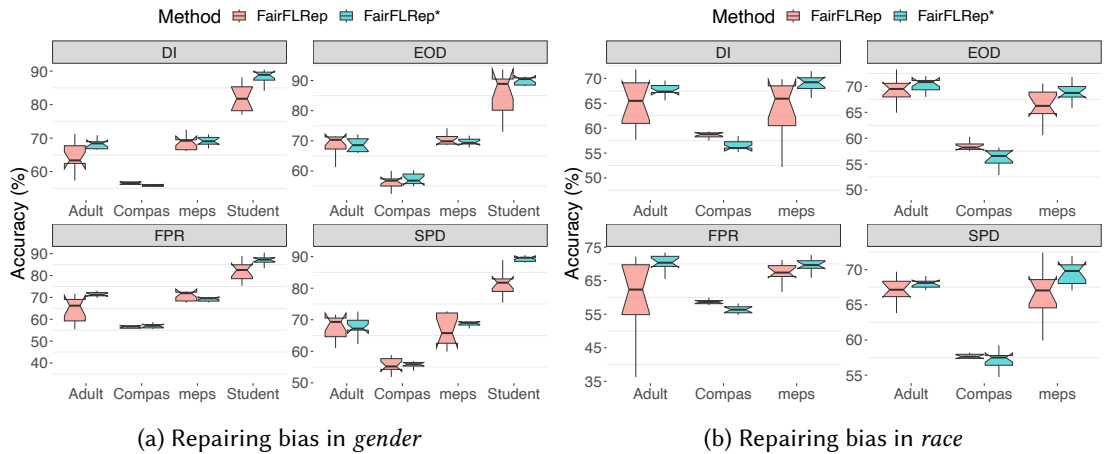

Fig. 17. RQ4 – `SettDiff` – Tabular datasets – Classification accuracy of models repaired using `FairFLRep` versus `FairFLRep∗`

6.5.3 `SettDiff` – *Tabular datasets.* Figure 16 compares the fairness outcomes of repairing bias related to *gender* and *race*, respectively, using our proposed method `FairFLRep` applied to the last layer, versus `FairFLRep∗` where the repair is applied to deeper or intermediate layers. According to the trends observed in Figure 16, `FairFLRep` generally achieves comparable or slightly better fairness outcomes across datasets. However, in some datasets (particularly MEPS and COMPAS), `FairFLRep∗` occasionally yields slightly better fairness improvements. This suggests that, for certain datasets, fairness gains may be more sensitive to adjustments made beyond the last layer.

To assess the trade-off between fairness and accuracy, Figure 17 reports the classification accuracy of models repaired using `FairFLRep` versus `FairFLRep∗`. Interestingly, while `FairFLRep∗` achieves better accuracy in some scenarios (most notably in MEPS and Student), the difference is not consistent. For Adult, and COMPAS, and even for MEPS when repaired for *gender*, `FairFLRep` performs comparably or occasionally better.

These results highlight a trade-off: deeper-layer repairs can improve both fairness and accuracy in some datasets (e.g., MEPS), but last-layer repair remains a strong, stable alternative, especially when simplicity and broad applicability are priorities.

The MEPS and Student datasets have significantly higher feature dimensionality (40 and 28 features, respectively) compared to COMPAS (6) and Adult (9). This difference likely contributes to the observed trend where deeper-layer repair outperforms or matches last-layer repair on MEPS and Student, but not necessarily on COMPAS or Adult, potentially due to:

- Models trained on datasets with many features often develop richer, deeper internal representations to capture intricate relationships among features. In such cases, bias or unfairness may be embedded not just at the output layer but throughout the intermediate layers of the network. Repairing only the last layer may thus be insufficient to fully address this embedded bias.
- With more features, especially if they interact or correlate, the model is likely to distribute relevant information across multiple layers. So, adjusting only the output may not shift the internal representations enough to produce fair outcomes. Repairing intermediate layers (as in FairFLRep∗) allows the model to relearn or reshape these internal feature interactions in a fairness-aware way.
- In contrast, datasets like COMPAS (6 features) and Adult (9 features) have lower dimensionality and possibly simpler decision boundaries. Biases in such models might be more "surface-level" and more easily correctable at the output layer without disturbing the rest of the network's structure. This explains why FairFLRep is often sufficient or even slightly better for these datasets.

> **Answer to RQ4 (SettDiff – Tabular datasets).** While FairFLRep (which repairs only the last layer) is generally effective at improving fairness with minimal disruption to model performance, dataset-specific characteristics can influence whether deeper-layer repairs (FairFLRep∗), targeting intermediate representations, offer additional advantages.

## 6.6 RQ5 – How efficient is FairFLRep in repairing fairness?

For any fairness repair technique to be practical in real-world applications, it must be efficient in terms of computational resources and time. This RQ examines the time it takes for FairFLRep to run in comparison to other approaches, with the goal of assessing whether FairFLRep can scale effectively to large datasets and complex models without imposing significant overhead.

*6.6.1 Image Datasets.* Table 8 compares the average time overhead (in minutes and seconds) of different repair techniques across the four image datasets and two models, with the values in brackets indicating the standard deviation of the time overhead across multiple runs (30 runs). According to the result, FairFLRep consistently shows faster execution times compared to Arachne across all datasets and models. FairArachne and FairFLRep exhibit very similar time overheads, often with marginal differences of a few seconds, while Arachne incurs significantly higher time overhead compared to both methods.

Despite both techniques running the same search problem with the same search budget, population size, and dataset, FairFLRep's faster performance can be attributed to its fitness function, which focuses on minimizing fairness violations. This differs from Arachne, which maximizes model accuracy, a goal that typically requires more extensive exploration. As a result, FairFLRep typically finds optimal solutions within 12 generations, while Arachne often requires 50 or more generations to reach similar optimization. This difference in convergence rates during the repair phase leads to `Arachne` taking significantly longer to complete the process.

Table 8. RQ5 – Image datasets – Comparison on Time Overhead (Minutes and Seconds) between `FairFLRep` and the Baselines.

| Dataset | Model | FairFLRep | FairArachne | Arachne | FairMOON |
|---|---|---|---|---|---|
| UTKFace | VGG16 | 5m 38s (0m 4s) | 5m 49s (0m 5s) | 30m 46s (17m 21s) | 19m 38s(4m 4s) |
| UTKFace | LeNet-5 | 4m 7s (0m 5s) | 4m 11s (0m 8s) | 17m 58s (18m 36s) | 7m 31s(0m 13s) |
| FairFace | VGG16 | 7m 15s (0m 4s) | 7m 15s (0m 4s) | 41m 18s (22m 52s) | 20m 22s(0m 57s) |
| FairFace | LeNet-5 | 5m 28s (0m 8s) | 5m 18s (0m 5s) | 14m 44s (3m 4s) | 8m 7s(0m 21s) |
| LFW | VGG16 | 5m 0s (0m 5s) | 4m 23s (0m 4s) | 22m 35s (0m 14s) | 16m 19s(1m 55s) |
| LFW | LeNet-5 | 2m 58s (0m 4s) | 2m 48s (0m 5s) | 15m 58s (0m 14s) | 5m 2s(0m 2s) |
| CelebA | VGG16 | 4m 50s (0m 10s) | 4m 34s (0m 9s) | 14m 29s (2m 12s) | 26m 56s(1m 47s) |
| CelebA | LeNet-5 | 3m 5s (0m 9s) | 3m 2s (0m 7s) | 9m 53s (1m 26s) | 5m 38s(0m 40s) |

For instance, in the `FairFace` dataset using the VGG16 model, `Arachne` takes 41 minutes and 18 seconds, while `FairFLRep` and `FairArachne` both finish in around 7 minutes and 15 seconds. This indicates that `Arachne` can be over five times slower than `FairFLRep`. The consistently lower time overhead of `FairFLRep` makes it highly scalable for large datasets and DNNs, which is crucial for real-world applications where computational resources and time are limited.

The close time performance between `FairFLRep` and `FairArachne` suggests that `FairFLRep`'s fair-aware strategy is as efficient as `FairArachne`'s during the fault localization phase. However, `FairFLRep`'s faster convergence in the repair phase compared to `Arachne` highlights the computational cost of `Arachne`'s accuracy-maximization approach, making `FairFLRep` the more efficient overall choice.

`FairMOON` also exhibits moderate time overhead across datasets. Its repair times are noticeably higher than `FairFLRep`, although still generally lower than `Arachne` in most cases. For example, on UTKFace (VGG16), `FairMOON` takes 19m 38s, compared to 5m 38s for `FairFLRep`. On FairFace (VGG16), `FairMOON` takes 20m 22s, compared to 7m 15s for `FairFLRep`. On CelebA (VGG16), `FairMOON` is actually slower than `Arachne` (26m 56s vs 14m 29s for `Arachne`). Although `FairMOON` shares the same fairness-aware repair phase as `FairFLRep` and `FairArachne`, its spectrum-based fault localization introduces additional computational costs. Specifically, `FairMOON` requires analyzing neuron activation spectra and prioritizing test inputs, which adds extra computation. While this approach is targeted toward fairness-aware input selection, it can incur substantial overhead, especially on larger or more complex datasets.

> **Answer to RQ5 – Image Datasets.** `FairFLRep` demonstrates a notable reduction in time overhead across all datasets and models compared to `Arachne` and `FairMOON`. This efficiency gain, combined with previous results showing improved fairness without harming accuracy, positions `FairFLRep` as a better candidate for real-world deployment, especially in environments where computational resources and time are critical factors, such as online systems or large-scale AI deployments.

*6.6.2 Tabular Datasets.* Table 9 reports the time overhead (in minutes and seconds) for `FairFLRep` and the baseline methods when repairing models across the tabular datasets under the `SettDiff` setting. The notation and structure follow that of Table 8.

Table 9. RQ5 – Tabular datasets – Comparison on Time Overhead (Minutes and Seconds) between `FairFLRep` and the Baselines.

| Dataset | Model | FairFLRep | FairArachne | Arachne | FairMOON | LIMI |
|---|---|---|---|---|---|---|
| Adult | Gender | 5m 11s (0m 27s) | 5m 16s (0m 14s) | 16m 37s (1m 26s) | 8m 34s (0m 32s) | 12m 55s (0m 30s) |
|  | Race | 5m 4s (0m 24s) | 5m 3s (0m 10s) | 21m 29s (3m 55s) | 8m 53s (0m 28s) | 11m 40s (0m 35s) |
| COMPAS | Gender | 2m 13s (0m 7s) | 2m 7s (0m 3s) | 17m 17s (1m 35s) | 5m 30s (0m 5s) | 0m 43s (0m 4s) |
|  | Race | 2m 1s (0m 1s) | 2m 8s (0m 2s) | 17m 48s (0m 33s) | 5m 25s (0m 9s) | 0m 43s (0m 3s) |
| Student | Gender | 0m 50s (0m 1s) | 0m 53s (0m 4s) | 4m 22s (0m 52s) | 5m 1s (0m 5s) |  |
| MEPS | Gender | 2m 22s (0m 5s) | 2m 41s (0m 4s) | 10m 49s (2m 6s) | 7m 5s (0m 13s) | 5m 35s (0m 9s) |
|  | Race | 2m 47s (0m 19s) | 2m 40s (0m 11s) | 13m 31s (3m 32s) | 7m 28s (0m 20s) | 15m 43s (0m 14s) |

According to Table 9, `FairFLRep` achieves competitive or lower mean time compared to the baselines across most datasets and sensitive attributes, while also maintaining low variability (small standard deviations). `FairArachne` exhibits time overheads very close to `FairFLRep`, often within a few seconds, because both methods share a similar fairness-aware repair phase but differ in fault localization. In contrast, `Arachne` incurs the highest time overhead, often significantly longer (e.g., 21m 29s ± 3m 55s for `Adult-Race`) compared to 5m 4s ± 0m 24s for `FairFLRep`. This overhead reflects `Arachne`'s exhaustive and non-fairness-aware repair search. `FairMOON` exhibits moderate runtime, faster than `Arachne` but slower than `FairFLRep`, due to its additional computational steps for spectrum-based neuron analysis. `LIMI` shows inconsistent performance efficiency—extremely fast on simpler datasets like `COMPAS` (~43 seconds mean) but significantly slower on some datasets such as `MEPS-Race` (15m 43s ± 0m 14s), suggesting limited scalability under increasing data complexity. While `LIMI` achieves rapid test input generation on simpler datasets by leveraging efficient latent space vector calculations (as reported in the original paper [71]), its efficiency declines on complex datasets where the surrogate boundary approximation becomes less accurate, requiring additional probing steps to find valid discriminatory instances. This explains the observed variability in `LIMI`'s time overhead across different datasets.

> **Answer to RQ5 – Tabular Datasets.** `FairFLRep` achieves a favorable balance between fairness performance and computational efficiency. It consistently offers lower or comparable mean repair times with lower variability, making it highly suitable for real-world scenarios where both fairness and efficiency are required.

## 7 Threats to Validity

*Construct Validity.* One potential threat to validity [69] lies in the choice of fairness metrics (e.g., *SPD*, *DI*, *EOD*, *FPR*) used to assess `FairFLRep`'s effectiveness. Although these metrics are widely used in fairness research, they represent group fairness and may not capture other notions, such as individual or causal fairness. This limited scope may influence our conclusions, as `FairFLRep`'s effectiveness is evaluated primarily through group-level metrics. If `FairFLRep`'s impact varies across different fairness definitions, this reliance on group-level metrics alone could overlook aspects like individual fairness or intersectional fairness. To mitigate this, we selected metrics that align with widely accepted definitions of fairness in classification tasks, acknowledging that future studies could benefit from exploring `FairFLRep`'s effects on a broader range of fairness metrics, especially those that capture intersectional and individual fairness nuances.

*Conclusion Validity.* This study primarily compares `FairFLRep` against `Arachne` and `FairArachne` as baselines to evaluate the effectiveness of our repair method. While `Arachne` is a relevant and established baseline, `FairArachne` is introduced as an internal ablation study combining `Arachne`'s fault-localization with `FairFLRep`'s repair component to provide a controlled comparison. This internal baseline helps assess `FairFLRep`'s specific contributions. However, the study does not encompass the full range of fairness repair techniques in the literature, such as adversarial debiasing, gradient-based repair, or fairness regularization approaches. Many of these techniques address fairness from different perspectives, focusing on problems like adversarial robustness or constraint-based fairness rather than model repair. A broader comparison could reveal additional insights, particularly concerning techniques designed for specific fairness types, such as individual or intersectional fairness. Future work will expand the range of comparisons to provide a more comprehensive assessment of `FairFLRep`'s effectiveness across diverse fairness contexts.

*Internal Validity.* This concerns the study design and methodology, specifically regarding consistency in experimental conditions and parameter settings across different baselines. Any variations in hyperparameters, repair and testing conditions, or model configurations during experiments could introduce unintended effects on `FairFLRep`'s performance relative to other methods. To address this, we standardized experimental settings, including identical search budgets, population sizes, and dataset splits across `FairFLRep`, `Arachne`, and `FairArachne`. All experiments were performed on the same Linux server with GPU support, and a warm restart was conducted for each new experiment. Additionally, a consistent approach to measuring fairness and accuracy across all experiments was applied to further minimize potential variability. Nonetheless, any unintentional inconsistencies in these settings could still affect the outcomes and interpretation of our findings.

*External Validity.* This threat refers to the ability to generalize the results to other network architectures and domains. A limitation of `FairFLRep` is its focus on CNN architectures. In this study, we primarily concentrate on CNNs, which are the dominant architecture for image classification tasks and are extensively used in fairness research. However, this focus may limit the method's applicability to other types of neural networks. In the future, we aim to extend our approach to additional network architectures, such as Recurrent Neural Networks (RNNs) for sequential data or other deep learning models used in different domains, to assess its broader applicability across various tasks and architectures. Additionally, although we evaluated `FairFLRep` on four widely used benchmark datasets (`UTKFace`, `FairFace`, `LFW`, and `CelebA`) and two popular models (`VGG16` and `LeNet-5`), generalizability to other datasets or models may be limited. To mitigate this, we deliberately selected these datasets and models due to their use in state-of-the-art fairness testing, as well as their diverse data distributions and biases.

## 8 Discussion and Conclusion

This work presents `FairFLRep`, a novel approach for mitigating bias in DNNs through an automated fault localization and repair technique. By identifying and adjusting the neuron weights that contribute to unfair behavior, `FairFLRep` effectively improves fairness across various sensitive attributes without significantly impacting model performance. The dual-categorization approach in `FairFLRep` enhances the flexibility of the fault localization method, allowing it to adapt to different types of bias by distinguishing cases where the sensitive attribute is the same as or different from the prediction class. This leads to a more nuanced understanding of the model's unfair behavior and enables tailored solutions for different types of biases. The two cases addressed by this approach are: (i) `SettSame` *(Same Attribute)*: Focuses on correcting misclassifications directly linked to the sensitive attributes. This helps reduce bias in predictions where the model disproportionately affects certain groups. (ii) `SettDiff` *(Different Attribute)*: Focuses on balancing the influence of sensitive attributes on predictions, ensuring underrepresented groups do not disproportionately influence the model's predictions. The evaluation shows that using `FairFLRep` for both fault localization and repair yields a more consistent and robust solution for improving fairness across different datasets, outperforming Arachne in most fairness metrics and datasets.

## 8.1 Key findings

The main findings of the paper are as follows:
- Arachne improves accuracy in some cases but struggles to fully address fairness issues, as seen in persistent gender bias.
- When fairness is considered only in the repair phase (i.e., FairArachne), there is a noticeable improvement in fairness across gender, age and race sub-groups, especially in datasets like LFW and UTKFace, where accuracy disparities between sub-groups become more balanced.
- Considering fairness in both the fault-localization and repair stages (i.e., FairFLRep) further minimizes disparities in accuracy within the sensitive community across datasets, with overall model accuracy either remaining competitive or improving compared to baselines.
- FairMOON, while occasionally matching or slightly outperforming FairFLRep on isolated metrics (e.g., *SPD* on LFW), shows high variability across datasets and fairness definitions. Its spectrum-based fault localization is too coarse for complex bias scenarios. It often sacrifices fairness for accuracy and underperforms in settings requiring stable subgroup fairness.
- Our findings support the effectiveness of last-layer repair as an effective, although deeper repairs offer some benefit in some scenarios. Moreover, our empirical evaluation on tabular datasets suggests that feature dimensionality and internal representation complexity are important factors when choosing the appropriate layer depth for fairness repair. This insight can guide future designs of fairness repair strategies–perhaps adaptive strategies that choose repair depth based on data complexity.

Additionally, FairFLRep demonstrates significant time efficiency across datasets and models, reducing time overhead compared to Arachne. It offers optimized fairness repair that is well-suited to real-world applications where computational resources are crucial.

## 8.2 Lessons learned

From the experimental results, we derived the following observations:
- *Single fairness metric optimization is sometime insufficient*: In the case of race disparities, our results clearly demonstrate that optimizing a model for one fairness metric often impacts other fairness metrics, sometime negatively. For example, when focusing on repairing *FPR*, it often results in worse outcomes for *DI* and *SPD* (and sometimes worsen *EOD* for tabular dataset). This shows that fairness is multi-dimensional, and addressing only one aspect can exacerbate other biases. This is particularly evident in more complex models like VGG16, where improvements in one metric (e.g., *DI*) often cause declines in others (e.g., *FPR* or *SPD*).
- Fairness metric behavior varies by modality: In image datasets, *SPD* and *EOD* tend to dominate in terms of general effectiveness. Optimizing for *SPD* often improves or at least maintains *EOD*, while *DI* and *FPR* are more volatile, often suffering large losses when other metrics are prioritized. These trends suggest that *DI* and *FPR* are harder to optimize in complex, high-dimensional tasks like face recognition, likely due to their sensitivity to specific group-level distributions.

    Contrastingly, in tabular datasets, *DI* and *SPD* emerge as the most generalizable and robust metrics. They consistently yield high fairness ties and minimal losses across other metrics, with limited impact on accuracy. *FPR* also preserves accuracy well but generalizes less consistently for fairness. *EOD*, while strong in image classification tasks, proves less reliable in tabular settings—frequently tying but rarely outperforming, and sometimes losing to simpler metrics. These findings suggest that simpler, group-level fairness definitions (e.g.,

*DI* and *SPD*) are better suited for structured, lower-dimensional tabular data, whereas more nuanced metrics like *EOD* are better aligned with the complex feature interactions present in image domains.

- *Model complexity affects fairness interactions*: The differences between `VGG16` and `LeNet-5` highlight that model complexity impacts the interactions between fairness metrics. The simpler `LeNet-5` model showed less severe losses across fairness metrics when a single fairness metric was repaired. This suggests that more complex models, such as `VGG16`, may amplify the trade-offs between fairness metrics, likely due to the model's ability to learn more nuanced (but sometimes biased) patterns.
- *Fairness metric correlations*: There is some evidence of positive correlations between certain fairness metrics. For instance, *DI* and *SPD* tend to show more similar results and are more likely to improve together. However, *FPR* and *EOD* show less consistency in their relationship, with improvements in one metric not necessarily leading to improvements in the other. This points to the need for a nuanced understanding of how different fairness metrics relate to each other.

## 8.3 Practical Implications

The findings of this study have several critical implications for researchers and practitioners aiming to develop fair, efficient, and robust AI systems:

- *Efficient repair*: `FairFLRep` demonstrates that fairness improvements can be achieved by modifying specific model weights, making it a highly efficient alternative for real-world applications. This weight-specific repair preserves much of the original model's learned behavior, reducing computational costs and time. For practitioners, adopting repair-based methods like `FairFLRep` enables faster and more cost-effective updates, particularly in production environments where other approaches like retraining may be resource-intensive and disruptive.
- *Adapting fairness repair for domain shifts*: Repair-based methods, like the approach used in `FairFLRep`, offer a promising way to handle domain shifts efficiently. Instead of requiring full model retraining, repair methods enable models to adapt to new data distributions and demographic shifts by adjusting only fairness-related weights. This dynamic approach maintains fairness as data characteristics evolve, offering a practical solution for scenarios where access to target domain data is limited or where computational resources are constrained.
- *Encouraging broader scope in fairness testing*: ML engineers and practitioners should broaden their approach to fairness testing by adopting a comprehensive range of metrics tailored to their application domain. Fairness testing should not be limited to a single metric but should cover various aspects of fairness, such as *SPD*, *DI*, *EOD*, and *FPR*. This can help capture a fuller picture of model fairness, especially in applications where biases could have real-world consequences.
- *Research focus on fairness metric effectiveness*: Future research should further investigate the relationships and interactions between various fairness metrics, examining how they impact different types of models and datasets. This is particularly important for applications where trade-offs across metrics may have significant social implications. Research should explore holistic repair algorithms that can balance improvements across multiple metrics while maintaining or enhancing model performance.
- *Societal and Legal Relevance*: `FairFLRep` aligns with increasing global emphasis on algorithmic accountability and fairness in AI, as emphasized by regulatory frameworks such as the EU AI Act [21], GDPR [22], and the U.S. EEOC guidelines [67]. They all stress the importance of non-discrimination and equitable treatment in automated decision-making. For example, the GDPR mandates that individuals should not be subject to decisions based solely on automated processing if those decisions have significant effects, while the EEOC requires fair treatment regardless of race, gender, or other protected attributes in employment-related algorithms. By targeting biased neural weights directly and offering subgroup-level fairness repair, `FairFLRep` supports compliance with such standards. Moreover, its ability to operate without retraining enhances auditability and post-hoc accountability–key elements in transparency-centered AI governance. The results also demonstrate reductions in disparity measures (e.g., DI < 0.2), aligning with legal fairness thresholds such as the "80% rule" commonly used to flag adverse impact in employment contexts.

In conclusion, `FairFLRep` offers a robust framework for fairness repair that caters to both academic and practical needs in developing fair and equitable DNNs. By applying these insights, researchers and practitioners can make substantial progress toward developing more socially responsible and fair AI systems for high-stakes applications.

## Acknowledgments

This research was supported by the DEEL Project (CRDPJ 537462-18), funded by the Natural Sciences and Engineering Research Council of Canada (NSERC) and the Consortium for Research and Innovation in Aerospace in Québec (CRIAQ), with contributions from Thales Canada Inc., Bell Textron Canada Limited, CAE Inc., and Bombardier Inc. Paolo Arcaini and Fuyuki Ishikawa are supported by the Engineerable AI Techniques for Practical Applications of High-Quality Machine Learning-based Systems Project (Grant Number JPMJMI20B8), JST-Mirai. Paolo Arcaini is also supported by the ASPIRE grant No. JPMJAP2301, JST.